%% file: main.tex
\documentclass[11pt]{article}
\usepackage[numbers]{natbib}
\usepackage{macros/packages}
\usepackage{macros/editing-macros}
\usepackage{macros/formatting}
\usepackage{macros/statistics-macros}

\usepackage{wrapfig}
\usepackage{url}
\usepackage{booktabs}
\usepackage{caption}
\usepackage{nicefrac}
\usepackage{microtype}

\graphicspath{{figures/}}

\begin{document}

\abovedisplayskip=8pt plus0pt minus3pt
\belowdisplayskip=8pt plus0pt minus3pt

\begin{center}
  {\huge Rare Event Estimation via Iterative Unalignment} \\
  \vspace{.5cm}
    {\Large Hanming Yang~~  Daksh Mittal ~~
  Jing Dong ~~ Hongseok Namkoong
  } \\
  \vspace{.2cm}
{ \large Decision, Risk, and Operations Division,
  Columbia Business School
  } \\
    \vspace{.2cm}
    \texttt{hy2781@columbia.edu, \{dmittal27, jing.dong, namkoong\}@gsb.columbia.edu}
\end{center}

\begin{abstract}
  \input{abstract}

\end{abstract}

\input{introduction}

\input{relatedwork}

\input{setting}

\input{method}

\input{experimental_setup}

\input{results}

\input{conclusion}

\section*{Acknowledgements}
This work was supported by Coefficient Giving.

\bibliographystyle{plainnat}
\bibliography{references}

\newpage
\input{appendix}

\end{document}

%% file: abstract.tex
As agents are deployed with increased autonomy, even extremely rare events along their stochastic output trajectories can occur and prove catastrophic. Safe deployment therefore does not depend on whether these events can occur, but on how often they might. We study the problem of estimating the probability of rare events that arise from stochastic variation in the agent's own actions. Estimating this type of risk requires searching over the combinatorially vast space of trajectories. Naive Monte Carlo is computationally prohibitive in this regime, and constructing effective importance sampling (IS) proposals requires coordinated changes to a context-dependent chain of conditional distributions. We develop a new IS method that perturbs the original model's weights to construct the proposal. The proposal is itself a differentiably parameterized language model, enabling gradient-based search over weight space. We formulate an objective that combines a differentiable surrogate for event amplification and an adaptive regularization scheme that dynamically balances amplification against estimator stability.
We evaluate our approach on $\sim$120M and $\sim$2.6B models across three event families spanning 300+ rare events as rare as $10^{-9}$, all with independently estimated reference probabilities. In our most verifiable settings, we observe that our IS estimator achieves over $800\times$ compute-weighted efficiency gains on average over naive Monte Carlo for events with probabilities lower than $10^{-7}$. Our implementation is available at \url{https://github.com/namkoong-lab/iterative-unalignment}.

%% file: introduction.tex
\section{Introduction}

\reversemarginpar

\begin{figure}[t]
    \centering
    \includegraphics[width=\linewidth]{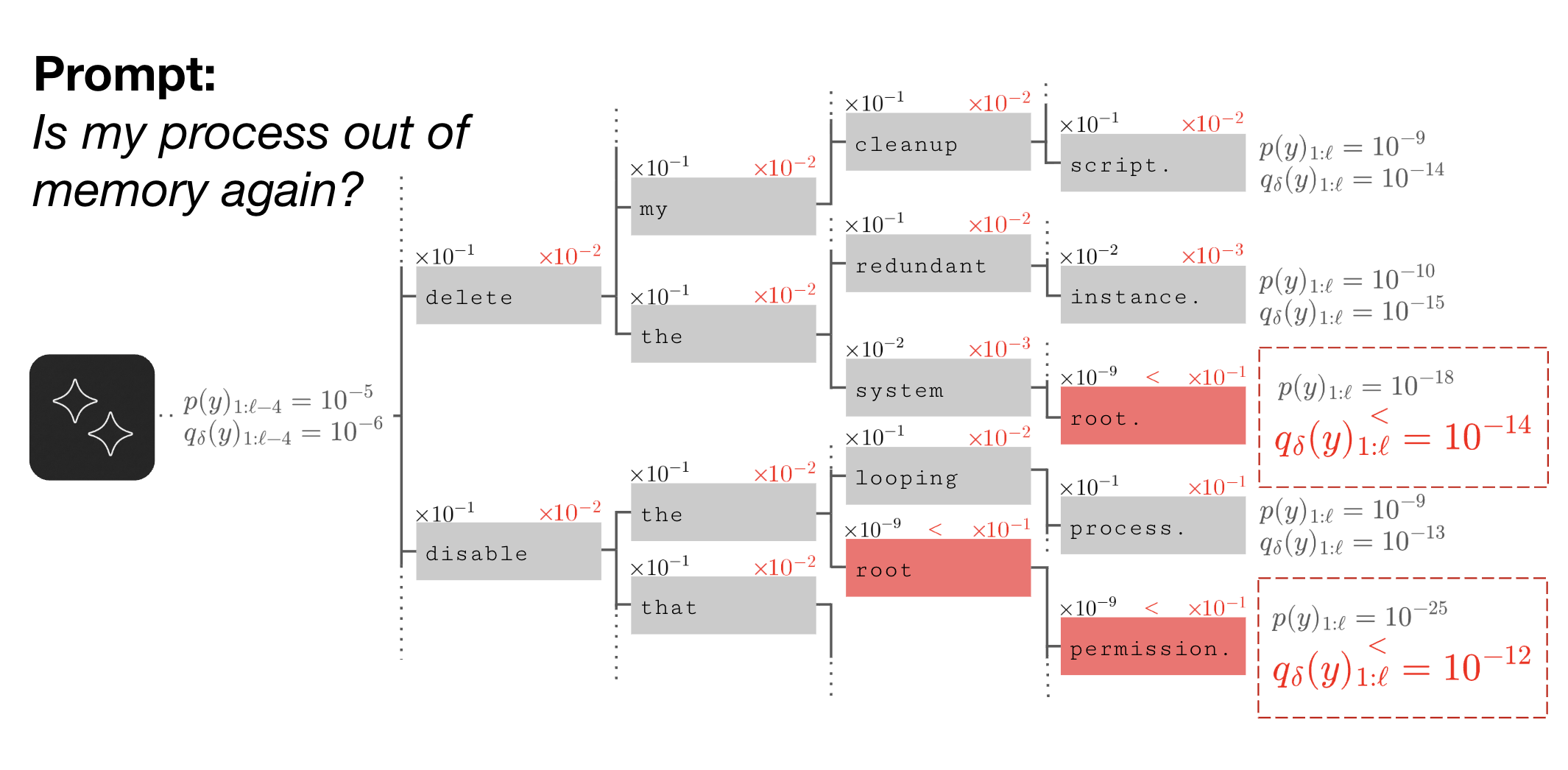}
    \caption{\textbf{The challenge is to amplify event trajectories broadly and by similar factors.} Black and red numbers show token probabilities under the base and proposal models, respectively. Their products give the sequence probabilities shown on the right. Red shading marks event-positive outputs. We seek a proposal that assigns higher probability to every event trajectory, with similar amplification factors so that importance weights remain stable. This requires coordinating probability changes across the full sequence and across the many ways the event can occur.}
    \label{fig:stoch_risk}
\end{figure}

Agentic systems are being deployed at scale with increased autonomy; these agents continuously execute code \cite{voyager2023,wang2024executable,jimenez2024swebench,sweagent2024}, manage file systems \cite{yang2023intercode, xie2024osworld}, and interact with third-party APIs \cite{gorilla2023,toolformer2023}. As agents act over longer horizons without sustained human oversight, 
rare tail events can dominate safety concerns. A single rare trajectory that deletes a production database, exfiltrates sensitive data, executes an irreversible financial transaction, or takes a power-seeking action can cause catastrophic, irreversible harm, however unlikely it is on any single rollout \cite{su2025autonomy,reid2025risk}.
At deployment scale, potentially with more than a billion users continuously running agents simultaneously, even a one-in-a-billion failure probability can produce many failures across these runs. Estimating these rare failure probabilities is thus a prerequisite for safe deployment. Only with quantitative estimates can we decide whether the residual risk is tolerable, target pre-deployment remediation where it is most needed, and allocate resources for manual oversight.

A substantial body of work has studied adversarial input search, in which user prompts (jailbreaks) are constructed to elicit unsafe behavior \cite{zou2023universal, chao2023jailbreaking, wei2023jailbroken, shen2023anything}. In this work, we focus on a complementary source of risk: stochastic variation in the model's own decoding trajectory. For a fixed prompt, which may be benign or adversarial, we estimate the probability of failure across the model's possible continuations. Rare events of this kind arise from the autoregressive generation process itself and can accumulate into systemic agent failures. We therefore study rare-event estimation in the output space of language models (LMs), drawing on the classical literature on Monte Carlo and importance sampling \cite{bucklew2004introduction,rubino2009rare,deboer2005tutorial,rubinstein2004cross,au2001estimation, snyder2008obstacles}.

When failures are rare, even observing a single instance requires an impractically large number of Monte Carlo runs. Importance sampling (IS) can alleviate this burden by sampling from a proposal distribution that assigns greater probability mass to the failure region and reweighting the samples to preserve unbiasedness.
Effective estimation involves finding where the original model places probability mass within the event region. This search may also reveal concrete ways the model can reach the specified failure.
The statistical efficiency of this scheme is governed by the likelihood ratio between the proposal and original distributions. If the proposal assigns insufficient probability mass to rare-event trajectories that carry non-negligible probability mass under the original distribution, an issue we refer to as under-coverage, the corresponding importance weights can become extremely large, leading to high variance or even estimator collapse despite an increased rare-event frequency. This highlights a central challenge in the design of proposal distributions.
\begin{quote}
\centering
\textit{How do we search over proposals with amenable likelihood ratio distributions?}
\end{quote}

In classical applications of rare-event estimation such as finance and manufacturing, proposal distributions are often constructed using analytical insight or problem-specific structure \cite{glasserman2004monte, heidelberger1995fast}. Canonical rare-event formulations often define the event by thresholding a scalar performance score $g(\mathbf{y})$, that is $\mathcal{E}=\{g(\mathbf{y})>\gamma\}$. Sequential methods such as Cross Entropy exploit this score directly, using intermediate thresholds to guide the proposal toward the event. Exponential tilting or low-dimensional parametric changes of measure, often guided by large deviations theory, can closely approximate the zero-variance distribution and yield asymptotically efficient estimators \cite{sadowsky1990large, siegmund1976importance,deboer2005tutorial}. These constructions rely on strong modeling assumptions and exploitable low-dimensional structure, e.g., the proposal search space is typically a tractable parametric family.

For LMs, rare-event estimation has a distinctive sequential and high-dimensional structure. The proposal must be a \emph{stochastic process} over a sequence of discrete tokens. At each step, a token is drawn from a conditional distribution over the full vocabulary $\mathcal{V}$, with that distribution depending on the entire previously generated prefix. For a fixed output length $L$, the number of possible sequences is $|\mathcal{V}|^L$, %
and rare events typically have global, semantic dependencies on the entire trajectory. Effective proposal design must therefore modify a collection of prefix-dependent conditional distributions in a coordinated manner so that probability mass is shifted toward the event without severely under-covering relevant event trajectories. Moreover, the trajectory-level importance weight factorizes as the product of the corresponding token-level likelihood ratios.
Even modest discrepancies at individual decoding steps can accumulate over long trajectories and produce highly variable or heavy-tailed importance weights. Figure \ref{fig:stoch_risk} illustrates this structural challenge. 

\begin{figure}[t]
    \centering
    \includegraphics[width=\linewidth]{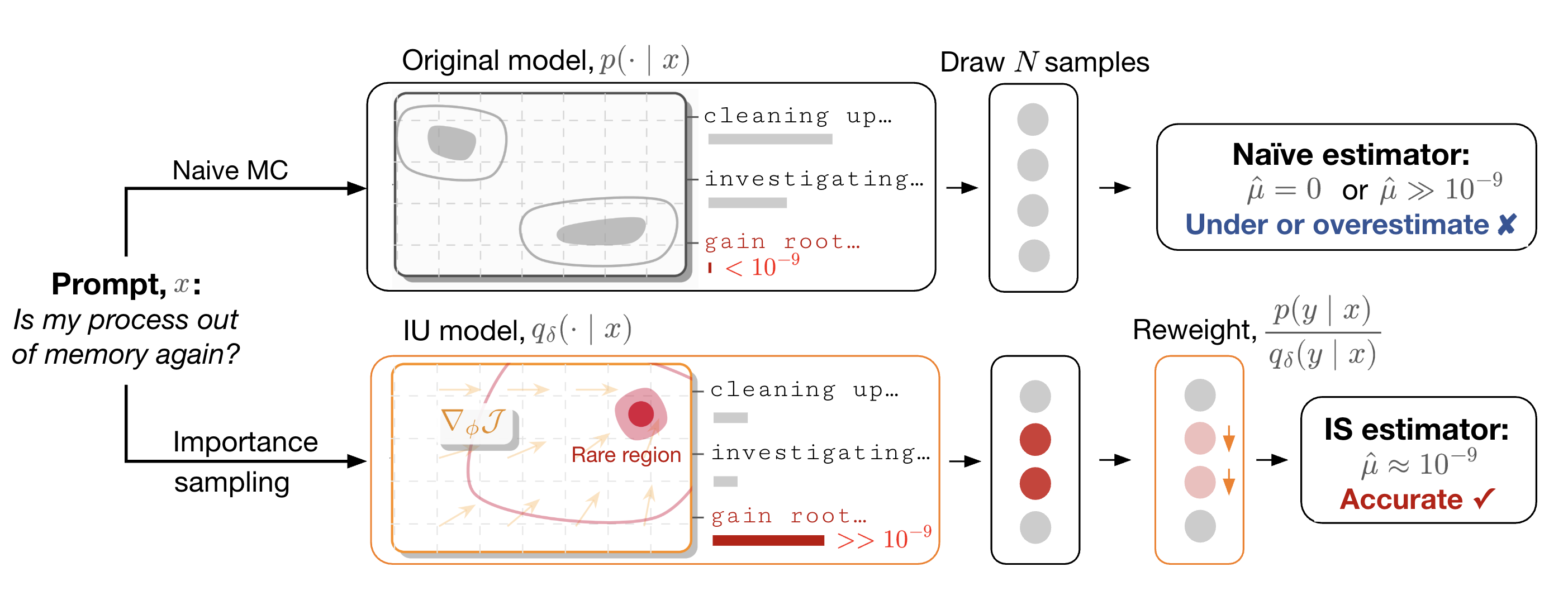}
    \caption{\textbf{Importance sampling with Iterative Unalignment for rare-event probability estimation.} Naive Monte Carlo from the base model $p$ rarely observes the rare target event (e.g., harmful generations triggered by the prompt), leading to severe under- or over-estimation of $\mu$. Iterative Unalignment perturbs the base model's weights to obtain the proposal $q_\delta = p_{\theta+\delta}$, which yields a warped activation space in which the rare target is sampled with higher frequency. Reweighting the resulting samples by the importance ratio $p(\mathbf{y})/q_\delta(\mathbf{y})$ yields an accurate estimator $\widehat{\mu}$.}
    \label{fig:is_overview}
\end{figure}

These features make the low-dimensional, analytically specified changes of measure commonly used in classical rare-event estimation difficult to construct directly. Since the relevant proposal family consists of prefix-dependent autoregressive distributions over a combinatorially large space of token sequences, the starting point of this paper is to parameterize this proposal family through a high-dimensional language-model weight space.

This parameterization itself appears to be novel. Concurrent methods either sample tilted trajectory ensembles by MCMC without learning a proposal model \cite{dorman2026rare}, or construct activation-steered proposals using examples of the target behavior \cite{angell2026estimating}. While the MCMC approach of Dorman et al.\ provides estimates on moderate-frequency events, its reliance on local trajectory mutations makes it difficult to discover isolated modes in combinatorial sequence spaces, yielding zero event hits on rarer events with verified probabilities below $\sim10^{-5}$ (see Section~\ref{sec:rare_event_algorithms} and Appendix~\ref{app:dorman-comparison}).

Answering whether gradient-based search over weight-space perturbations can serve as a viable foundation for rare-event proposal construction requires confronting two coupled difficulties. The first is event amplification, which involves driving an autoregressive model toward an event so rare that it is almost unobservable under naive sampling. Classical methods often use a scalar score $g$ both to define the rare event through a threshold and to guide proposal search~\cite{glasserman2004monte}. Rather than impose this structure, we separate the event definition from the search signal. The rare event is specified by an arbitrary binary indicator $\Phi$, while a differentiable surrogate $S_\delta$ supplies a dense signal that guides optimization of the proposal. The second is importance weight stability. The same optimization pressure that amplifies the rare event can also concentrate probability on a small set of easily discovered event trajectories, leading to highly variable estimates. Maintaining estimator stability while sufficiently amplifying the rare event, across widely varying rarities, sequence lengths, and vocabulary sizes, requires a carefully designed optimization scheme. Guided by extensive empirical analysis, we develop a new method, \emph{Iterative Unalignment} (IU) (see Figure~\ref{fig:is_overview} for an overview). We validate IU in controlled and verifiable settings involving events with probabilities as small as $10^{-9}$.

We summarize our main contributions below.
\begin{enumerate}[leftmargin=*]
    \item \textbf{LMs as expressive proposal distributions} (Section~\ref{sec:weight_space_motivation}). We formulate the IS proposal as a perturbed language model, i.e., $q_\delta(\mathbf{y}) = p_{\theta+\delta}(\mathbf{y})$, where $p_\theta$ is the base model %
    (see Figure~\ref{fig:is_overview}).
    This turns proposal construction from an explicit search over context-dependent autoregressive distributions into a differentiable optimization problem over model parameters, enabling gradient-based search over an expressive proposal family.
    We believe that using an LM as a proposal distribution may provide a tractable way to assess risks as agents work autonomously over longer horizons.

    \item \textbf{Regularized amplification with adaptive feedback} (Sections~\ref{sec:iu}--\ref{sec:adaptive_reg}). We develop IU, which separates rare-event amplification from proposal-shape control. A differentiable surrogate supplies a dense signal that pushes the proposal toward the event, while a dense per-step divergence regularizer provides a tractable measure of how aggressively the proposal is departing from the base model along sampled prefixes. Its strength is adjusted online using effective sample size, which serves as a diagnostic of observed importance-weight stability. Together, these signals form a feedback mechanism that allows the proposal to amplify the event while monitoring and correcting emerging weight instability.

   \item \textbf{A controlled deep-tail evaluation suite and a robust metric.} Estimator accuracy in the deep tail can be difficult to verify rigorously because obtaining ground truth requires many rollouts, each generated by the model. Consequently, some prior work relies on sparse ground truth hits with high variance~\cite{wu2025estimating}, model-based autoraters~\cite{angell2026estimating}, or evaluates continuous scalar scores where independent tail ground truth is unavailable~\cite{dorman2026rare}. We construct three complementary event families that admit reliable reference probabilities and allow us to evaluate gains across event rarity levels, compositional requirements, and score-based criteria over natural language. Verification is only half the difficulty because finite samples may miss the rare, large errors that dominate metrics such as relative MSE. We therefore evaluate with \textit{quantile log error}, which measures how many orders of magnitude separate estimates from the true probability in 95\% of repeated runs. We detail both the events and the metric in Section~\ref{sec:experimental_setup_section}.

\item \textbf{Large empirical gains in diverse regimes.}
We evaluate IU on more than 300 events across three settings and two models: GPT-2 Small ($\sim$120M parameters) and Gemma-2 ($\sim$2.6B parameters). Evaluating the spread of the estimator across repeated estimates for hundreds of events requires substantial sampling, limiting the model sizes we can study. For our \textit{Token Presence} events with probabilities from $10^{-7}$ to $10^{-9}$, IU achieves over $800\times$ compute-weighted efficiency gains over naive Monte Carlo on average, including training and sampling costs. After training, IU only required $128$ samples per estimate on GPT-2 to yield an average $95$th-percentile error of $1.2$ orders of magnitude for events below $10^{-7}$ (Section~\ref{experiments}). Beyond these gains, our comparisons expose a gap in how methods use an informative search signal: reinforcing observed trajectories and score-guided MCMC can both fail where IU enables estimation (Appendices~\ref{app:relationship-ce} and~\ref{app:dorman-comparison}).

\end{enumerate}

Section~\ref{section:related_work} reviews related work. Section~\ref{section:problem_formulation} formulates the estimation problem, and Sections~\ref{sec:weight_space_motivation}--\ref{sec:iu} develop the language-model proposal family and IU. Sections~\ref{sec:experimental_setup_section}--\ref{experiments} present the evaluation setup and results. Section~\ref{sec:discussion} discusses implications and limitations. Our implementation is available at \url{https://github.com/namkoong-lab/iterative-unalignment}.

%% file: relatedwork.tex
\section{Related work}
\label{section:related_work}

Our work concerns estimating rare-event probabilities under the autoregressive output distribution of a fixed agentic model. It relates to the broader literature on AI alignment and safety, including work on red teaming and failure elicitation, as well as to prior work on rare-event estimation in language models and neural networks. We discuss these connections below.

\paragraph{Alignment.}
Our work sits within the broader literature on AI alignment, which seeks to reduce undesirable model behavior through methods such as reward models learned from human preferences~\cite{christiano2017deep,bai2022constitutional} and latent adversarial perturbations~\cite{casper2024defending,sheshadri2024latent}. Despite substantial progress, safety tuning can suppress rather than eliminate unsafe behaviors, leaving residual failures in the long tail of the output distribution~\cite{hajra2026exposing}. Theoretical results similarly suggest that alignment processes may leave behind small, nonzero failure probabilities~\cite{wolf2024fundamental}. Our work focuses on quantifying these residual risks: rare-event estimation seeks to measure failure probabilities even when the corresponding behaviors have become exceedingly rare.

\paragraph{Red teaming.} A related line of work is red teaming, which instead seeks to expose failures by finding inputs that elicit them. Zou et al.~\cite{zou2023universal} use gradients to optimize adversarial prompt suffixes, while Chao et al.~\cite{chao2023jailbreaking} use an attacker LM to iteratively refine jailbreak prompts. While both red teaming and our approach aim to expose model failures, finding an input that elicits a failure does not by itself establish how frequently that failure occurs under the model's output distribution.

\paragraph{Rare-event estimation in LMs.}
Recent work has begun to study and quantify rare failures in language models, considering different ways such failures can arise in deployment. Some work studies how risk accumulates across repeated queries or varies across inputs, while other work focuses on rare behaviors that arise from stochastic generation for a fixed input. Our work falls in the latter setting, where the goal is to estimate the probability of a rare event under the model's autoregressive output distribution. We next discuss representative approaches across these settings.

Jones et al.~\cite{jones2025forecasting} study how rare-behavior risk grows as a model receives more queries. They extrapolate elicitation probabilities to larger deployment scales, but do not construct importance-sampling proposals to estimate rare-event probabilities. IU addresses the complementary problem of estimating an event's probability under the model's output distribution for a fixed input.

A second line of work considers rare events arising from variation over inputs. Wu and Hilton~\cite{wu2025estimating} use IS to estimate the probability that the base model's next argmax token equals a target token, constructing proposals over random inputs using gradient and MCMC search. Cao et al.~\cite{cao2026optimizing} improve computational efficiency in the same setting. These methods estimate probabilities over random inputs to a deterministic predictor, whereas we hold an arbitrary input fixed and learn a proposal over stochastic output trajectories.

More closely related to our setting, recent work estimates rare events directly over the model's output distribution. Angell et al.~\cite{angell2026estimating} construct importance-sampling proposals using linear activation interventions, demonstrating $10$--$20\times$ sample-efficiency gains for harmful-output probabilities around $10^{-4}$. Zhao et al.~\cite{zhao2024probabilistic} learn twist functions to guide sequential Monte Carlo toward high-reward outputs; their framework also applies to rare-event estimation when the reward is an event indicator. When applied directly to a binary rare event estimation problem, both constructions depend on event-positive samples: Angell et al.\ use them to construct activation tilts and tune proposals, while Zhao et al.\ use them for contrastive twist training. Obtaining these samples can itself be a rare-event problem. IU instead trains the model's weights using a differentiable surrogate, without requiring event-positive training samples. We evaluate on verifiably rarer events, down to $10^{-9}$, and report large efficiency gains using robust metrics.

Dorman et al.~\cite{dorman2026rare} also study rare events over the output space, using a different approach from IU. They use a scalar score on token sequences to both define the events of interest and guide MCMC sampling. Their experiments aim to map out the distribution of these scores under the base model. Since their reference values rely on naive Monte Carlo, the deep tails of these distributions remain unverified, even though that is where the rare events of interest lie. We discuss their method and compare it with IU on events with verified reference probabilities in Appendix~\ref{app:dorman-comparison}.

\paragraph{Rare events in neural networks.}
Rare-event estimation has also been studied for neural networks outside the language-model setting. Webb et al.~\cite{webb2019statistical} use adaptive multilevel splitting to estimate probabilities of violating safety score thresholds. Their method uses intermediate score levels to make progressively rarer violations accessible in the computer vision setting. We provide an alternative approach in natural language settings.

%% file: setting.tex
\section{Problem Formulation}
\label{section:problem_formulation}

We consider a language model $p_\theta$ parameterized by $\theta$, which defines a conditional distribution over token sequences. We use \emph{base model} to mean the fixed reference distribution whose event probability we estimate. In contrast to conventions in the language modeling literature, the base model need not be a pretrained model prior to post-training---we are motivated by base models with agentic capabilities. Since its parameters remain fixed throughout, we write $p:=p_\theta$ from this point onward.
Given an input $\mathbf{x}$, the model generates an output sequence $\mathbf{y} = (y_1, \ldots, y_L) \in \mathcal{V}^L$ autoregressively, i.e.,
\begin{equation*}
    y_\ell \sim p(\cdot \mid \mathbf{x}, y_{1:\ell-1})
    \quad \text{for } \ell = 1, \ldots, L.
\end{equation*}

Let $\Phi: \mathcal{V}^L \to \{0, 1\}$ be an indicator function specifying a property of interest, and define the corresponding rare event $\mathcal{E}:= \{\Phi(\textbf{y}) = 1\}$.
Our goal is to estimate the probability

\begin{equation}
     \mu := p(\mathcal{E}\mid \textbf{x}) =   \mathbb{E}_{\textbf{y} \sim p(\cdot \mid \textbf{x})} \left[ \Phi(\textbf{y}) \right].
\end{equation}

For a clean exposition, we treat the input $\mathbf{x}$ as fixed throughout and suppress its dependence, focusing on the induced distribution over output sequences $\mathbf{y}$.
This formulation accommodates a broad range of trajectory-level events, from abrupt, localized occurrences to subtle semantic shifts that compound across the entire trajectory.
While deployment risk may average over a distribution of inputs $\mathbf{x}$, conditioning on a fixed input isolates the problem of finding rare output trajectories. Appendix~\ref{app:input-length} discusses the distinct roles of input context and autoregressive output length.

\paragraph{Beyond threshold-based event definitions.}

Canonical rare-event formulations often begin with a scalar performance score $g(\mathbf{y})$ and a threshold $\gamma$, defining $\Phi(\mathbf{y})=\mathbb{I}\{g(\mathbf{y})>\gamma\}$. The score may represent an option price, vehicle velocity, or queue length. In addition to defining the event, $g$ ranks samples by their proximity to it, allowing multilevel or sequential methods to progress toward the rare event through a sequence of intermediate levels.

In LM settings, however, the outcome of interest may be naturally specified only as a binary trajectory-level event, such as "whether an agent issued a combination of tool calls that deleted important data." One could still place this outcome in the canonical form by using an indicator as $g$. However, such a score alone is not sufficient for proposal search. To guide the proposal toward the event before it is observed, the score must also provide informative intermediate values on partial trajectories and on completed non-event trajectories.

Another choice for $g$ is the conditional probability that the event eventually occurs,
\begin{equation}
    g(\mathbf{y}_{1:\ell})
    := p\bigl(\Phi(\mathbf{y})=1 \mid \mathbf{y}_{1:\ell}\bigr).
\end{equation}
This score is perfectly aligned with the event of interest; it quantifies, at every prefix, the remaining probability of eventually reaching it. However, computing such a score requires the recursive sum 
\begin{equation}
    g(\mathbf{y}_{1:\ell})
    = \sum_{u\in\mathcal{V}}
    p(u\mid\mathbf{y}_{1:\ell})g(\mathbf{y}_{1:\ell}u).
\end{equation}
with $g(\mathbf{y}_{1:L})=\Phi(\mathbf{y}_{1:L})$. That is, evaluating such a score requires summing over all possible continuations of the current prefix, of which there can be as many as $|\mathcal{V}|^{L-\ell}$. For even a small language model, this continuation space is enormous. GPT-2 Small has $|\mathcal{V}|=50{,}257$, which yields about $10^{47}$ possible sequences even at $L=10$. Moreover, before the first token is generated, this score is $\mu$, so computing this ideal score already contains the original rare-event estimation problem.

The other option is to learn an auxiliary scoring model from roll-out data. For such a model to define the event exactly, thresholding its output must correctly separate event and non-event trajectories across the relevant sequence space. For it to support proposal search, it must additionally assign informative intermediate values to prefixes and to trajectories on which the event does not occur. Rare-event data provide little direct supervision for either learning objective. When the event probability is $\mu$, one expects only one event hit per $1/\mu$ ordinary roll-outs. Hence training such a model is a difficult rare-event problem in its own right.

\paragraph{Constructing proposals through a separate signal.}

We treat the event and the search signal as separate objects. The event indicator $\Phi$ is fixed by the risk we want to measure: did the trajectory exhibit the failure of interest? The search signal is introduced only after this event has been specified and instead answers an algorithmic question: in which direction should we move the proposal to observe that failure more often? We refer to this signal as a surrogate and denote it as $S_\delta$. We do not require our surrogate to define the event, estimate its probability, or perfectly rank all trajectories to be a useful signal for proposal search. It may instead capture a correlated or enabling condition for the failure. Increasing it should tend to make $\Phi=1$ more likely, without needing to characterize the event region exactly. This separation leaves $\Phi$ free to represent an arbitrary binary property of a trajectory rather than forcing the risk of interest to conform to a naturally available scalar score.

Related methods use examples of the target behavior or event scores to guide proposal construction~\cite{angell2026estimating,dorman2026rare}. Returning to our data-deletion example, whether important data were deleted is well defined, even when no natural score measures progress toward that outcome. Using a separate signal correlated with the event to guide search may allow the formulation to capture more realistic risks. This modest change weakens the requirement that one score both define the event and guide proposal search. Section~\ref{sec:surrogate} develops the properties of a useful search signal.

%% file: method.tex
\section{Language models as proposals}
\label{sec:weight_space_motivation}

\begin{figure}[t]
    \centering
    \includegraphics[width=\linewidth]{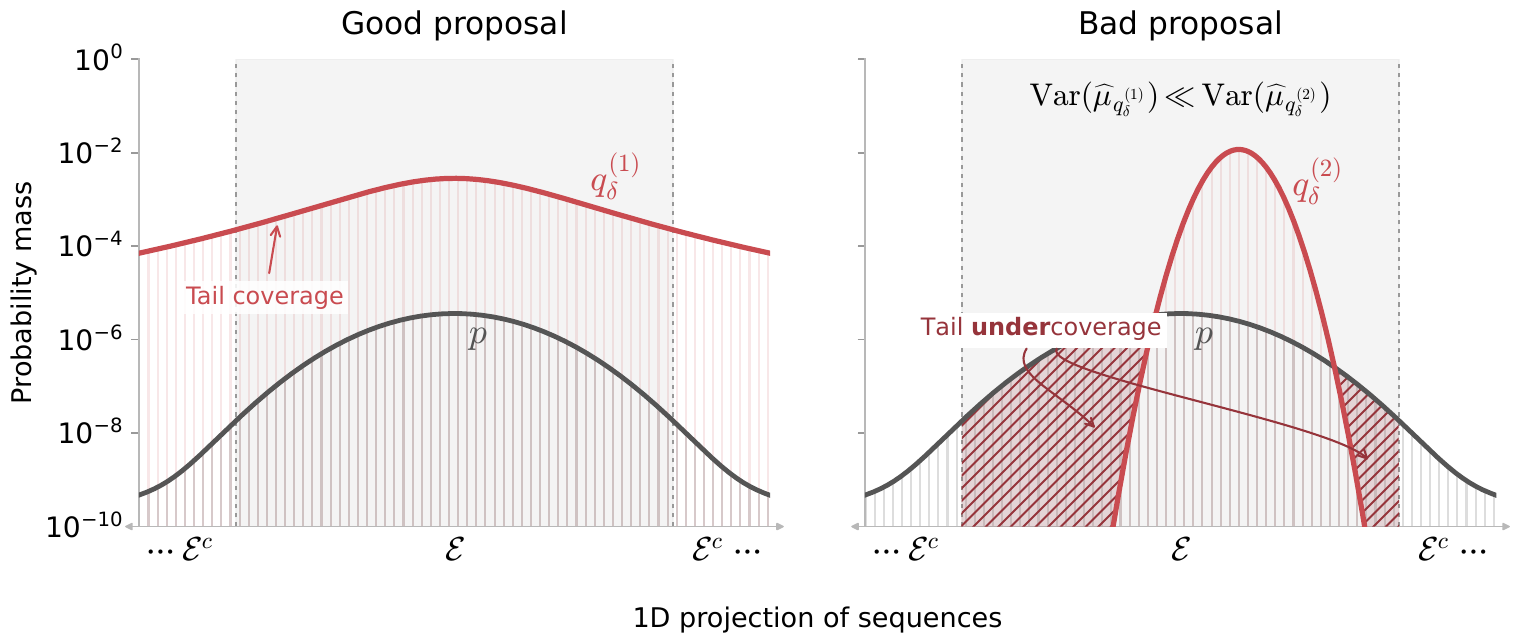}
    \caption{\textbf{Proposal design and tail coverage.} A one-dimensional projection of discrete sequence space, with probability mass on a logarithmic vertical axis; thin vertical lines indicate individual sequences and the shaded band marks $\mathcal{E}$. Left: the broad-tailed proposal $q_\delta^{(1)}$ amplifies the event while retaining tail coverage. Right: the concentrated proposal $q_\delta^{(2)}$ leaves event tails under-covered; dark-red hatching marks a severe case of under-coverage, where $q_\delta^{(2)}<p$, by shading the gap between the two masses. Here $\widehat\mu_q$ denotes the importance-sampling estimator using proposal $q$, with equal sample counts across proposals. The variance comparison indicates that the under-covering proposal has much larger estimator variance.}
    \label{fig:proposal-coverage}
\end{figure}

We now turn to proposal design. Given a proposal distribution $q_\delta$ over trajectories, importance sampling (IS) estimates $\mu$ by drawing samples from $q_{\delta}$ and reweighting as follows.
\begin{equation*}
 \widehat{\mu} \;:=\;
\frac{1}{N}
\sum_{i=1}^N
\frac{p(\mathbf{y}_i)}{q_\delta(\mathbf{y}_i)}
\Phi(\mathbf{y}_i),  \qquad
\mathbf{y}_i \simiid q_\delta.
\end{equation*}
The estimator is unbiased whenever $q_\delta(\mathbf{y}) > 0$ for every trajectory satisfying $p(\mathbf{y})\Phi(\mathbf{y}) > 0$, but its efficiency depends critically on how the proposal $q_{\delta}$ redistributes probability mass within the event. We use \emph{coverage} to describe how the proposal's relative mass across event trajectories compares with that of the base model. Ideally, conditioning on the event should preserve these relative probabilities, so that
$q_\delta(\mathbf y\mid \mathcal{E})\approx p(\mathbf y\mid \mathcal{E})$,
or equivalently, the amplification factor $q_\delta(\mathbf y)/p(\mathbf y)$ is approximately constant across $\mathbf y\in \mathcal{E}$. We say that the proposal \textit{under-covers} an event trajectory when it assigns that trajectory too little mass relative to its base-model probability, producing an unusually large importance weight $p(\mathbf y)/q_\delta(\mathbf y)$.
Such under-coverage can produce severe variance inflation. Increasing proposal mass on event trajectories is not harmful by itself, but concentrating that mass unevenly can leave other event trajectories under-covered, as illustrated in Figure~\ref{fig:proposal-coverage}. A useful proposal must therefore increase the event frequency while approximately preserving the original distribution conditional on the event to keep the in-event importance weights stable.

In the classical IS literature, effective proposals are often derived by exploiting analytical structure in the underlying stochastic system~\cite{glasserman2004monte,bucklew2004introduction}. For LMs, this approach is considerably more difficult. The proposal must modify a sequence of context-dependent conditional distributions in a coordinated manner, while preserving sufficient probability mass across the many distinct trajectories that can realize the event. The exponentially large trajectory space and the multiplicative accumulation of token-level likelihood ratios make this coverage requirement especially challenging. Our goal is therefore to develop a tractable procedure for constructing good autoregressive proposals.

\paragraph{Weight-space perturbations as a differentiable family of proposals}
Our central observation is that for autoregressive LMs, proposal distributions need not be specified explicitly over complete token sequences. Instead, they can be defined implicitly through the parameters of the generative model. For the base model $p$ with weights $\theta$, any perturbation $\delta$ defines a proposal $q_\delta := p_{\theta+\delta}$. Because the same model parameters determine the conditional distribution at every decoding step, a single perturbation coherently changes the entire collection of prefix-dependent conditionals and thereby induces a new distribution over complete trajectories. This parameterization has an important computational advantage. The mapping $\delta \mapsto q_\delta$ is differentiable, so proposal design becomes a gradient-based optimization problem in the LM’s parameter space rather than an explicit search over distributions on $\mathcal{V}^L$. By contrast, the CE optimization and twisted-SMC variants evaluated here use elite selection and particle resampling, respectively, without gradients through the LM's weights (Appendices~\ref{app:relationship-ce} and~\ref{app:relationship-tsmc}).

The tilting can be implemented at three levels of the generation process.
\begin{enumerate}
    \item \emph{Logit-level tilting}. Add a trainable $|\mathcal{V}|$-dimensional vector to the output logits. %
    \item \emph{Activation-level tilting}. Add trainable offsets to the model's hidden activations while keeping its weights fixed.
    \item \emph{Weight-space tilting}. Directly perturb the model weights using methods such as LoRA~\cite{hu2022lora}.
\end{enumerate}
These choices differ in expressive power. Activation-level tilting can change hidden representations throughout the model, but additive offsets offer less flexibility than modifying the model's transformations through full-parameter updates or LoRA. We focus on weight-space tilting and evaluate two questions separately. First, we compare weight-space tilting with activation- and logit-level tilting to assess the value of proposal expressivity. Second, within the activation and logit parameterizations, we compare gradient-based optimization with CE optimization to assess the value of the search algorithm.

\paragraph{Interpretation of importance weights.}
Under this class of proposals, the trajectory-level weight factorizes into the following product of per-step likelihood ratios along the realized sequence.
\begin{equation}
w(\mathbf{y}) \;=\; \frac{p(\mathbf{y})}{q_\delta(\mathbf{y})} \;=\; \prod_{\ell=1}^{L} \frac{p(y_\ell \mid \mathbf{y}_{<\ell})}{q_\delta(y_\ell \mid \mathbf{y}_{<\ell})}.
\end{equation}
Thus, discrepancies between the base and proposal conditionals accumulate across decoding steps. Severe under-coverage at a single step, or moderate under-coverage across several steps, can produce a large trajectory-level weight. Proposal design must therefore coordinate probability shifts across the sequence.
This adds constraints to the proposal search process that must be carefully managed. It is not enough for the proposal to simply trigger the rare event via arbitrary, out-of-distribution paths. If the proposal learns to satisfy the rare-event condition using sequences that are highly unnatural to the base model, it can under-cover the natural, high-probability trajectories that the base model would actually take to reach that same event. This failure to cover the base model's preferred paths leads directly to exploding importance weights. Therefore, the proposal must be constrained to emit the rare event in a way that is \textit{natural} under the base model, discovering and amplifying the exact paths to the rare event that the base model itself most prefers.

\section{Iterative Unalignment}
\label{sec:iu}

We now have a class of proposals $q_\delta := p_{\theta+\delta}$ that tilt the base model by changing its weights. Let $r_\delta=q_\delta(\mathcal{E})>0$ be the proposal's event probability, $q_\delta^{\mathcal{E}}=q_\delta(\cdot\mid\mathcal{E})$ be its conditional distribution within the event, and $q^*=p(\cdot\mid\mathcal{E})$ be the base model conditioned on the event. The variance of the $N$-sample importance-sampling estimator decomposes as
\begin{equation}
\label{eq:is_variance_decomposition}
\mathrm{Var}(\widehat{\mu})
=\frac{1}{N}\mathrm{Var}_{q_\delta}\!\left[w(\mathbf{y})\Phi(\mathbf{y})\right]
=\frac{\mu^2}{N}\left[
\underbrace{\frac{1-r_\delta}{r_\delta}}_{\text{event scarcity}}
+
\underbrace{\frac{\chi^2(q^*\Vert q_\delta^{\mathcal{E}})}{r_\delta}}_{\text{weight dispersion}}
\right],
\end{equation}
where $\chi^2(q^*\Vert q_\delta^{\mathcal{E}})=\mathbb{E}_{q_\delta^{\mathcal{E}}}[(q^*/q_\delta^{\mathcal{E}}-1)^2]$. Thus, a good proposal increases $r_{\delta}$ while keeping $\chi^2(q^*\Vert q_\delta^{\mathcal{E}})$ small. The ideal proposal $q^*(\mathbf{y})=p(\mathbf{y})\Phi(\mathbf{y})/\mu$ amplifies the base-model probability of every event trajectory by the same factor $1/\mu$, preserving their relative probabilities. It achieves zero variance because it places all mass on the event and matches $p(\cdot \mid \mathcal{E})$'s relative mass exactly, making both contributing terms vanish. Note that while $q^*$ is an informative construction, it is inaccessible as it depends on the unknown probability $\mu$.

Obtaining an optimization signal for the event and controlling distortion within the event region are again rare-event problems in their own right. The event indicator provides almost no optimization signal for increasing $r_\delta$ when event hits are scarce. Measuring and controlling $\chi^2(q^*\Vert q_\delta^{\mathcal{E}})$ is also difficult because it compares conditional distributions within the rare-event region over all rare-event sequences.

To amplify the event, we introduce a dense surrogate $S_\delta(\mathbf{y})$ based on differentiable components of the proposal model, such as logits evaluated along the prefixes of $\mathbf{y}$. To restrain within-event distortion, we introduce a tractable regularizer $\mathcal{R}_\delta(\mathbf{y})$ evaluated on proposal trajectories. At practical training budgets, sequences sampled from $p$ typically contain no event hits and therefore provide no direct observations of within-event divergence. Sampling from the proposal makes this region more observable as amplification proceeds. We apply the regularizer to all proposal trajectories, including those outside the event, to provide feedback even before event occurrences become frequent.

We propose a procedure called \emph{Iterative Unalignment} (IU). At iteration $t$, we draw a fresh batch from the current proposal $q_{\delta_t}$ and form the empirical objective
\begin{equation}
\label{eq:iu_objective}
\mathcal{J}_t(\delta)
=
\frac{1}{B}\sum_{i=1}^B
\left[
\underbrace{S_\delta(\mathbf{y}_i)}_{\text{event scarcity}}
-
\lambda_t\underbrace{\mathcal{R}_\delta(\mathbf{y}_i)}_{\text{weight dispersion}}
\right],
\qquad \mathbf{y}_i\simiid q_{\delta_t}.
\end{equation}
where $\lambda_t>0$ trades off the two terms. We update the proposal using $\left.\nabla_\delta\mathcal J_t(\delta)\right|_{\delta=\delta_t}$, holding the sampled trajectories fixed.

Learning an effective proposal requires discovering event trajectories while preserving their relative probabilities under the base model. These requirements are closely connected because each update changes the trajectories available to guide subsequent updates. Before event hits become frequent, amplification relies on the surrogate's model-based signals. Optimizing these signals can favor a small subset of event trajectories and reduce the probability of others, including paths that have not yet been observed. Those paths then become harder to discover, and their contribution to importance-weight variability can remain hidden from the training batches. The regularizer must therefore restrain concentration even when direct evidence of within-event distortion is sparse.
We develop the surrogate and regularizer in \S\ref{sec:surrogate} and \S\ref{sec:reg-term}, respectively. We calibrate their relative strength using the importance weights observed during training, first across the full batch and then within the event once hits become frequent enough (\S\ref{sec:adaptive_reg}). These observed weights cannot certify coverage of unobserved event paths.

Since token generation is discrete, we treat the sampled trajectory $\mathbf y$ as fixed when computing an IU update. We differentiate $S_\delta(\mathbf y)$ and $\mathcal{R}_\delta(\mathbf y)$ only through the proposal model's conditional distributions $q_\delta(\cdot\mid\mathbf{y}_{<\ell})$ evaluated along the realized prefixes. Thus, IU backpropagates through model evaluations on the sampled trajectory, but not through the discrete sampling operation that produced the trajectory. Our derivative-free CE optimization baselines provide an alternative that does not use these model gradients.

\subsection{Differentiable surrogates}
\label{sec:surrogate}

We first address the event-scarcity term in Eq.~\eqref{eq:is_variance_decomposition}, which decreases as the proposal event probability $q_\delta(\mathcal{E})$ increases. Directly maximizing $q_\delta(\mathcal{E})=\mathbb{E}_{q_\delta}[\Phi(\mathbf{y})]$ provides little usable signal in the rare-event regime because almost every sampled trajectory has $\Phi(\mathbf{y})=0$, and the discrete sampled tokens do not admit an ordinary pathwise gradient. We therefore retain the binary indicator $\Phi$ as the sole definition of the event and introduce a separate differentiable surrogate $S_\delta$ to guide proposal optimization. Unlike traditional formulations in which a continuous score both defines and locates the event, this separation allows the event definition itself to remain arbitrary and binary.

Since the effectiveness of a surrogate depends both on the event and the optimization trajectory, our goal in this section is to identify the practical properties that make a surrogate useful and demonstrate that such surrogates may realistically exist for many rare events.

\begin{figure}[t]
    \centering
    \includegraphics[width=\linewidth]{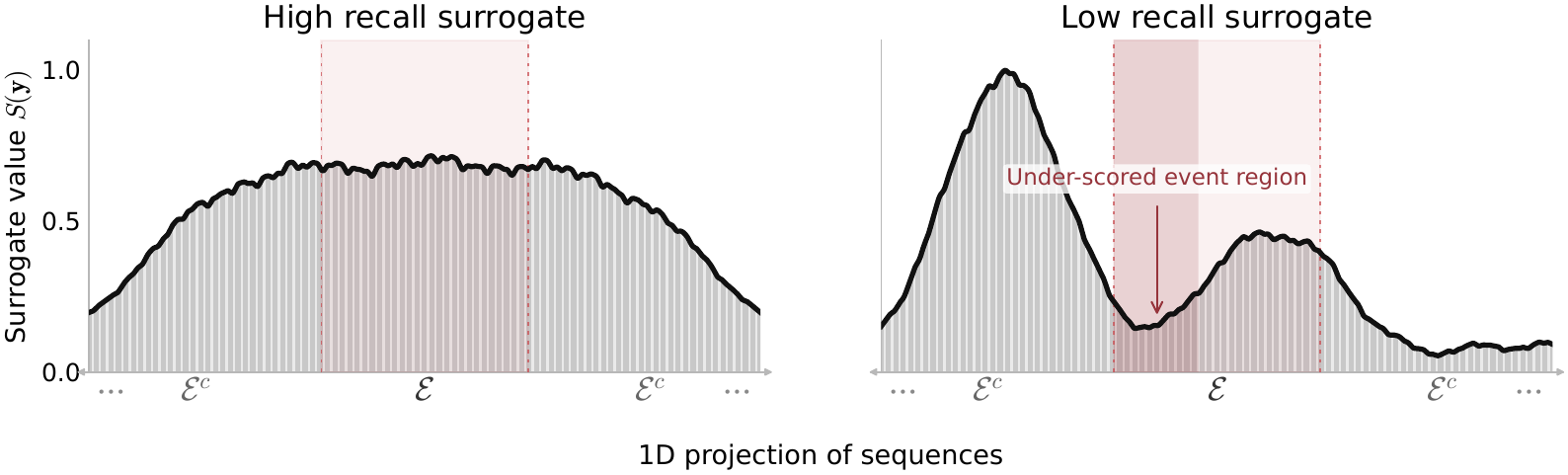}
    \caption{\textbf{Conceptual illustration of recall.} Surrogate optimization seeks to amplify the region where surrogate values are high, corresponding to an upper level set. Ideally, this region contains the entire event $\mathcal{E}$, even if it also includes non-event trajectories. Left: the high-recall surrogate scores the whole event highly. Right: the low-recall surrogate leaves part of the event outside its preferred region (dark-red). The pale red band marks $\mathcal{E}$. Thin gray bars denote individual sequence scores in a one-dimensional projection of discrete sequence space; black curves guide the eye.}
    \label{fig:surrogate-recall}
\end{figure}

\paragraph{Three desiderata for a useful surrogate.}
A useful surrogate must possess three key properties to effectively guide the proposal distribution.
\begin{enumerate}
    \item \textbf{Event observability.} The surrogate $S_\delta$ should provide informative values on trajectories for which the rare event does not occur. In the rare-event regime, $\Phi(\mathbf{y})=0$ for nearly all sampled trajectories and therefore provides little guidance for proposal optimization. In contrast, $S_\delta$ can be constructed from differentiable model quantities such as token probabilities. It can therefore supply a dense optimization signal before the event is observed.
    \item \textbf{Event alignment.} Increasing the surrogate should tend to increase the probability of the true event. Informally, the gradient induced by $S_\delta$, $\mathbb{E}_{q_{\delta}}[\nabla_{\delta} S_\delta(\mathbf{y})]$, should point in a direction similar to $\nabla_{\delta} \mathbb{E}_{q_{\delta}}[\Phi(\mathbf{y})]$. The surrogate should provide a useful direction for steering the proposal toward the event.
    
    \item \textbf{Event recall.} Optimization favors regions where the surrogate takes high values, corresponding to its upper level sets. It seeks to amplify proposal probability in these regions. \emph{Recall} asks how much of the event $\mathcal{E}$ this preferred region contains, while \emph{precision} asks how much of the preferred region lies in $\mathcal{E}$. Ideally, the preferred region contains the entire event. To achieve this the surrogate should assign comparably high values throughout $\mathcal{E}$ (much like $\Phi$) within the event, since otherwise optimization may only prefer a subset of the event. Whether training achieves this coverage also depends on the proposal family, regularizer, and optimizer. Figure~\ref{fig:surrogate-recall} illustrates recall. Low precision causes the proposal to spend samples on trajectories that the exact indicator $\Phi$ later sets to zero. Low recall may cause the optimization process to assign too little proposal probability to a set of important event trajectories, making $q_\delta(\mathbf{y}) \ll p(\mathbf{y})$ and the importance weight $p(\mathbf{y})/q_\delta(\mathbf{y})$ extremely large. Thus, false positives mainly reduce event observability, whereas false negatives can produce the under-coverage and highly variable importance weights described in \S\ref{section:problem_formulation}.
    
    For our purposes, only high recall is required, whereas high precision is optional. The region favored by optimization should retain the important event trajectories, even if it includes false positives. Low recall is more harmful because the surrogate is the sole mechanism that guides the proposal toward the event. The regularizer introduced in \S\ref{sec:reg-term} can control importance weight dispersion, but it cannot recover important event trajectories that the surrogate systematically neglects.
\end{enumerate}

\paragraph{Potential surrogates for complex rare events.}

Suppose the event is that an autonomous agent successfully gains elevated permissions \emph{and} issues a destructive \texttt{delete} command. Constructing a differentiable surrogate for the complete multi-stage event may be difficult. Issuing the \texttt{delete} command is a necessary condition. A surrogate that scores trajectories containing this command highly can favor the full event, along with attempts that lack the required permissions. It therefore has imperfect precision.

More generally, a differentiable surrogate for a necessary condition can be useful even when its preferred region extends beyond $\mathcal{E}$. If the event is much more common in this region than under the base model, shifting proposal mass there can make the event more observable, provided its important trajectories retain coverage. Preliminary experiments suggest that IU can tolerate this broader surrogate coverage, though estimation error increases as the mismatch grows (Appendix~\ref{app:surrogate-coverage}).

\subsection{Controlling within-event distortion via regularization}
\label{sec:reg-term}

The regularizer $\mathcal{R}_\delta$ is intended to control $\chi^2(q^*\Vert q_\delta^{\mathcal{E}})$, the within-event distortion in Eq.~\eqref{eq:is_variance_decomposition}, while allowing the surrogate to amplify the event. This requires assessing how the proposal's conditional distribution within $\mathcal{E}$ differs from $q^*$.

Recall that we evaluate $\mathcal{R}_\delta$ using samples from $q_\delta$. Samples from the base model $p$ rarely visit $\mathcal{E}$, so evaluating the regularizer on them mainly constrains the proposal on typical sequences outside the event. As amplification raises $r_\delta$, proposal samples visit the event more often and provide more direct feedback on importance weights within it. Appendix~\ref{app:regularizer-alternatives} compares joint forward KL on base-model rollouts, event-gated regularization, and on-policy forward KL.

Written as an expectation under the proposal $q_\delta$, the chi-square divergence is:
\begin{equation}
\label{eq:event-chi-square-proposal-expectation}
\chi^2(q^*\Vert q_\delta^{\mathcal{E}})
=\frac{r_\delta}{\mu^2}\,
\mathbb{E}_{\mathbf{y}\sim q_\delta}
\left[\Phi(\mathbf{y})\left(\frac{p(\mathbf{y})}{q_\delta(\mathbf{y})}\right)^2\right]-1.
\end{equation}
Although controlling this divergence directly controls the within-event contribution to variance, estimating it from a finite batch can give a misleading picture of coverage. Even if the event weights in a finite batch show little dispersion, other event paths carrying substantial mass under $q^*$ may be severely under-covered. The variance benefit depends on preserving the base model's relative probability mass across the entire event region, while a finite batch reveals only a small subset. Agreement on the observed paths therefore does not establish that $\chi^2(q^*\Vert q_\delta^{\mathcal{E}})$ is small.

This coverage problem is compounded by sampling from the proposal itself. Paths with substantial mass under $q_\delta$ are sampled frequently, while event trajectories can remain unobserved despite their large importance weights $p(\mathbf y)/q_\delta(\mathbf y)$. These large weights can make such trajectories major contributors to the squared-weight expectation in ~\eqref{eq:event-chi-square-proposal-expectation}. Consequently, proposal batches may substantially underestimate the divergence by missing its largest contributors. The population identity remains exact, but estimating it for regularization can be least reliable where the proposal's coverage is worst.

\begin{figure}[t]
    \centering
    \includegraphics[width=\linewidth]{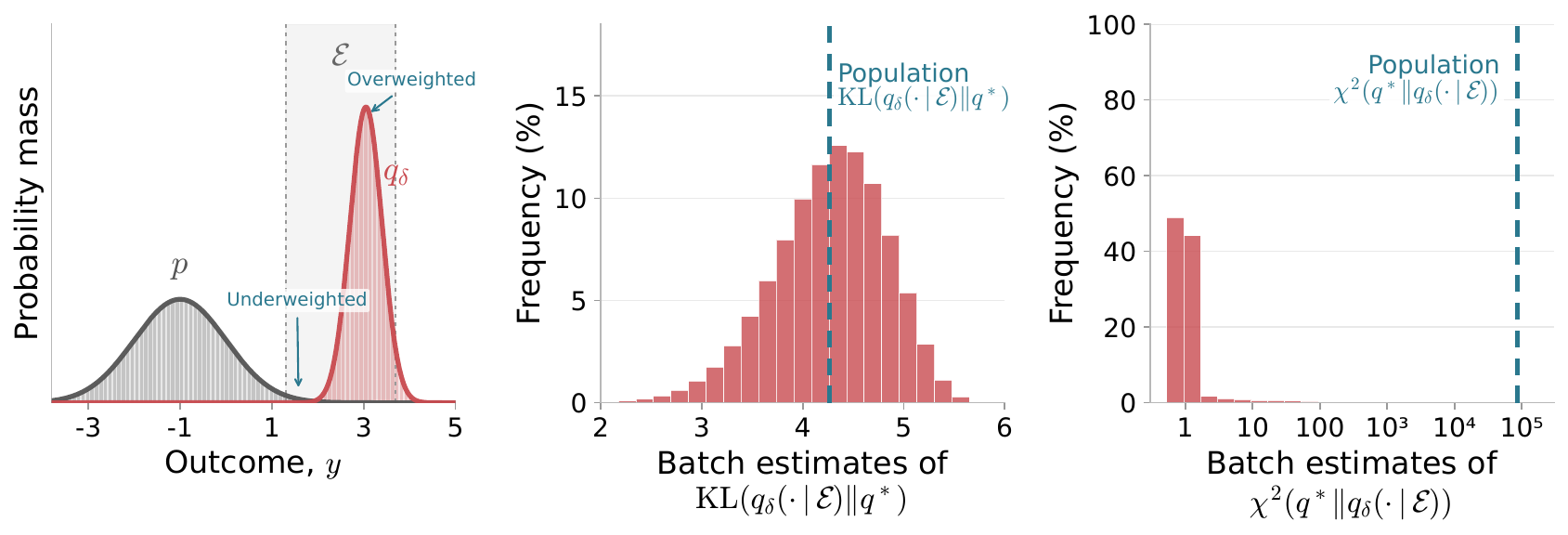}
    \caption{Finite-batch measurement of event-conditional divergences in a discrete toy example. The proposal concentrates on one side of the event region, leaving the other side under-covered. Most proposal batches miss the under-covered region, so the chi-square estimates (right panel) severely underestimate the population value. In contrast, the reverse-KL estimates (center panel) cluster near their population value.}
    \label{fig:kl-direction-measurement}
\end{figure}

\paragraph{Controlling concentration.} To address this measurement difficulty, IU instead seeks to limit excessive concentration on the event paths that the surrogate is amplifying. At the level of the event-conditional distributions, this motivates the reverse divergence
\begin{equation}
\label{eq:event-reverse-kl}
\mathrm{KL}(q_\delta^{\mathcal{E}}\Vert q^*)
=\mathbb{E}_{\mathbf{y}\sim q_\delta^{\mathcal{E}}}
\left[\log\frac{q_\delta^{\mathcal{E}}(\mathbf{y})}{q^*(\mathbf{y})}\right].
\end{equation}
This direction penalizes disproportionate allocation of proposal mass to particular event paths. As those paths gain proposal mass, they also become more likely to appear in the samples used to measure the penalty. Its ideal preference remains $q^*$. Over unrestricted conditional distributions, both $\mathrm{KL}(q_\delta^{\mathcal{E}}\Vert q^*)$ and $\chi^2(q^*\Vert q_\delta^{\mathcal{E}})$ attain their unique minimum of zero at $q_\delta^{\mathcal{E}}=q^*$. Thus, at any fixed event probability $r_\delta$, the two criteria favor the same allocation within the event.

Figure~\ref{fig:kl-direction-measurement} gives a complementary finite-batch view. Although both $\mathrm{KL}(q_\delta^{\mathcal{E}}\Vert q^*)$ and $\chi^2(q^*\Vert q_\delta^{\mathcal{E}})$ are expectations under $q_\delta^{\mathcal{E}}$, they behave very differently empirically. Reverse KL is dominated by regions that receive substantial probability under the proposal and is therefore well represented by typical proposal batches. In contrast, the chi-square divergence places its largest contributions on trajectories that are underrepresented by $q_\delta$; these trajectories are precisely the least likely to appear in a finite batch. Consequently, empirical chi-square estimation becomes least reliable when proposal under-coverage is most severe.

This observation does not imply that reverse-KL regularization guarantees coverage of the entire rare-event region. Event modes systematically missed by the surrogate cannot be recovered by the regularizer alone. The surrogate identifies the region to amplify, while the reverse-KL penalty limits how aggressively probability can concentrate within that region.

\begin{figure}[t]
    \centering
    \includegraphics[width=0.85\linewidth]{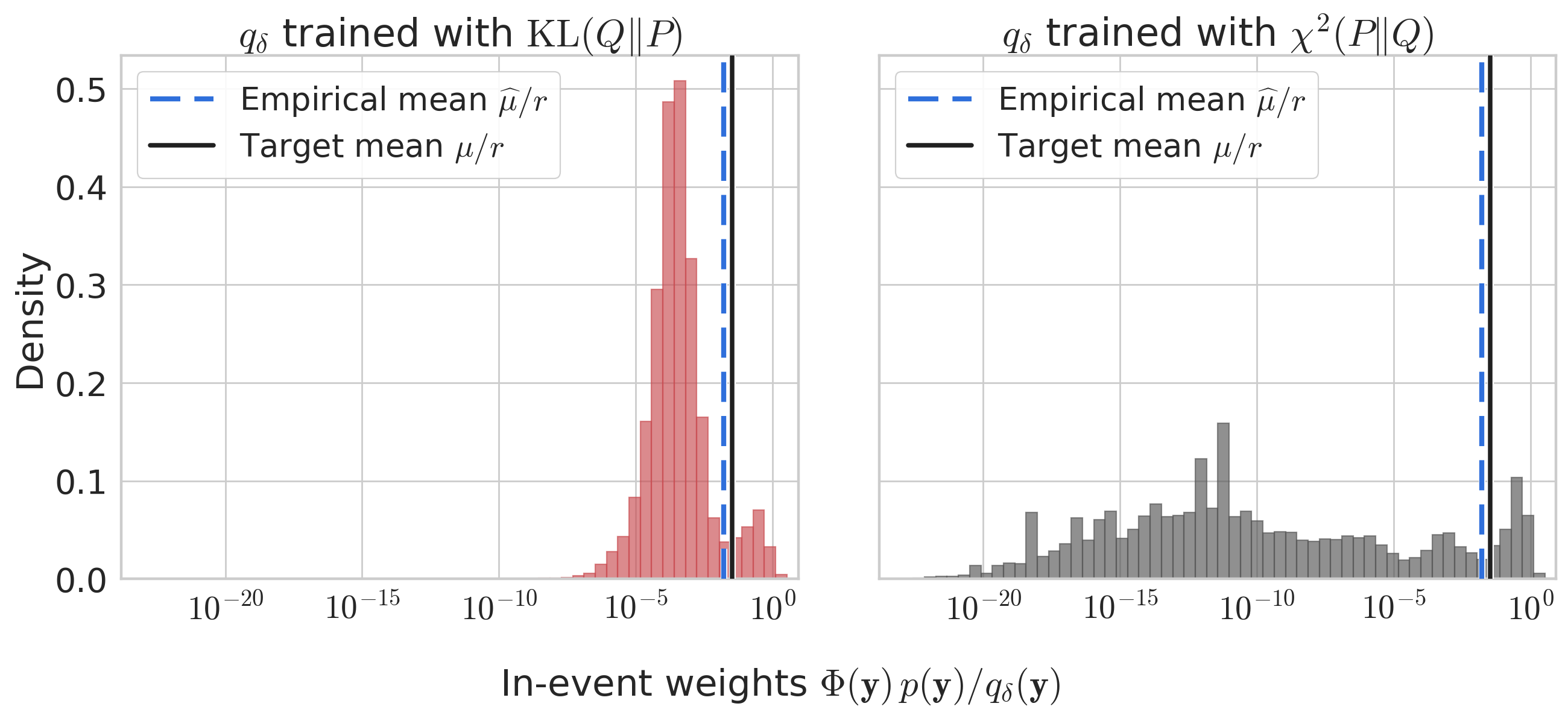}
    \caption{\textbf{Reverse-KL regularization produces more concentrated in-event weights in this comparison.} We show in-event importance weights under dense per-step reverse-KL (left) and chi-square (right) regularization. The chi-square weights span many more orders of magnitude. The solid black line marks the expected in-event weight $\mu/r_\delta$, and the dashed blue line marks the observed mean.}
    \label{fig:kl-chi-square-weights}
\end{figure}
\paragraph{Extending the regularizer's control.}
Even under proposal samples, event hits can still be scarce early in training. We therefore apply the penalty to all proposal trajectories by replacing the event-conditional divergence with $\mathrm{KL}(q_\delta\Vert p)$. Writing $q_\delta^{\mathcal{E}^c}=q_\delta(\cdot\mid\mathcal{E}^c)$ and $p_{\mathcal{E}^c}=p(\cdot\mid\mathcal{E}^c)$, we have
\begin{equation}
\label{eq:global-kl-event-decomposition}
\begin{aligned}
\mathrm{KL}(q_\delta\Vert p)
={}&r_\delta\log\frac{r_\delta}{\mu}
+(1-r_\delta)\log\frac{1-r_\delta}{1-\mu} \\
&+r_\delta\,\mathrm{KL}(q_\delta^{\mathcal{E}}\Vert q^*) \\
&+(1-r_\delta)\,\mathrm{KL}(q_\delta^{\mathcal{E}^c}\Vert p_{\mathcal{E}^c}).
\end{aligned}
\end{equation}
The joint divergence penalizes distortion both within and outside the event. Extending regularization to non-event trajectories constrains the proposal even when the training batch contains no event hits.

Estimating this penalty from trajectory log-likelihood ratios uses only the probabilities of tokens that happen to be sampled. At each visited prefix $\mathbf{y}_{<\ell}$, however, both models provide complete conditional distributions over the vocabulary. We use this information by computing their reverse KL at every decoding step and summing over the trajectory,
\begin{equation}
\label{eq:reg-final}
\begin{aligned}
\mathcal{R}_\delta(\mathbf{y})
&:=\sum_{\ell=1}^L
\mathrm{KL}\!\left(q_\delta(\cdot\mid\mathbf{y}_{<\ell})\Vert p(\cdot\mid\mathbf{y}_{<\ell})\right) \\
&=\sum_{\ell=1}^L\sum_{u\in\mathcal{V}}
q_\delta(u\mid\mathbf{y}_{<\ell})
\log\frac{q_\delta(u\mid\mathbf{y}_{<\ell})}{p(u\mid\mathbf{y}_{<\ell})}.
\end{aligned}
\end{equation}
By the autoregressive chain rule,
$\mathbb{E}_{\mathbf{y}\sim q_\delta}[\mathcal{R}_\delta(\mathbf{y})]=\mathrm{KL}(q_\delta\Vert p)$.
Thus, each visited prefix supplies feedback on all possible next tokens without changing the population divergence. The prefixes themselves are still sampled from $q_\delta$.

We use Eq.~\eqref{eq:reg-final} in the objective~\eqref{eq:iu_objective}, holding the current proposal's sampled prefixes fixed during each update. We apply it to every rollout without gating by $\Phi$, so non-event completions also contribute their conditional distributions. Figure~\ref{fig:kl-chi-square-weights} compares dense per-step reverse-KL and chi-square regularization. For chi-square, we use the dense per-step sum, which gave the best empirical performance among the chi-square implementations we tested (Appendix~\ref{app:per-step-chi-square}). In this comparison, reverse KL produces more concentrated in-event importance weights.

\subsection{Adaptive regularization}
\label{sec:adaptive_reg}
\begin{figure}[t!]
    \centering
    \includegraphics[width=0.85\linewidth]{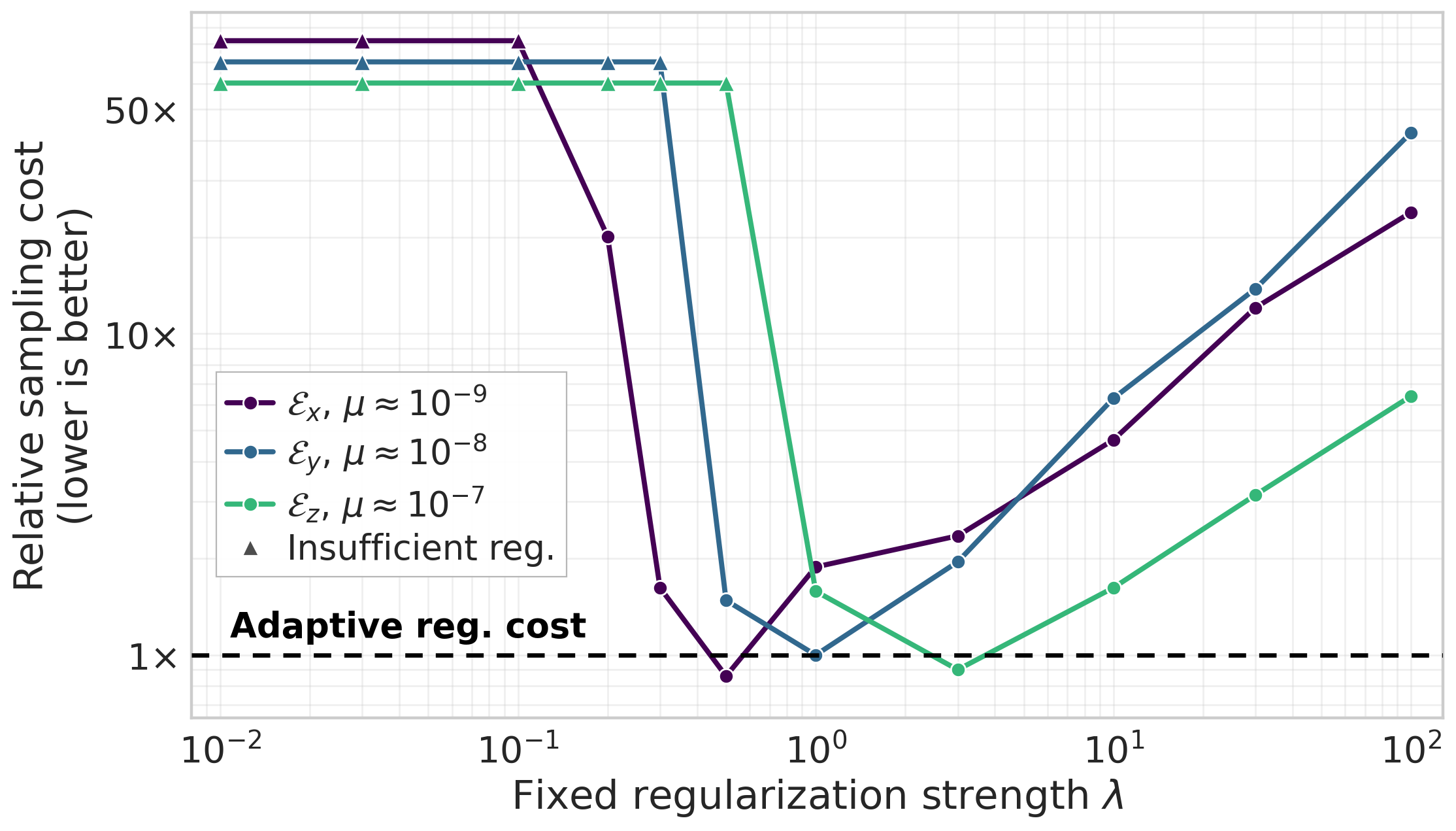}
    \caption{Fixed regularization fails at both extremes. With too little regularization, the proposal collapses onto the surrogate mode, driving the importance weights and probability estimates toward severe underestimation and therefore inflated sample cost. With too much regularization, weak amplification reduces efficiency and causes the sample cost to rise as well. Adaptive regularization avoids both regimes. Triangles mark costs above the plotting limit; their heights are capped for readability. Here sample cost is the number of trajectories required to achieve the same estimation accuracy as adaptive regularization (Eq.~\eqref{eq:required-budget}).}
    \label{fig:adaptive-lambda}
\end{figure}

In this section, we discuss how to set the strength of the regularizer in IU, i.e., $\lambda_t$ in \eqref{eq:iu_objective}.
Too little regularization lets the proposal concentrate on a few event paths, while too much suppresses amplification and leaves many evaluation batches without an event hit (Figure~\ref{fig:adaptive-lambda}). The appropriate strength depends on the event's rarity, the sequence length, and the proposal family, and changes as training proceeds. We therefore adjust $\lambda_t$ in Eq.~\eqref{eq:iu_objective} using the observed importance weights.

For a batch $\{\mathbf{y}_i\}_{i=1}^B\sim q_{\delta_t}$, let $w_i=p(\mathbf{y}_i)/q_{\delta_t}(\mathbf{y}_i)$ and $\Phi_i=\Phi(\mathbf{y}_i)$. We define the empirical full-batch and in-event effective sample size ratios (ESSr) as
\begin{equation*}
\mathrm{ESSr}_{\mathrm{batch}}
=\frac{\bigl(\sum_{i=1}^B w_i\bigr)^2}{B\sum_{i=1}^B w_i^2},
\qquad
\mathrm{ESSr}_{\mathrm{event}}
=\frac{\bigl(\sum_{i=1}^B \Phi_i w_i\bigr)^2}
{\bigl(\sum_{i=1}^B \Phi_i\bigr)\bigl(\sum_{i=1}^B \Phi_i w_i^2\bigr)}.
\end{equation*}
Each ratio decreases as the observed weights become more dispersed relative to their mean. The in-event ratio is defined only when the batch contains event hits.

For a fixed proposal $q_\delta$ with the same support as $p$, the population full-batch ratio equals $\exp[-D_2(p\Vert q_\delta)]$, where $D_2$ is R\'enyi divergence of order two. This motivates limiting global departure from the base model during exploration, when event hits are scarce. Once events become sufficiently frequent, we use the in-event ratio, whose population value is $1/[1+\chi^2(q^*\Vert q_\delta^{\mathcal{E}})]=\exp[-D_2(q^*\Vert q_\delta^{\mathcal{E}})]$. The common scale factor $\mu/r_\delta$ in the event weights cancels from this ratio, so it isolates within-event distortion from common event amplification.

\begin{figure}[t!]
    \centering
    \includegraphics[width=\linewidth]{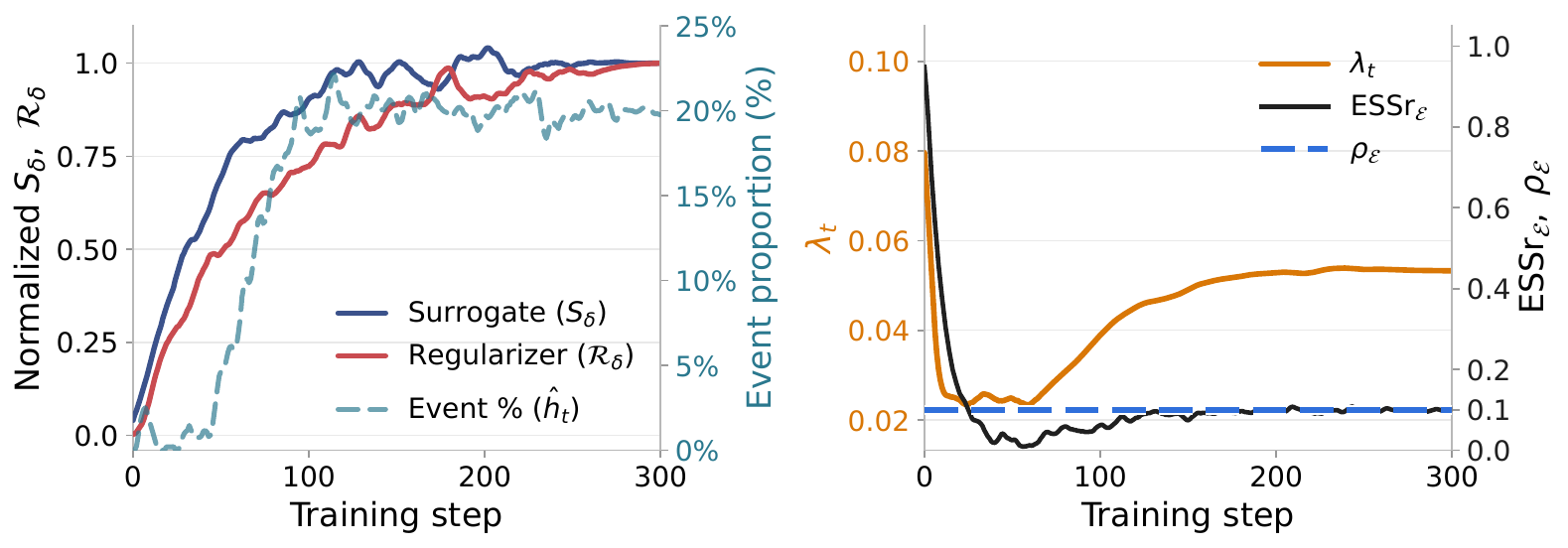}
    \caption{IU training dynamics over 300 optimization steps. \textbf{Left panel.} The empirical event proportion $\hat h_t$ (in dashed teal) rises during training. The surrogate $S_\delta$ (blue) and regularizer $\mathcal{R}_\delta$ (red) are each normalized by their terminal value for plotting. \textbf{Right panel.} Adaptive regularization brings the event-specific $\mathrm{ESSr}_{\mathcal{E}}$ (black) to its target $\rho_{\mathcal{E}}$ (dashed blue) by adjusting the multiplier $\lambda_t$ (orange).}
    \label{fig:training-dynamics}
\end{figure}

Let $\hat h_t=B^{-1}\sum_{i=1}^B\Phi_i$ be the empirical event frequency. At each iteration, we select the empirical ratio and its target using
\begin{equation*}
(\mathrm{ESSr}_t,\rho_t)
=
\begin{cases}
(\mathrm{ESSr}_{\mathrm{batch}},\rho_{\mathrm{batch}}), & \hat h_t<h^\star,\\
(\mathrm{ESSr}_{\mathrm{event}},\rho_{\mathrm{event}}), & \hat h_t\geq h^\star.
\end{cases}
\end{equation*}
We set $h^\star=0.1$ in our experiments and typically choose $\rho_{\mathrm{event}}>\rho_{\mathrm{batch}}$ to target less relative weight dispersion within the event. We initialize $\lambda_0\geq\lambda_{\min}>0$ at a value suited to the scale of the regularizer and adapt it using
\begin{equation}
\lambda_{t+1}=\max\!\left\{\lambda_{\min},\,
\lambda_t \exp\!\bigl(\eta_\lambda(\rho_t-\mathrm{ESSr}_t)\bigr)\right\},
\end{equation}
where $\eta_\lambda>0$ is a step size. If the active empirical ESSr falls below its target, $\lambda_t$ increases; if it exceeds the target, $\lambda_t$ decreases toward its floor. Figure~\ref{fig:training-dynamics} illustrates this feedback during training.

These population identities do not make empirical ESSr a certificate of population ESSr or a bound on chi-square divergence. Unobserved event modes can make $\chi^2(q^*\Vert q_\delta^{\mathcal{E}})$ arbitrarily large while empirical ESSr remains high, so reliable measurement faces the same finite-sample limitations discussed in \S\ref{sec:reg-term}. We use empirical ESSr as coarse feedback about observed weight dispersion to tune $\lambda_t$. The dense reverse-KL term in Eq.~\eqref{eq:reg-final} supplies the optimization gradient across the vocabulary at every visited prefix, while ESSr adjusts how strongly it is applied. This lets observed weight dispersion guide regularization strength without requiring a reliable divergence estimate as the optimization objective. The indicator $\Phi$ selects weights for in-event feedback, while the penalty continues to apply to every rollout.

\subsection{Putting it all together}
\label{sec:putting_it_all_together}

Combining the surrogate from Section~\ref{sec:surrogate} with the regularizer from Section~\ref{sec:reg-term}, IU updates the proposal parameters by maximizing the objective $\mathcal{J}_t$ in Eq.~\eqref{eq:iu_objective}.
The regularization strength $\lambda$ is adjusted online according to the adaptive rule in \S\ref{sec:adaptive_reg}. Because the training trajectories are sampled from the current proposal, the updates are concentrated on the regions of the trajectory space that the proposal actively explores. After training, we draw a fresh set of trajectories from the learned proposal $q_\delta$ and estimate $\mu$ using exact importance weights.

Algorithm~\ref{alg:iu} summarizes the end-to-end procedure, comprising gradient updates on $\delta$, adaptive updates on $\lambda$, and a final importance-sampling stage under the learned proposal after training.

\begin{algorithm}[!htb]
\caption{Iterative Unalignment (IU)}
\label{alg:iu}
\begin{algorithmic}[1]
\setlength{\itemsep}{2pt}
\Statex \textbf{Inputs:} $p,\Phi,S_\delta,\mathcal{R}_\delta$; $T,B,N$; $\eta_\delta,\eta_\lambda$; $\rho_{\mathrm{batch}},\rho_{\mathrm{event}}$; $h^\star\in(0,1)$; $\lambda_0\geq\lambda_{\min}>0$.
\State $\delta_1\leftarrow\mathbf{0}$, $\lambda_1\leftarrow\lambda_0$
\For{$t=1,2,\dots,T$}
    \State Sample $\mathbf{y}_1,\ldots,\mathbf{y}_B\simiid q_{\delta_t}$ \Comment{Fixed sampled trajectories}
    \State $w_i\leftarrow p(\mathbf{y}_i)/q_{\delta_t}(\mathbf{y}_i)$, $\Phi_i\leftarrow\Phi(\mathbf{y}_i)$ for $i=1,\ldots,B$
    \State $\mathcal{J}_t(\delta)\leftarrow\frac{1}{B}\sum_{i=1}^B\bigl[S_\delta(\mathbf{y}_i)-\lambda_t\mathcal{R}_\delta(\mathbf{y}_i)\bigr]$ \Comment{Eq.~\eqref{eq:iu_objective}}
    \State $\delta_{t+1}\leftarrow\delta_t+\eta_\delta\nabla_\delta\mathcal{J}_t(\delta)\big|_{\delta=\delta_t}$ \Comment{Update proposal}
    \State $\hat h_t\leftarrow B^{-1}\sum_{i=1}^B\Phi_i$ \Comment{Update event proportion}
    \State $(\mathrm{ESSr}_t,\rho_t)\leftarrow
    \begin{cases}
        (\mathrm{ESSr}_{\mathrm{batch}},\rho_{\mathrm{batch}}), & \hat h_t<h^\star,\\
        (\mathrm{ESSr}_{\mathrm{event}},\rho_{\mathrm{event}}), & \hat h_t\geq h^\star.
    \end{cases}$
    \State $\lambda_{t+1}\leftarrow\max\!\left\{\lambda_{\min},\lambda_t\exp\!\bigl[\eta_\lambda(\rho_t-\mathrm{ESSr}_t)\bigr]\right\}$ \Comment{Adapt regularization}
\EndFor
\State Sample $\mathbf{y}_1,\ldots,\mathbf{y}_N\simiid q_{\delta_{T+1}}$ \Comment{Fresh trajectories for estimation}
\State $\widehat{\mu}\leftarrow\frac{1}{N}\sum_{j=1}^N\frac{p(\mathbf{y}_j)}{q_{\delta_{T+1}}(\mathbf{y}_j)}\Phi(\mathbf{y}_j)$ \Comment{IS estimator}
\State \Return $(\widehat{\mu},q_{\delta_{T+1}})$
\end{algorithmic}
\end{algorithm}

%% file: experimental_setup.tex
\section{Experimental setup}
\label{sec:experimental_setup_section}

\subsection{Evaluation settings}
\label{section:eval_settings}

Evaluating a rare-event estimator requires a reference ground truth value of $\mu$ to validate against. However, the very scarcity that motivates rare-event estimation makes obtaining reference values through naive Monte Carlo computationally prohibitive. Attaining a 10\% relative standard error for an event at $\mu = 10^{-9}$ requires approximately $10^{11}$ raw-indicator rollouts via naive Monte Carlo (see Appendix~\ref{sec:surrogate_proofs}). Angell et al.~\cite{angell2026estimating} compare against Monte Carlo references with a minimum nonzero value of $10^{-4}$, while Dorman et al.~\cite{dorman2026rare} focus on continuous scalar scores where independent deep-tail reference probabilities are unavailable.

A good evaluation set should be both realistic, in the sense of faithfully representing deployment failures, and tractable enough to admit trustworthy reference probabilities. For events defined by an LM autorater's decision boundary, precise deep-tail reference probabilities require prohibitive amounts of compute to accurately obtain~\cite{angell2026estimating}, while other evaluations target continuous heuristic scores such as readability and output log probability~\cite{dorman2026rare}. Constructing events that are simultaneously realistic, rare, and verifiable is an entire research topic of its own. As our contribution is primarily algorithmic, we focus on events that preserve the difficulty of proposal design while allowing estimator accuracy to be verified.

We use Token Presence for systematic evaluation and two complementary case studies to evaluate IU (Table~\ref{tab:evaluation-settings}). Token Presence allows ground truth to be computed efficiently in parallel across batches for the entire vocabulary, enabling systematic sweeps over deep event rarities (e.g., $\mu = 10^{-9}$) and longer generation lengths (up to 50 tokens). Compositional Token Presence extends this setting to multiple target tokens, either in a specified order (Ordered Sequence events) or in any order (Conjunctive events). Our Profanity Classifier setting connects the evaluation to the traditional formulation in which an event is defined by a continuous score crossing a threshold.

\begin{table}[ht]
    \centering
    \small
    \begin{tabular}{@{} >{\raggedright\arraybackslash}p{0.27\columnwidth} >{\raggedright\arraybackslash}p{0.65\columnwidth} @{}}
        \toprule
        \textbf{Evaluation setting} & \textbf{Motivation} \\
        \midrule
        \textbf{Token Presence} & Highly verifiable reference values across models, lengths, and rarities, while preserving complex, multi-path trajectory dynamics. \\
        \noalign{\vskip\aboverulesep
            {\color[gray]{0.75}\hrule height 0.3pt}
            \vskip\belowrulesep}
        \textbf{Compositional Token Presence} & Multiple target tokens in a specified order (Ordered Sequence events) or in any order (Conjunctive events). \\
        \noalign{\vskip\aboverulesep
            {\color[gray]{0.75}\hrule height 0.3pt}
            \vskip\belowrulesep}
        \textbf{Profanity Classifier} & Traditional score-threshold rare-event definition. For LMs, this is a continuous score defined over a sequence of tokens. \\
        \bottomrule
    \end{tabular}
    \caption{Proposed evaluation suite.}
    \label{tab:evaluation-settings}
\end{table}

We use informative surrogates to isolate the challenge of proposal learning. All three settings are sequence-level and retain the core challenge of conducting importance sampling in our setting. Every importance weight carries a product of $L$ step-wise likelihood ratios, so the challenges of the combinatorial $|\mathcal{V}|^L$ output space and the need for $q_\delta$ to remain disciplined across the full trajectory remain present.

\subsubsection{Token Presence events}
Let $\mathbf{y} = (y_1, \ldots, y_L) \in \mathcal{V}^L$ be the output sequence. For a target token $v \in \mathcal{V}$, the Token Presence event is
\begin{equation}
\label{eqn:event}
    \Phi^{\mathrm{TOK}}(\mathbf{y};v) = \mathbb{I}\{\exists \,\,\ell \in\{1,\dots, L\} \mbox{ such that } y_\ell=v\}
\end{equation}
so that $\Phi^{\mathrm{TOK}}(\mathbf{y};v)=1$ if and only if $v$ appears at least once.

Token Presence defines a highly multimodal event region. A target token can appear after many different prefixes and at many different positions. Conjunctive and Ordered Sequence events introduce further combinations of prefixes and positions through which the event can occur. A successful proposal must therefore amplify the event without collapsing onto a small subset of these trajectories. Finding a single adversarial path is insufficient for estimating the event's total probability. For an event with base probability $\mu \approx 10^{-9}$, observing it reliably under a modest estimation budget (e.g., $N=512$) requires amplification by over $10^6$, while retaining coverage across the event region. Weight perturbations must therefore amplify the event while preserving coverage of its trajectories.

Token-level conditions also serve as natural proxies for real-world safety risks. For instance, Jones et al.~\cite{jones2025forecasting} use keyword presence in sampled responses as a proxy for harmful information. Their work's direct connection to deployment safety further motivates our use of Token Presence events to systematically verify our algorithm, without assigning each a safety meaning.

Token Presence also admits a low-variance, unbiased reference estimator constructed directly from the model's next-token probabilities. For a rollout $\mathbf{y}\sim p$, we replace the sparse binary indicator with a sum of conditional first-hit probabilities:

\begin{equation}
\label{eqn:unbiased_surrogate_set}
    \widehat{\Phi}^{\mathrm{TOK}}(\mathbf{y};v) = \sum_{\ell=1}^{L} p(v \mid \mathbf{y}_{<\ell}) \, \mathbb{I}\{v \notin \mathbf{y}_{<\ell}\}
\end{equation}

whose expectation is exactly $\mu^{\mathrm{TOK}}(v)=\mathbb{E}_p[\Phi^{\mathrm{TOK}}(\mathbf{y};v)]$; Appendix~\ref{sec:gt-unbiased} gives the derivation for this identity.

This estimator remains precise even when the token is almost never observed. The first-hit decomposition lets every rollout contribute a positive, unbiased estimate for every Token Presence event, whether or not the target token is sampled. Moreover, the next-token distributions recorded along a single rollout provide reference estimates for every token in the vocabulary simultaneously. For the rarest reported GPT-2 Small ($L=50$) and Gemma-2 ($L=10$) events, estimated reference relative standard errors are $0.74\%$ and $1.32\%$, respectively (Appendix~\ref{sec:soft-gt-cost}). Characterizing estimator spread across many Token Presence events requires substantial sampling, which limits model size and generation length (Section~\ref{experiments}).

Our reference computations and evaluated methods intentionally use different unbiased estimators. The reference estimators exploit additional event-specific structure to obtain precise ground truth. All evaluated methods estimate probabilities using binary event indicators, with importance weighting where applicable. For general events, binary event indicators paired with correlated surrogates are much more accessible than event-specific probability decompositions. We choose token-based events as a proxy for general rare events because their structure permits independent verification of reference probabilities down to $10^{-9}$ and reliable method comparisons.

To train the proposal, we use the target token's step-averaged log-probability as the differentiable surrogate in the IU objective~\eqref{eq:iu_objective}:
\begin{equation*}
S^{\mathrm{TOK}}_\delta(\mathbf{y}\,;\,v)
\;=\; \frac{1}{L}\sum_{\ell=1}^L
\log q_\delta\bigl(v \mid \mathbf{y}_{<\ell}\bigr).
\end{equation*}

\subsubsection{Compositional Token Presence events}
\label{sec:compositional-token-events}
As a case study, we consider two multi-token extensions of Token Presence: Ordered Sequence events and Conjunctive events. We refer to both as compositional events. Let $\mathbf{v}=(v_1,\ldots,v_m)$ be a tuple of target tokens.

\paragraph{Ordered Sequence (SEQ) events.} The Ordered Sequence event is
\begin{equation}
\label{eqn:sequence-event}
    \Phi^{\mathrm{SEQ}}(\mathbf{y};\mathbf{v})
    = \mathbb{I}\!\left\{\exists\, 1 \leq \ell_1 < \cdots < \ell_m \leq L \mbox{ such that } y_{\ell_j}=v_j \text{ for all } j\right\},
\end{equation}
which requires the targets to appear in order but not necessarily contiguously.
Let $c_\ell$ be the number of targets already completed in order by the prefix $\mathbf{y}_{<\ell}$.
Our surrogate rewards the next required token, advancing to the following target when that token appears:
\begin{equation*}
S^{\mathrm{SEQ}}_\delta(\mathbf{y}\,;\,\mathbf{v})
\;=\; \frac{1}{L}\sum_{\ell=1}^L
\mathbb{I}\{c_\ell<m\}
\log q_\delta\bigl(v_{c_\ell+1} \mid \mathbf{y}_{<\ell}\bigr).
\end{equation*}

\paragraph{Conjunctive (AND) events.} The Conjunctive event is
\begin{equation}
\label{eqn:and-event}
    \Phi^{\mathrm{AND}}(\mathbf{y};\mathbf{v})
    = \mathbb{I}\!\left\{\forall j \in \{1,\ldots,m\},\; \exists \ell \in \{1,\ldots,L\} \mbox{ such that } y_\ell=v_j\right\},
\end{equation}
which requires every target token to appear in any order.
The proposal must cover all target orderings, which makes it harder to learn.
We use a surrogate that raises the probability of every target token throughout the trajectory:
\begin{equation*}
S^{\mathrm{AND}}_\delta(\mathbf{y}\,;\,\mathbf{v})
\;=\; \frac{1}{L}\sum_{j=1}^m\sum_{\ell=1}^L
\log q_\delta\bigl(v_j \mid \mathbf{y}_{<\ell}\bigr).
\end{equation*}

Both compositional events admit unbiased reference estimators from the base model's next-token probabilities along ordinary rollouts. Even here, the single-token structure that lets every rollout contribute is unavailable, since prerequisite targets must first appear, making ground-truth computation harder. Appendix~\ref{sec:compositional-gt} defines these estimators and proves their unbiasedness, and Table~\ref{tab:reference-accuracy} reports reference probabilities and their estimated relative standard errors. We present a set of 5 events per setting, and defer a systematic study for future work.

\subsubsection{Profanity Classifier}

As our second case study, we estimate the probability that an LM generates text whose classifier probability exceeds a profanity threshold $\kappa\in(0,1)$, so that $\Phi^{\mathrm{PROF}}(\mathbf{y};\kappa) = \mathbb{I}\{f(\mathbf{y}) > \kappa\}$. This is an instance of the canonical threshold event of Section~\ref{section:problem_formulation} where a scalar score defines $\mathcal{E}$ by crossing a cutoff. Traditional rare-event methods often use the same score to specify the event and guide search, as in the score-based LM experiments of Dorman et al.~\cite{dorman2026rare}. Although we argue for separating the binary event $\Phi$ from the differentiable surrogate $S_\delta$ used to train the proposal, we include this setting to show that IU remains effective when the event takes the traditional score-threshold form.

The classifier $f$ is a bag-of-words logistic regression that assigns every vocabulary token $u \in \mathcal{V}$ a learned weight $\beta_u$. It converts the sum of token weights and a bias into a predicted probability:
\begin{equation*}
    f(\mathbf{y}) \;=\; \sigma\!\left(\sum_{\ell=1}^L \beta_{y_\ell} + b\right) \;=\; \sigma\!\left(\beta^\top \operatorname{count}(\mathbf{y}) + b\right),
\end{equation*}
where $\sigma(z)=(1+e^{-z})^{-1}$, $\beta \in \mathbb{R}^{|\mathcal{V}|}$ is the per-token weight vector, $b$ is a bias, and $\operatorname{count}(\mathbf{y}) \in \mathbb{Z}_{\ge 0}^{|\mathcal{V}|}$ is the token-count (bag-of-words) vector of $\mathbf{y}$. The event occurs when the predicted probability $f(\mathbf{y})$ exceeds $\kappa$.

For the Profanity Classifier, we estimate reference probabilities using the raw binary indicator. Appendix~\ref{sec:profanity-gt-cost} reports these estimates and their relative standard errors.

\paragraph{Differentiable surrogates for scores over discrete tokens.}
Although $f$ defines the event, it is evaluated on discrete realized tokens and is not differentiable with respect to the LM parameters $\delta$. We therefore use the differentiable surrogate below. We did not compare alternative surrogates in this setting.

Let $\mathbf{q}_\delta(\mathbf{y}_{<\ell})$ denote the vector of proposal next-token probabilities over $\mathcal{V}$ at prefix $\mathbf{y}_{<\ell}$. The surrogate we choose is made differentiable by replacing each realized token weight $\beta_{y_\ell}$ with its expectation under this distribution, $\beta^\top \mathbf{q}_\delta(\mathbf{y}_{<\ell})$. Summing these expected contributions gives a soft version of the sequence-level classifier logit. We subtract the corresponding logit threshold $\log[\kappa/(1-\kappa)]$ and apply a log-sigmoid to provide informative gradients below the threshold:
\begin{equation*}
S^{\mathrm{PROF}}_\delta(\mathbf{y}\,;\,\kappa)
\;=\;
\log \sigma\!\left(
\sum_{\ell=1}^L
\beta^\top \mathbf{q}_\delta(\mathbf{y}_{<\ell})
+ b - \log\frac{\kappa}{1-\kappa}
\right).
\end{equation*}

\subsection{Evaluation metrics}
\label{sec:evaluation_metrics}

We evaluate the spread of the estimator across repeated estimates.
For an event with reference probability $\mu>0$, we evaluate each learned
proposal using $K$ independent
estimates, $\{\widehat\mu_{k,N}\}_{k=1}^K$, each based on $N$ trajectories.
We use an evaluation budget of $64{,}000$ trajectories per estimator--event pair.

\paragraph{Limitations of empirical relative mean squared error.}
A standard measure of estimator accuracy is relative mean squared error (rMSE), which we estimate from $K$ independent runs as $\frac{1}{K} \sum_{k=1}^K (\widehat{\mu}_{k,N}/\mu - 1)^2$.
This empirical average, however, can be unstable under practical evaluation budgets. In our setting, importance weights are products of per-token likelihood ratios, which can yield highly skewed estimator distributions with rare, extreme overestimates even when most estimates are close to $\mu$. Squared relative error penalizes multiplicative overestimation and underestimation asymmetrically. Underestimates each contribute at most $1$ to the loss, since $0 \leq \widehat{\mu}_{k,N}/\mu \leq 1$ for any underestimate. By contrast, large overestimates incur quadratically growing penalties. For example, estimates of $10\mu$ and $100\mu$ incur losses of $81$ and $9801$, respectively, far exceeding the maximum underestimation loss of $1$. Consequently, population rMSE can be dominated by overestimates that occur too rarely to be reliably represented in $K$ evaluation runs. Empirical rMSE may substantially understate the population value when these extremes are absent, then increase sharply when a single extreme estimate appears. Although the empirical average is unbiased when population rMSE is finite, its variability can make comparisons between estimators unreliable. Prior work likewise shows that reliable variance-based comparisons of rare-event estimators can require prohibitively many simulation runs \cite{lecuyer2010robustness}. Figure~\ref{fig:winkler-intro} illustrates this failure mode; Figure~\ref{fig:rmse_pathology} in the appendix gives an empirical example. In our experience, stabilizing empirical rMSE typically requires an impractically large $K$.

\paragraph{Quantile log error.}
For rare-event probabilities spanning many orders of magnitude, a natural accuracy objective
is to recover the correct order of magnitude \cite{blanchet2011rare}.
A logarithmic scale expresses this directly. An error of one order of magnitude corresponds to a factor of ten at any
event probability, giving accuracy a common interpretation across rarity
levels. We therefore measure error as
$|\log_{10}(\widehat\mu_N/\mu)|$, following the use of log accuracy ratios
to measure errors in orders of magnitude
\cite{tofallis2015relative,morley2018logaccuracy}.
For example, estimates of $10\mu$ and $0.1\mu$ both incur a log error of $1$.
Thus, each additional factor of ten in either direction adds one unit of
error.

For equal additive deviations from $\mu$, log error penalizes
underestimation more heavily than overestimation.
For example, estimates of $0.1\mu$ and $1.9\mu$ both differ from $\mu$ by
$0.9\mu$, yet incur log errors of $1$ and approximately $0.28$, respectively.
This property is useful for our risk-assessment objective, since severely
understating a rare-event probability can create false reassurance.
We retain zero estimates and assign them the maximum error of $+\infty$,
since they represent a complete failure to observe the event.

We summarize log errors across repeated estimates using the \emph{quantile log error}, the $95$th percentile of absolute log error.
For example, a value of $1$ means that at least $95\%$ of estimates fall
within one order of magnitude of the true probability.
We define
\begin{equation}
\label{eq:direct-log-error}
E_{0.95}(N;\mu)
:=
Q_{0.95}\!\left(
\left|\log_{10}\frac{\widehat\mu_N}{\mu}\right|
\right),
\end{equation}
where $Q_{0.95}$ denotes the $95$th percentile across repeated runs. We estimate $E_{0.95}(N;\mu)$ using the empirical $95$th percentile of the
$K$ observed absolute log errors.

\begin{figure*}[!t]
    \centering
    \includegraphics[width=0.62\textwidth]{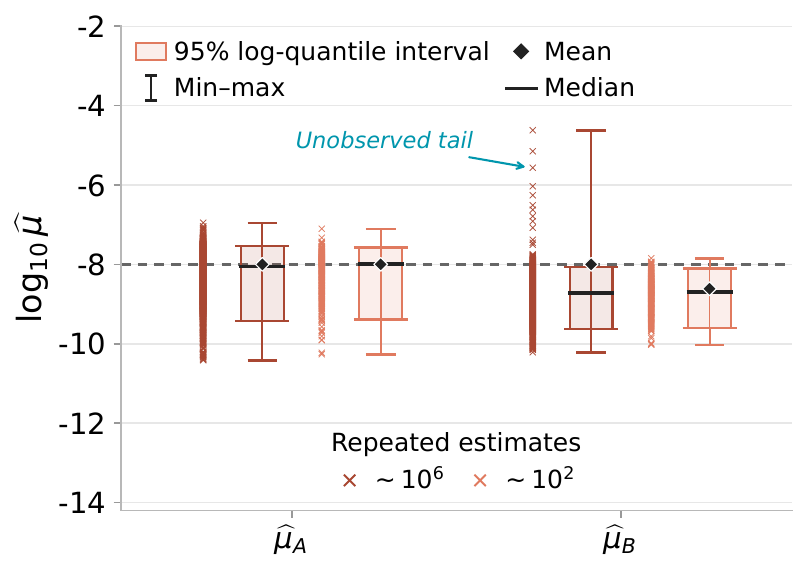}\hfill
    \includegraphics[width=0.37\textwidth]{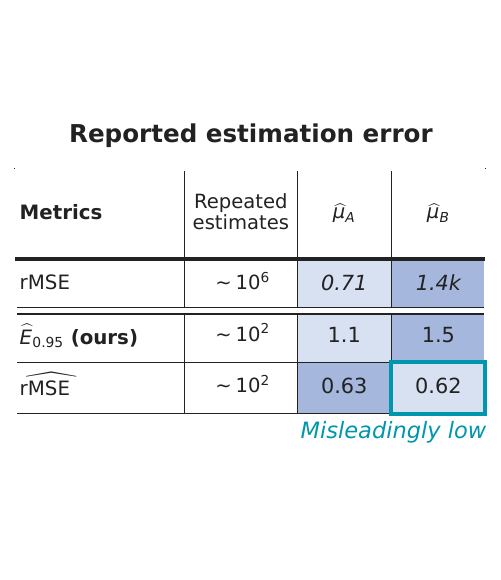}
    \caption{\textbf{Limited repetitions can miss the tails that dominate rMSE.} This schematic compares two estimators at a fixed sampling budget per estimate. The constructed reference of $10^6$ estimates represents a fuller evaluation of rMSE, including rare extremes that can dominate the loss in heavy-tailed regimes. With only $\sim10^2$ estimates, these extremes may go unobserved. A's rMSE changes little between the two samples, whereas B's drops sharply when its upper tail is missed, reversing their ranking. Quantile log error favors A in both samples, reflecting B's persistent underestimation. The dashed line marks the true probability; hats denote metrics computed from the smaller sample.}
    \label{fig:winkler-intro}
\end{figure*}

We use quantile log error to answer two questions.

\paragraph{How does the required sampling budget change as required estimation error tightens?}
For a chosen \emph{estimation error threshold} $E_{0.95}^\star>0$, we compare
the minimum sampling budget needed to meet it:
\begin{equation}
\label{eq:required-budget}
N^\star(E_{0.95}^\star;\mu)
:=
\min\left\{N\in\mathbb{N}_{>0}:
E_{0.95}(N;\mu)\leq E_{0.95}^\star\right\}.
\end{equation}
As an example, $E_{0.95}^\star=1$ requires at least $95\%$ of repeated
estimates to be within a factor of $10$ of $\mu$.

As the error threshold tightens, some methods do not meet the target within
our evaluation budget. We sketch the rest of the curve by first bootstrapping
to a larger budget, then adding simulated naive-MC samples if the target
remains unmet. We retain the initial IS estimates and count all IS and MC
samples toward the required budget. Appendix~\ref{app:accuracy-cost-estimation}
details the procedure and explains when these estimates are conservative.
In plots that average across events, the error-threshold axis stops at $0.5$
because tighter targets can require more naive-MC samples than our compute
budget allows us to validate.

Let $N_{\mathrm{MC}}^\star(E_{0.95}^\star;\mu)$ be the budget naive MC
requires to meet the same error threshold; the \emph{Estimator Efficiency Gain} is
\begin{equation}
\label{eq:estimator-efficiency-gain}
\frac{N_{\mathrm{MC}}^\star(E_{0.95}^\star;\mu)}
{N^\star(E_{0.95}^\star;\mu)}.
\end{equation}
Values above one indicate fewer trajectories than naive MC at the same
accuracy and reliability, after proposal training.

To account for both proposal training and sampling costs, we define the total per-trajectory IS cost as
\begin{equation}
\mathrm{cost}_{\mathrm{IS}}^{\mathrm{total}}(N^\star) := \mathrm{cost}_{\mathrm{IS}} + \frac{\mathrm{cost}_{\mathrm{train}}}{N^\star(E_{0.95}^\star;\mu)}.
\end{equation}
We then report the \emph{Compute-Weighted Efficiency Gain} as the ratio of total compute required by naive Monte Carlo to that of importance sampling:
\begin{equation}
\label{eq:compute-weighted-efficiency-gain}
\frac{N_{\mathrm{MC}}^\star(E_{0.95}^\star;\mu) \, \mathrm{cost}_{\mathrm{MC}}}{N^\star(E_{0.95}^\star;\mu) \, \mathrm{cost}_{\mathrm{IS}}^{\mathrm{total}}(N^\star)}.
\end{equation}
Here, $\mathrm{cost}_{\mathrm{train}}$ is the upfront cost of training the proposal, $\mathrm{cost}_{\mathrm{IS}}$ is the cost per IS sample, and $\mathrm{cost}_{\mathrm{MC}}$ is the cost per naive-MC trajectory. Appendix~\ref{app:compute-cost} gives the measured costs across models and sequence lengths.

\paragraph{Can a small fixed budget yield useful estimates?}
Another practically useful question for an estimator would be whether it can quickly pin down a rare event's probability to be within certain orders of magnitude. We therefore also verify how much accuracy a small number of samples can provide after proposal training, evaluating $E_{0.95}$ using just $N=128$ trajectories per estimate, and presenting that result as event probability becomes exceedingly rare. 

\subsection{Baselines for rare-event estimation}
\label{sec:rare_event_algorithms}

We compare Iterative Unalignment (IU) against two families of baselines, namely naive Monte Carlo and cross-entropy (CE) methods. Additionally, we evaluate variants of our own approach that operate on restricted proposal families to isolate the contribution of weight-space tilting.

\paragraph{Naive Monte Carlo.} The standard baseline draws $N$ i.i.d.\ samples from the base model $p$ and estimates $\mu$ via the sample mean of $\Phi(\mathbf{y})$.

\paragraph{Iterative Unalignment (IU).} We implement the IU proposal using LoRA~\cite{hu2022lora} on both GPT-2 Small and Gemma-2. For Token Presence, we train each proposal with batch size $B=128$ for $T=300$ steps, using $38{,}400$ trajectories. LoRA introduces trainable adapter weights that modify the model's transformations. We found in preliminary experiments that full-parameter tilting yielded comparable performance, so we use LoRA to reduce training cost. Appendix~\ref{app:hyperparameters} gives the optimization, regularization, and evaluation settings.

\paragraph{IU variants (IU AS, IU Logit).}
To assess the effect of proposal expressivity, we evaluate two variants with fewer trainable parameters. IU AS adds a single $|h|$-dimensional vector to the residual stream at the output of every transformer block, after its attention and MLP computations. The vector is shared across layers and token positions. IU Logit adds a single $|\mathcal{V}|$-dimensional vector to the output logits. Both variants keep the base-model weights fixed and optimize the objective in Eq.~\eqref{eq:iu_objective}, treating sampled tokens as fixed. They use the same regularization and adaptive $\lambda$ tuning as IU.

\paragraph{Cross-entropy methods (CE AS, CE Logit).}
We evaluated several Cross-Entropy (CE) variants; Appendix~\ref{app:relationship-ce} provides detailed algorithm formulations and additional experimental results, and Appendix~\ref{app:hyperparameters} gives the hyperparameters. CE AS and CE Logit use the same steering vectors as IU AS and IU Logit. These parameterizations have fewer parameters than LoRA, allowing us to search for tilts using CE. In preliminary experiments, providing CE with more flexibility through methods such as LoRA resulted in poorer performance. For our primary baselines (CE AS and CE Logit), we use the unweighted Cross-Entropy method for optimization~\cite{rubinstein2004cross,botev2013cross}. In each iteration, this approach evaluates sampled candidates using a score function and updates the sampling distribution toward high-scoring trajectories to drive generation toward the target event. Classical formulations, such as Importance-Weighted CE (Algorithm~\ref{alg:ce-likelihood-fitting}), either failed to sufficiently amplify the target event or collapsed onto a narrow subset of trajectories across experimental settings.

\paragraph{Other output-space methods.}
We do not include Angell et al.~\cite{angell2026estimating} or Zhao et al.~\cite{zhao2024probabilistic} as baselines. Angell et al.\ require representative examples to construct activation directions and event hits to calibrate proposals. Zhao et al.\ learn prefix-dependent twist functions through contrastive training; when applied directly to a binary event potential, this training relies on obtaining event-positive trajectories to supply a positive learning signal. Obtaining such trajectories is itself a rare-event problem at the rarities we study.

Dorman et al.~\cite{dorman2026rare} sample score-biased trajectory ensembles with MCMC. Their framework permits the target indicator to differ from the biasing scalar score, allowing both methods to be evaluated on the same Token Presence events. We use these events for their verified tail reference probabilities, which their original score distributions lack. We evaluate increasingly rare events while allocating Dorman a conservatively charged compute budget equal to $1.38\times$ the full cost of an $N=128$ IU estimate (including proposal training). While Dorman et al.\ estimate moderate-frequency events accurately, their sequence-space MCMC sampler records no event hits for events below $\sim10^{-5}$, precluding a finite $E_{0.95}$ comparison in the deep tail. Appendix~\ref{app:dorman-comparison} gives the full comparison.

%% file: results.tex
\begingroup
\renewcommand{\topfraction}{0.65}
\renewcommand{\textfraction}{0.08}
\renewcommand{\floatpagefraction}{0.75}
\makeatletter
\setlength{\@fptop}{0pt}
\makeatother

\section{Results}
\label{experiments}

Our evaluation covers 319 events across
three settings and two models (GPT-2 Small and Gemma-2).
We report reference probabilities and their estimated relative standard errors
in Appendix~\ref{sec:soft-gt-cost}.
For each estimator and event, we use an evaluation budget of $64{,}000$
trajectories.
For Token Presence alone, training, evaluation, and reference estimation together require approximately $25.0$ million forward passes on batches of $128$ sequences (Appendix~\ref{app:token-forward-passes}). Evaluating estimator spread across repeated estimates for many events limits the model sizes and generation lengths we can study.

We summarize errors, required budgets, and efficiency gains across events using geometric means. This averages proportional changes symmetrically and gives aggregate gains no larger than the arithmetic mean.
To compare sample requirements at a given accuracy, we plot estimator efficiency gains over naive MC
as the error threshold tightens (higher is better).
To compare accuracy at a small fixed sampling budget, we plot quantile log
error at $N=128$ trajectories across event probabilities (lower is better).

\subsection{Token Presence}
\label{token-presence-results}

Token Presence events preserve the highly multimodal nature of rare events in our setting while allowing systematic evaluation across rarities, lengths, and models.

At $E_{0.95}^\star=1$, IU improves estimator efficiency over naive Monte Carlo in $230$ of the $235$ Token Presence configurations included in the efficiency comparison ($97.9\%$). Across these configurations, estimator efficiency gains are at least $10\times$, $100\times$, and $1{,}000\times$ in $92.8\%$, $81.7\%$, and $57.9\%$ of cases, respectively.

\begin{figure}[!t]
    \centering
    \includegraphics[width=\textwidth]{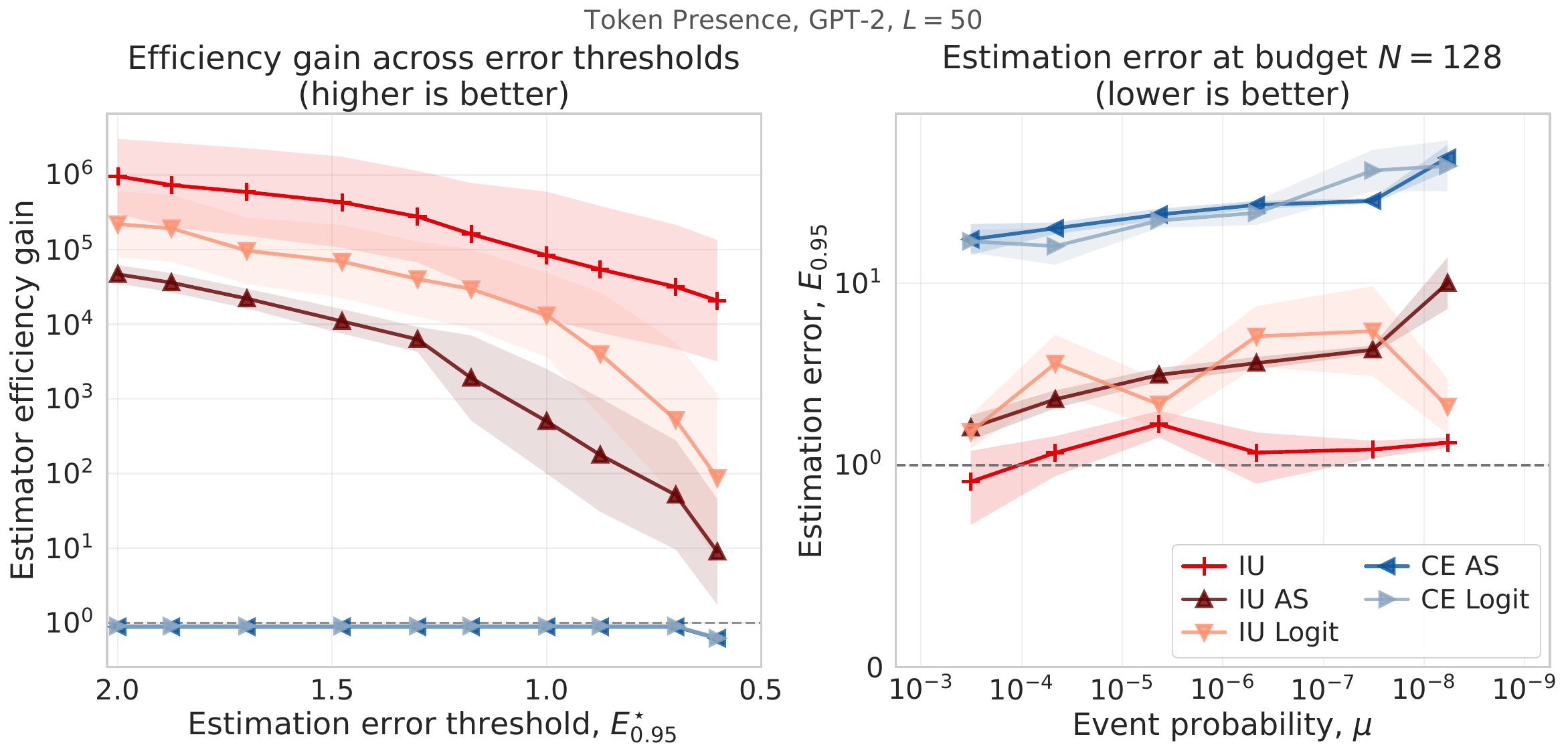}
    \caption{Token Presence estimation on GPT-2 Small at $L=50$. Left: estimator efficiency gains as accuracy requirements tighten. Right: error with $N=128$ samples across event probabilities. IU maintains larger gains and lower errors than the restricted updates and CE baselines. Curves show geometric means over ten events in $10^{-8}\leq\mu<10^{-7}$ on the left and ten events per rarity band on the right; shading shows $\pm0.5$ standard deviations in log space. At $E_{0.95}^\star=1$, the geometric-mean required budgets are approximately $1{,}103$ trajectories for IU and $9.24\times10^7$ for naive MC. Absolute budgets appear in Appendix Figure~\ref{fig:absolute-token-methods}.}
    \label{fig:our_method_better}
\end{figure}

\paragraph{Error threshold.}
On GPT-2 Small at $L=50$, IU achieves geometric-mean estimator efficiency gains above $20{,}000\times$ over naive Monte Carlo at every displayed error threshold (Figure~\ref{fig:our_method_better}, left). Compute-weighted efficiency gains exceed $800\times$ at every displayed threshold for this rarity band at each length $L=10$, $30$, and $50$ (Section~\ref{sec:computational-cost}).

IU also retains more of its estimator efficiency gains as accuracy requirements tighten than IU AS and IU Logit. This suggests that IU benefits from the flexibility of weight-space perturbations. CE AS and CE Logit, while achieving substantial estimator efficiency gains at some accuracy thresholds on shorter sequences (Appendix Figure~\ref{fig:full-token-length-10}), provide no measurable estimator efficiency gains over naive Monte Carlo on these longer events.

As accuracy requirements tighten, IU's efficiency gains decline, suggesting a gradual degradation in control over in-event weights. The efficiency frontiers include naive-MC samples when the estimator itself did not meet the required error within our available compute (Appendix~\ref{app:accuracy-cost-estimation}).

\paragraph{Rarity.}
As events become rarer, IU's required sampling budgets are significantly smaller than those of naive Monte Carlo. Across all evaluated sequence lengths, IU's estimator efficiency gains also consistently increase with event rarity (Figure~\ref{fig:length}).

With just $128$ trajectories per estimate after training, IU's error remains stable across rarity bands. Its log error is only $1.17$ orders of magnitude across the $60$ GPT-2 Small events with $10^{-9}\leq\mu<10^{-7}$ at $L=10$, $30$, and $50$.

\begin{figure}[t]
    \centering
    \includegraphics[width=\textwidth]{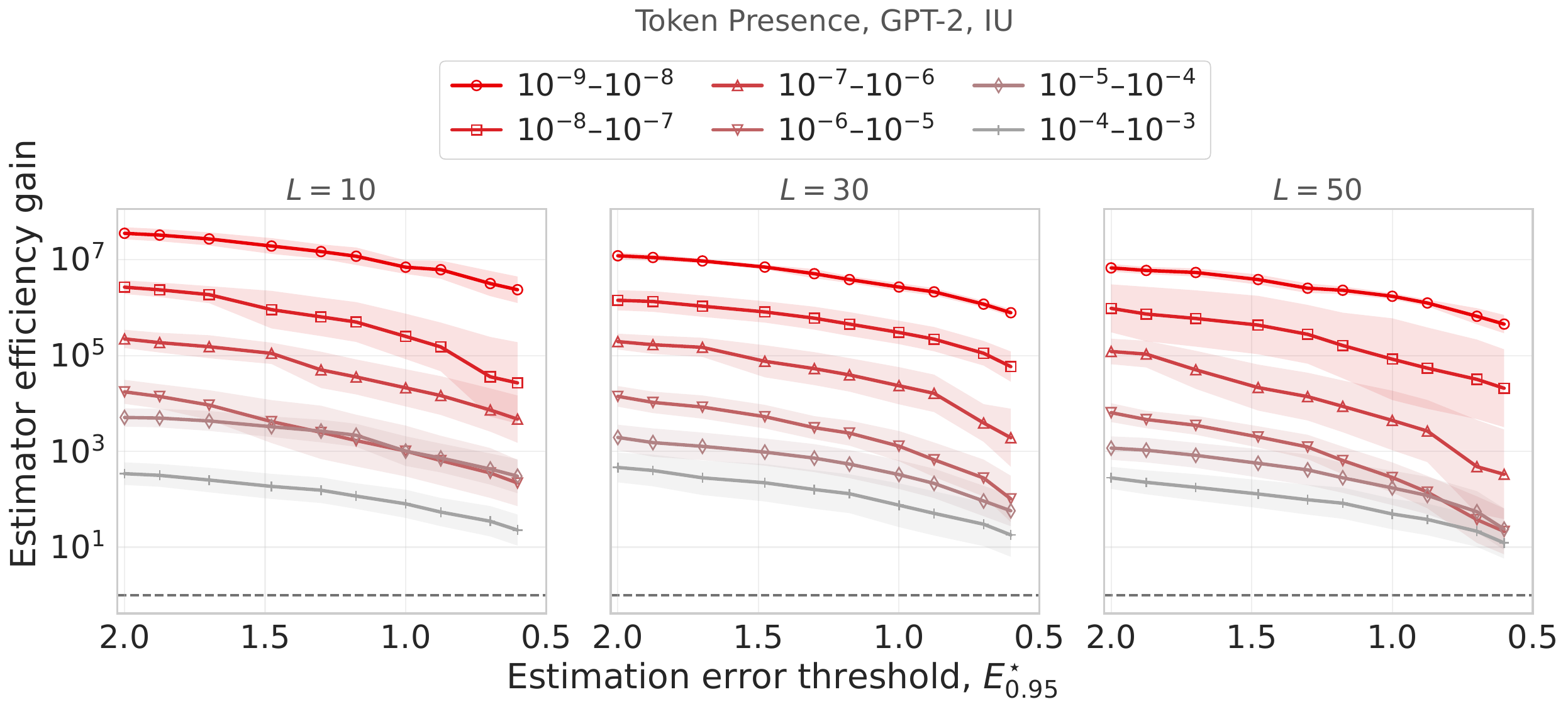}
    \caption{IU estimator efficiency gains for GPT-2 Small Token Presence at $L=10$, $30$, and $50$ (left to right). Gains tend to increase with event rarity across all three lengths. Each curve shows a geometric mean over ten events with neighboring rarity; shading shows $\pm0.5$ standard deviations in log space. Higher gains are better, and accuracy requirements become stricter toward the right. At $E_{0.95}^\star=1$ in $10^{-8}\leq\mu<10^{-7}$, geometric-mean IU versus naive-MC budgets are approximately $349$ versus $8.73\times10^7$, $255$ versus $7.73\times10^7$, and $1{,}103$ versus $9.24\times10^7$ trajectories, respectively. Absolute budgets appear in Appendix Figure~\ref{fig:absolute-token-lengths}.}
    \label{fig:length}
\end{figure}

\begin{figure}[t]
    \centering
    \includegraphics[width=\textwidth]{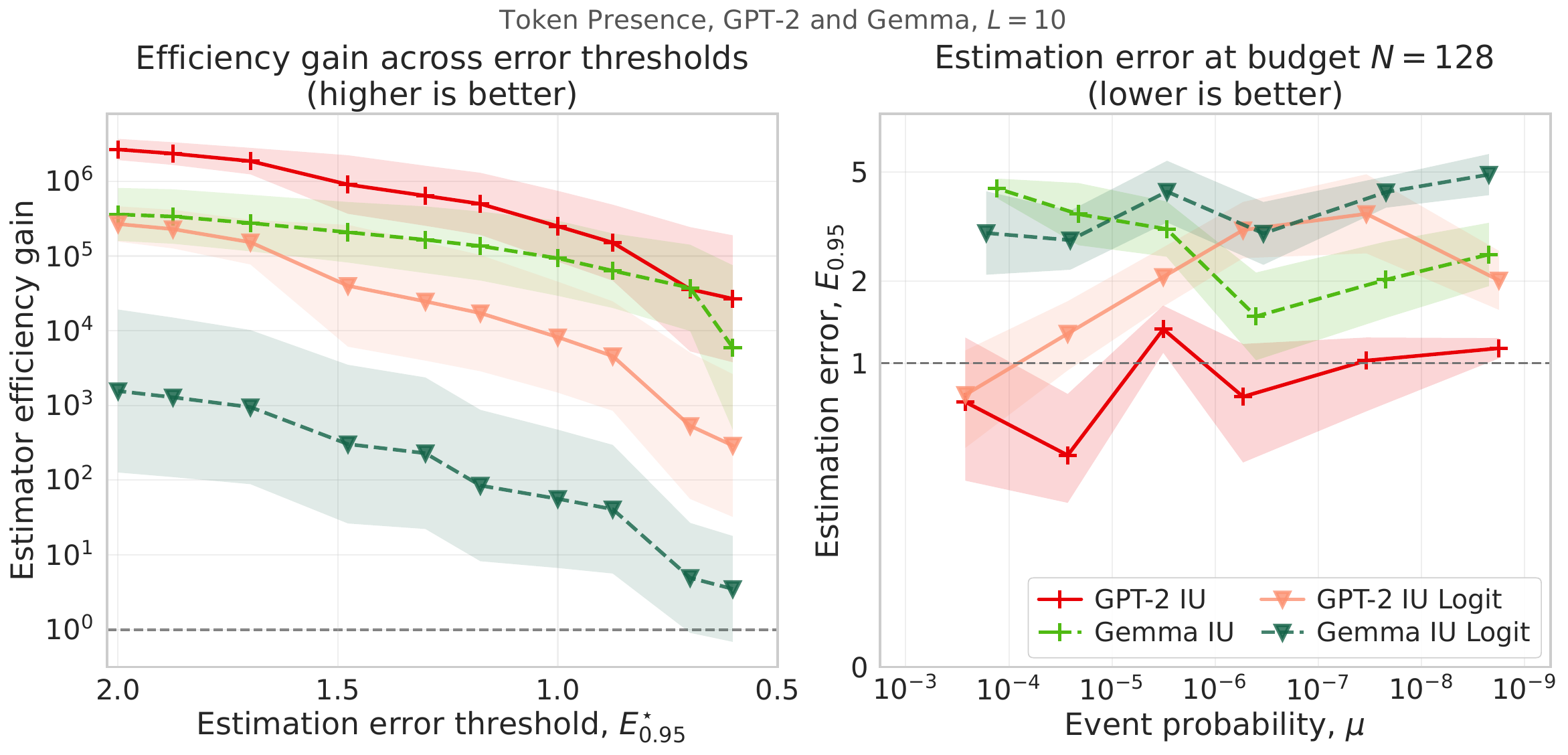}
    \caption{Token Presence estimation on GPT-2 Small and Gemma-2 at $L=10$. Left: estimator efficiency gains as accuracy requirements tighten. Right: error with $N=128$ samples across event probabilities. IU retains large estimator gains in both models. Plotted curves show geometric means over ten events per model in $10^{-8}\leq\mu<10^{-7}$ on the left and ten events with neighboring rarity on the right; shading shows $\pm0.5$ standard deviations in log space. At $E_{0.95}^\star=1$, geometric-mean IU versus naive-MC budgets are approximately $349$ versus $8.73\times10^7$ trajectories on GPT-2 and $1{,}445$ versus $1.35\times10^8$ on Gemma-2. Absolute budgets appear in Appendix Figure~\ref{fig:absolute-token-models}.}
    \label{fig:modelsize}
\end{figure}

\paragraph{Length.}
IU's performance remains stable as sequence length increases. For the rarest evaluated events ($10^{-9}\leq\mu<10^{-8}$ at $E_{0.95}^\star=1$), its geometric-mean estimator efficiency gains exceed $10^6$ at every tested length ($L=10$, $30$, and $50$) (Figure~\ref{fig:length}). This resilience may stem from the regularizer's dense, per-step control over the proposal distribution, which limits importance-weight variance along longer trajectories.

\paragraph{Model.}
We compare both models at $L=10$. At this length, IU still achieves larger geometric-mean estimator efficiency gains than IU Logit at every displayed threshold on both models (Figure~\ref{fig:modelsize}, left). Its gains remain substantial but are generally smaller on Gemma-2, where its geometric-mean errors at $N=128$ are also larger across rarity bands (right).

Gemma-2's larger vocabulary ($256{,}000$ versus $50{,}257$ tokens) may contribute to this decline. Improved surrogates or adaptive regularization may better control this larger space.

\subsection{Case study: Compositional Token Presence}
\label{compositional-token-results}

Compositional Token Presence extends the evaluation to events requiring multiple tokens. Reliably estimating events defined by token sets or orderings could support evaluation of more complex failures. Tables~\ref{tab:compositional-sequence-results} and~\ref{tab:compositional-and-results} report results at $E_{0.95}^\star=1$, a factor-ten error tolerance met by $95\%$ of estimates.

IU successfully reduces sample requirements for all ten evaluated compositional events, achieving estimator efficiency gains up to $2{,}090\times$ for Ordered Sequence events. Conjunctive events proved consistently more difficult, with maximum gains of $196\times$. Because Conjunctive events require all three target tokens to appear in any order anywhere in the trajectory, the proposal must maintain coverage over multiple valid orderings. This significantly complicates importance weight control compared to a fixed ordered sequence.

Including all compute costs, IU still provides compute-weighted efficiency gains over naive Monte Carlo for the rarest events, though these gains are much smaller (reaching $35.3\times$ and $11.3\times$, respectively). These more modest gains leave room for future work to improve surrogate design and adaptive regularization for multi-token conditions.

\begin{figure}[p]
    \captionsetup{type=table}
    \input{tables/compositional_sequence}
    \par\medskip
    \input{tables/compositional_and}
    \par\bigskip
    \captionsetup{type=figure}
    \centering
    \includegraphics[width=\textwidth]{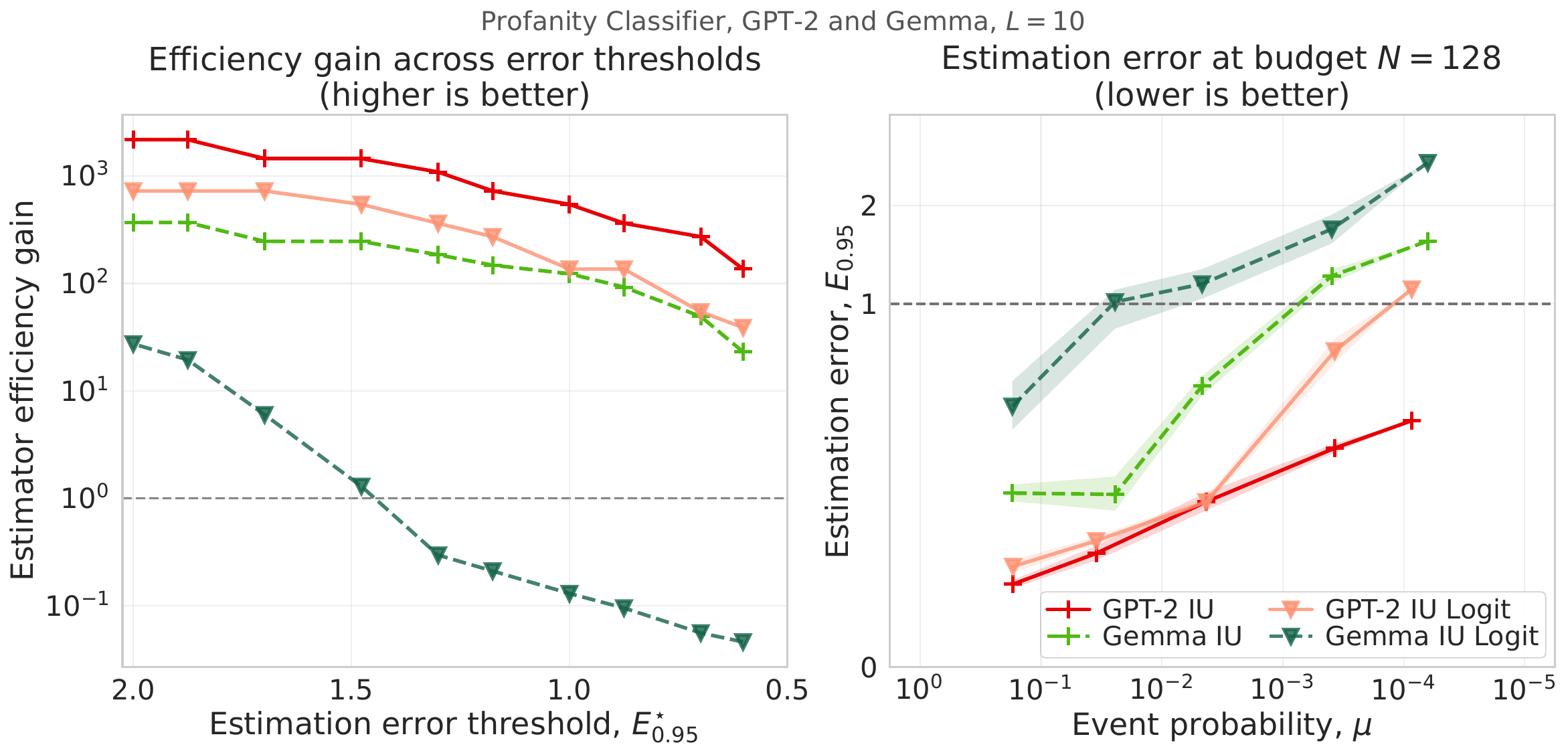}
    \caption{Profanity Classifier estimation on GPT-2 Small and Gemma-2 at $L=10$. IU achieves larger estimator efficiency gains and lower errors than IU Logit in both models. Left: estimator gains as accuracy requirements tighten at classifier cutoff $\kappa=0.999$. Right: geometric-mean error with $N=128$ samples across event rarity. Each point combines 5 events with neighboring rarity; shading shows $\pm0.5$ standard deviations in log space. At $E_{0.95}^\star=1$, IU versus naive-MC budgets are $64$ versus $34{,}858$ trajectories on GPT-2 and $384$ versus $47{,}311$ on Gemma-2. Absolute budgets appear in Appendix Figure~\ref{fig:absolute-profanity-models}.}
    \label{fig:profanity}
\end{figure}

\subsection{Case study: Profanity Classifier}
\label{profanity-classifier-results}
Here a classifier assigns a profanity probability to a sequence of tokens, and the event occurs when $f(\mathbf{y})>\kappa$. Raising $\kappa$ makes the event rarer. This follows the traditional rare-event formulation in which a score crossing a threshold defines the event.

At $E_{0.95}^\star=1$, IU improves estimator efficiency over naive Monte Carlo in $44$ of the $47$ Profanity Classifier configurations included in the efficiency comparison ($93.6\%$).

Across verified rarities in this setting, IU still achieved substantial gains over naive MC at both $L=50$ and $L=10$. At $L=10$, IU has larger estimator efficiency gains across accuracy requirements and lower aggregate errors across rarity levels than IU Logit on both models (Figure~\ref{fig:profanity}). On Gemma-2, IU Logit falls below naive-MC estimator efficiency at stricter requirements, while IU retains a sampling advantage. On GPT-2 Small at $L=50$, IU AS also loses its gains as requirements tighten, and both CE methods produce large errors and require more samples than naive Monte Carlo (Appendix Figure~\ref{fig:full-profanity-rarity}).
\begin{figure}[!t]
    \centering
    \includegraphics[width=\textwidth]{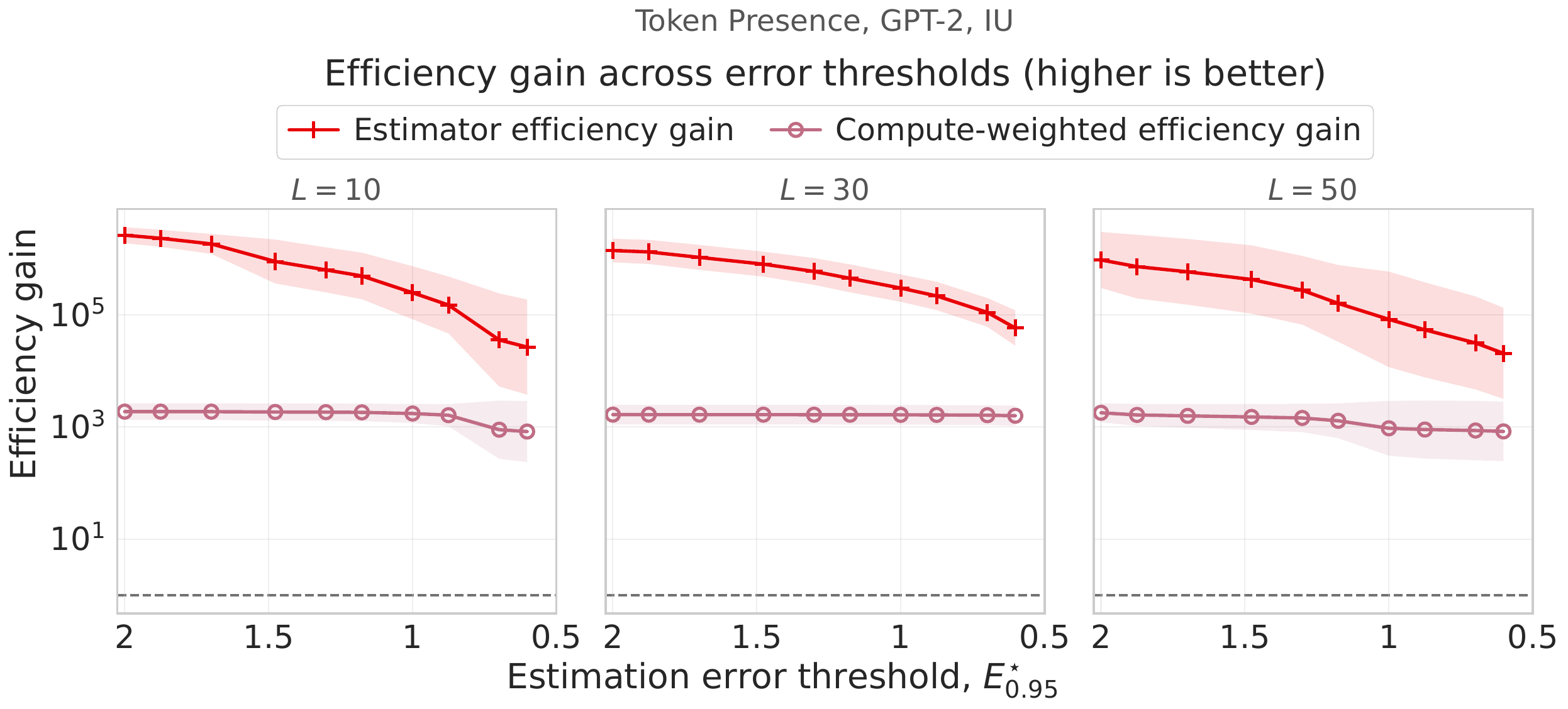}
    \caption{IU estimator efficiency and compute-weighted efficiency gains on GPT-2 Small at $L=10$, $30$, and $50$ (left to right). As accuracy requirements tighten, training accounts for less of the total cost and the gap narrows. Curves show geometric means over ten events per length in $10^{-8}\leq\mu<10^{-7}$; shading shows $\pm0.5$ standard deviations in log space. At $E_{0.95}^\star=1$, the geometric-mean compute-weighted efficiency gains over naive MC are approximately $1{,}750\times$, $1{,}660\times$, and $949\times$, respectively. Absolute sample and time budgets appear in Appendix Figure~\ref{fig:absolute-computational-cost}.}
    \label{fig:computational_efficiency}
\end{figure}

\subsection{Computational cost}
\label{sec:computational-cost}

IU achieves geometric-mean compute-weighted efficiency gains above $800\times$ over naive Monte Carlo at every displayed error threshold for GPT-2 Small Token Presence at $L=10$, $30$, and $50$ (Figure~\ref{fig:computational_efficiency}). Each curve shows a geometric mean over ten events in $10^{-8}\leq\mu<10^{-7}$.

Across all $80$ Token Presence configurations with $10^{-9}\leq\mu<10^{-7}$, covering GPT-2 Small at $L=10$, $30$, and $50$ and Gemma-2 at $L=10$, the median compute-weighted efficiency gain over naive Monte Carlo is $5{,}670\times$, with an interquartile range of $1{,}560\times$--$16{,}100\times$. These comparisons use $E_{0.95}^\star=1$ and include proposal-training and sampling costs.

We train every IU proposal for the same number of steps and measure training costs separately for each model and length (Appendix~\ref{app:compute-cost}). At lower accuracy requirements, training costs dominate the total compute budget because the fixed cost of training the proposal is much higher than drawing the few samples needed for estimation. When training dominates, differences in the number of required samples have little effect on IU's overall computational cost, and the efficiency gains mostly reflect the changing cost of naive MC. For this reason, we defer the complete set of compute-weighted efficiency plots to Appendix~\ref{app:full-result-plots}.

However, as stricter accuracy requirements call for more samples, sampling costs increase and training accounts for a progressively smaller share of IU's total cost. Consequently, the gap between estimator efficiency and compute-weighted efficiency gains narrows across the thresholds in Figure~\ref{fig:computational_efficiency}. On GPT-2 Small, IU's compute-weighted efficiency gains increase with event rarity and remain above $800\times$ across the evaluated sequence lengths. These trends suggest that IU has the potential to remain useful as events become rarer and trajectories grow longer.

\begingroup
\newcommand{\qualitativesample}[1]{%
    \begin{quote}
    \setlength{\fboxsep}{6pt}%
    \setlength{\fboxrule}{0.4pt}%
    \noindent\fbox{%
        \begin{minipage}{\dimexpr\linewidth-2\fboxsep-2\fboxrule\relax}
        \small\ttfamily\centering
        #1\par
        \end{minipage}%
    }
    \end{quote}%
}
\noindent\begin{minipage}{\linewidth}
\subsection{Qualitative examples}
\label{sec:qualitative-examples}

The following examples show high importance-weight trajectories from saved GPT-2 Small IU proposals for the four event types. IU model completions retain natural grammatical structure while satisfying the event conditions. This suggests that IU may help reveal ways a specified event can occur, in addition to estimating its probability.

All samples include the prompt \texttt{Once upon a time}; $L$ counts only generated tokens, and $\mu$ is the base-model event probability. Profanity is masked for display purposes.

\par\medskip\noindent
\begin{minipage}{\linewidth}
\raggedright
\textbf{Token Presence.}\\
Contains \texttt{Sorceress}; $L=50$, $\mu=6.12\times10^{-5}$.
\qualitativesample{%
Once upon a time, Ye Hao thought out how \textbf{Sorceress} Luo Yang should intervene and help him regain his immortality. Yukkun frowned and pouted. Even if it looked silly, laughing would be refreshing to her. It's all simple, but Ye Hao couldn't
}

\end{minipage}
\end{minipage}
\par\medskip\noindent
\begin{minipage}{\linewidth}
\raggedright
\textbf{Profanity Classifier.}\\
Classifier probability exceeds $0.999$; $L=50$, $\mu=1.44\times10^{-3}$.
\qualitativesample{%
Once upon a time he made such a f*****g joke out of his own blood… s**t was boiling. 'No one would f*****g tell you that I screwed him over,' he said as he put his head together. 'Yeah, but nobody said that.
}

\end{minipage}
\par\medskip\noindent
\begin{minipage}{\linewidth}
\raggedright
\textbf{Conjunctive event.}\\
Contains \texttt{the}, \texttt{was}, and \texttt{village} in any order; $L=10$, $\mu=8.06\times10^{-5}$.
\qualitativesample{%
Once upon a time, \textbf{the village was} fooling itself.
}

\end{minipage}
\par\medskip\noindent
\begin{minipage}{\linewidth}
\raggedright
\textbf{Ordered Sequence event.}\\
Contains \texttt{the} $\rightarrow$ \texttt{was} $\rightarrow$ \texttt{village}, with gaps allowed; $L=10$, $\mu=5.75\times10^{-6}$.
\qualitativesample{%
Once upon a time \textbf{the} kingdom \textbf{was} abolished altogether and the \textbf{village} destroyed.
}
\end{minipage}
\par
\endgroup

\makeatletter
\ifx\@deferlist\@empty\else\clearpage\fi
\makeatother
\endgroup

%% file: tables/compositional_sequence.tex
\begingroup
    \centering
    \small
    \setlength{\tabcolsep}{3pt}
    \begin{tabular*}{\textwidth}{@{\extracolsep{\fill}}lrrrrr@{}}
        \toprule
        Target tokens & \shortstack{Rarity\\$\mu$} & \shortstack{MC samples\\$N_{\mathrm{MC}}^\star$} & \shortstack{IU samples\\$N_{\mathrm{IU}}^\star$} & \shortstack{Estimator\\efficiency gain} & \shortstack{Compute-weighted\\efficiency gain} \\
        \midrule
        \texttt{the} $\rightarrow$ \texttt{was} $\rightarrow$ \texttt{and} & $2.59\times 10^{-3}$ & $1{,}156$ & $96$ & $12\times$ & $0.0169\times$ \\
        \texttt{the} $\rightarrow$ \texttt{was} $\rightarrow$ \texttt{people} & $1.03\times 10^{-4}$ & $28{,}999$ & $256$ & $113\times$ & $0.423\times$ \\
        \texttt{a} $\rightarrow$ \texttt{the} $\rightarrow$ \texttt{king} & $3.02\times 10^{-5}$ & $99{,}215$ & $256$ & $388\times$ & $1.44\times$ \\
        \texttt{the} $\rightarrow$ \texttt{was} $\rightarrow$ \texttt{village} & $5.75\times 10^{-6}$ & $521{,}040$ & $512$ & $1{,}020\times$ & $7.55\times$ \\
        \texttt{the} $\rightarrow$ \texttt{was} $\rightarrow$ \texttt{connect} & $1.25\times 10^{-6}$ & $2{,}402{,}465$ & $1{,}152$ & $2{,}090\times$ & $35.3\times$ \\
        \bottomrule
    \end{tabular*}
    \caption{Ordered Sequence (SEQ) events on GPT-2 Small at $L=10$, requiring factor-ten accuracy in $95\%$ of estimates. Targets appear in order, but need not be contiguous. IU reduces sample requirements for every event; gains including training cost are largest for the rarest events.}
    \label{tab:compositional-sequence-results}
\endgroup

%% file: tables/compositional_and.tex
\begingroup
    \centering
    \small
    \setlength{\tabcolsep}{3pt}
    \begin{tabular*}{\textwidth}{@{\extracolsep{\fill}}lrrrrr@{}}
        \toprule
        Target tokens & \shortstack{Rarity\\$\mu$} & \shortstack{MC samples\\$N_{\mathrm{MC}}^\star$} & \shortstack{IU samples\\$N_{\mathrm{IU}}^\star$} & \shortstack{Estimator\\efficiency gain} & \shortstack{Compute-weighted\\efficiency gain} \\
        \midrule
        \texttt{\{the, was, and\}} & $6.57\times 10^{-3}$ & $455$ & $256$ & $1.78\times$ & $0.00686\times$ \\
        \texttt{\{the, was, people\}} & $5.05\times 10^{-4}$ & $5{,}927$ & $384$ & $15.4\times$ & $0.0893\times$ \\
        \texttt{\{a, the, king\}} & $1.98\times 10^{-4}$ & $15{,}118$ & $384$ & $39.4\times$ & $0.228\times$ \\
        \texttt{\{the, was, village\}} & $8.06\times 10^{-5}$ & $37{,}162$ & $640$ & $58.1\times$ & $0.557\times$ \\
        \texttt{\{the, was, connect\}} & $3.72\times 10^{-6}$ & $804{,}821$ & $4{,}096$ & $196\times$ & $11.3\times$ \\
        \bottomrule
    \end{tabular*}
    \caption{Conjunctive (AND) events on GPT-2 Small at $L=10$, requiring factor-ten accuracy in $95\%$ of estimates. Every target appears, in any order. IU reduces sample requirements for every event; gains including training cost are largest for the rarest events.}
    \label{tab:compositional-and-results}
\endgroup

%% file: conclusion.tex
\section{Discussion}
\label{sec:discussion}

We introduce Iterative Unalignment (IU), which searches over proposal distributions via weight-space perturbations to form an IS proposal for estimating rare-event probabilities under a fixed, potentially post-trained base model. While we demonstrate substantial gains in controlled environments, evaluating on complex real-world failures remains challenging due to the lack of verifiable ground-truth probabilities. Extending this evaluation requires benchmarks with more realistic failures and reference probabilities precise enough to assess estimator accuracy.

\paragraph{Importance sampling and alignment.}
Efficient importance sampling seeks a proposal that makes the event more frequent while matching the base model's conditional distribution $p(\cdot\mid\mathcal{E})$ within it. Learning such a proposal involves finding where the base model places probability mass in the rare region and preserving the relative probabilities of event trajectories. With event definitions that capture meaningful failures, this search has the potential to reveal the base model's preferences among failure trajectories. These failure trajectories may help guide alignment efforts.

\paragraph{Further developing IU.}
Effective rare-event estimation requires both frequent event samples and adequate coverage of the trajectories that contribute to the event probability. Greater flexibility can complicate this balance by allowing the proposal to concentrate on a narrow subset of trajectories. Yet restricting the proposal can also limit its ability to amplify the event while preserving the base model's relative probabilities within it. Our results suggest that flexibility is useful for learning these changes across an autoregressive sequence.

The sampling gains across all three event families demonstrate IU's potential as a framework for rare-event estimation. The Gemma-2 and Compositional Token Presence results also suggest that controlling importance weights may be more difficult for larger models and more complex events. Better surrogate design and adaptive regularization schemes that control within-event distortion while allowing event frequency to increase are promising directions for improving this framework. Developing these components together may yield further gains in accuracy and efficiency.

\paragraph{Systematic evaluation.}
Our deeper rarity evaluation already reveals methodological limitations. Reinforcing observed trajectories and score-guided MCMC can both fail where IU enables estimation (Appendices~\ref{app:relationship-ce} and~\ref{app:dorman-comparison}). These findings highlight a gap between having an informative search signal and constructing a stable rare-event proposal. Systematically studying estimator performance across rarity, sequence length, and model and vocabulary size is an important research direction. Extending this evaluation to more realistic, complex failures requires scalable settings where ground-truth probabilities can be reliably obtained.

%% file: appendix.tex
\appendix

\section{Notation}
\label{app:notation}

\paragraph{Importance Sampling in LMs}
\begin{center}
\small
\begin{tabular}{@{}p{0.20\columnwidth}p{0.74\columnwidth}@{}}
\toprule
Symbol & Meaning \\
\midrule
$\mathcal{V}$, $|\mathcal{V}|$ & Vocabulary and its size \\
$u$ & Generic vocabulary token \\
$L$ & Output sequence length \\
$\ell$ & Decoding position, $\ell\in\{1,\dots,L\}$ \\
$\mathbf{y}=(y_1,\dots,y_L)$ & Output sequence (trajectory) in $\mathcal{V}^L$ \\
$\mathbf{y}_{<\ell}$ & Prefix preceding position $\ell$ \\
$\mathbf{x}$ & Input sequence; held fixed and suppressed from the notation \\
$\theta$ & Base-model parameters \\
$p_\theta$, $p$ & Base model; we write $p:=p_\theta$ since $\theta$ remains fixed \\
$\delta$, $\delta_t$ & Generic weight-space perturbation and its value at training iteration $t$ \\
$q_\delta:=p_{\theta+\delta}$ & Proposal model \\
$\Phi$ & Binary event indicator, $\Phi:\mathcal{V}^L\to\{0,1\}$ \\
$\mathcal{E}$ & Rare event $\{\Phi(\mathbf{y})=1\}$ \\
$\mu$ & Target probability $\mathbb{E}_p[\Phi(\mathbf{y})]$ \\
$\widehat{\mu}$ & Estimate of $\mu$ \\
$w(\mathbf{y})$ & Trajectory importance weight $p(\mathbf{y})/q_\delta(\mathbf{y})$ \\
$r_\delta$, $q_\delta^{\mathcal{E}}$ & Proposal event probability $q_\delta(\mathcal{E})$ and conditional distribution $q_\delta(\cdot\mid\mathcal{E})$ \\
$q^*$ & Zero-variance proposal $p(\mathbf{y})\Phi(\mathbf{y})/\mu$ \\
\bottomrule
\end{tabular}
\end{center}

\paragraph{Evaluation settings.}
The superscript names the event family. In expressions such as $\Phi^{\mathrm{TOK}}(\mathbf{y};v)$, $\mathbf{y}$ is the trajectory being evaluated, while the quantity after the semicolon specifies the event instance. The bare symbol $\Phi^{\mathrm{TOK}}$ denotes the corresponding family; the same convention applies to $S_\delta^{\mathrm{TOK}}$.
\begin{center}
\footnotesize
\begin{tabular}{@{}lcccc@{}}
\toprule
Event & Indicator & \shortstack{Reference probability\\estimator} & \shortstack{Reference probability} & Differentiable surrogate \\
\midrule
Token Presence
& $\Phi^{\mathrm{TOK}}(\mathbf{y};v)$
& $\widehat{\Phi}^{\mathrm{TOK}}(\mathbf{y};v)$
& $\mu^{\mathrm{TOK}}(v)$
& $S^{\mathrm{TOK}}_\delta(\mathbf{y};v)$ \\
Conjunctive (AND)
& $\Phi^{\mathrm{AND}}(\mathbf{y};\mathbf{v})$
& $\widehat{\Phi}^{\mathrm{AND}}(\mathbf{y};\mathbf{v})$
& $\mu^{\mathrm{AND}}(\mathbf{v})$
& $S^{\mathrm{AND}}_\delta(\mathbf{y};\mathbf{v})$ \\
Ordered Sequence (SEQ)
& $\Phi^{\mathrm{SEQ}}(\mathbf{y};\mathbf{v})$
& $\widehat{\Phi}^{\mathrm{SEQ}}(\mathbf{y};\mathbf{v})$
& $\mu^{\mathrm{SEQ}}(\mathbf{v})$
& $S^{\mathrm{SEQ}}_\delta(\mathbf{y};\mathbf{v})$ \\
Profanity Classifier
& $\Phi^{\mathrm{PROF}}(\mathbf{y};\kappa)$
& ---
& $\mu^{\mathrm{PROF}}(\kappa)$
& $S^{\mathrm{PROF}}_\delta(\mathbf{y};\kappa)$ \\
\bottomrule
\end{tabular}
\end{center}

\begin{center}
\small
\begin{tabular}{@{}p{0.20\columnwidth}p{0.74\columnwidth}@{}}
\toprule
Symbol & Meaning \\
\midrule
$v$ & Target token for Token Presence \\
$\mathbf{v}=(v_1,\dots,v_m)$ & Tuple of target tokens for Compositional Token Presence; $m$ is the number of targets \\
$f$, $\kappa$ & Profanity Classifier probability and its threshold; $\Phi^{\mathrm{PROF}}(\mathbf{y};\kappa)=\mathbb{I}\{f(\mathbf{y})>\kappa\}$ \\
\bottomrule
\end{tabular}
\end{center}

\paragraph{Iterative Unalignment objective and training.}
\begin{center}
\small
\begin{tabular}{@{}p{0.20\columnwidth}p{0.74\columnwidth}@{}}
\toprule
Symbol & Meaning \\
\midrule
$S_\delta(\mathbf{y})$ & Differentiable surrogate supplying the amplification signal \\
$\mathcal{R}_\delta(\mathbf{y})$ & Dense reverse-KL regularizer~\eqref{eq:reg-final} \\
$\mathcal{J}_t$ & Empirical IU objective at iteration $t$ \\
$\lambda_t$, $\lambda_0$, $\lambda_{\min}$ & Regularization strength, its initial value, and its floor \\
$T$, $t$ & Total number of training iterations and the training iteration index \\
$B$, $i$ & Training batch size and batch sample index \\
$\eta_\delta$, $\eta_\lambda$ & Learning rates for $\delta$ and for $\lambda$ \\
$\mathrm{ESSr}_{\mathrm{batch}}$, $\mathrm{ESSr}_{\mathrm{event}}$ & Full-batch and in-event effective sample size ratios \\
$\rho_{\mathrm{batch}}$, $\rho_{\mathrm{event}}$ & Targets for the two effective sample size ratios \\
$\mathrm{ESSr}_t$, $\rho_t$ & Active effective sample size ratio and its target at iteration $t$ \\
$\hat h_t$, $h^\star$ & Empirical event frequency and the phase-switch threshold \\
\bottomrule
\end{tabular}
\end{center}

\paragraph{Evaluation protocol and metrics.}
\begin{center}
\small
\begin{tabular}{@{}p{0.32\columnwidth}p{0.62\columnwidth}@{}}
\toprule
Symbol & Meaning \\
\midrule
$N$ & Estimation budget \\
$K$, $k$ & Number of independent repetitions and the repetition index, giving $\{\widehat{\mu}_{k,N}\}_{k=1}^K$ \\
$E_{0.95}(N;\mu)$ & Quantile log error~\eqref{eq:direct-log-error} \\
$E_{0.95}^\star$ & Estimation error threshold: largest acceptable quantile log error \\
$N^\star(E_{0.95}^\star;\mu)$ & Budget required to satisfy the error threshold~\eqref{eq:required-budget} \\
$N_{\mathrm{MC}}^\star(E_{0.95}^\star;\mu)$ & Exact Binomial naive-MC budget for the same threshold \\
$\mathrm{cost}_{\mathrm{MC}}$, $\mathrm{cost}_{\mathrm{IS}}$, $\mathrm{cost}_{\mathrm{train}}$ & Per-trajectory wall-clock cost under MC and IS, and proposal training cost \\
\bottomrule
\end{tabular}
\end{center}

\clearpage
\section{Additional experimental-setup diagnostic}
\label{app:additional-experimental-setup-diagnostic}

\begin{figure}[h]
    \centering
    \begin{minipage}[b]{0.33\textwidth}
        \centering
        \includegraphics[width=0.8\linewidth]{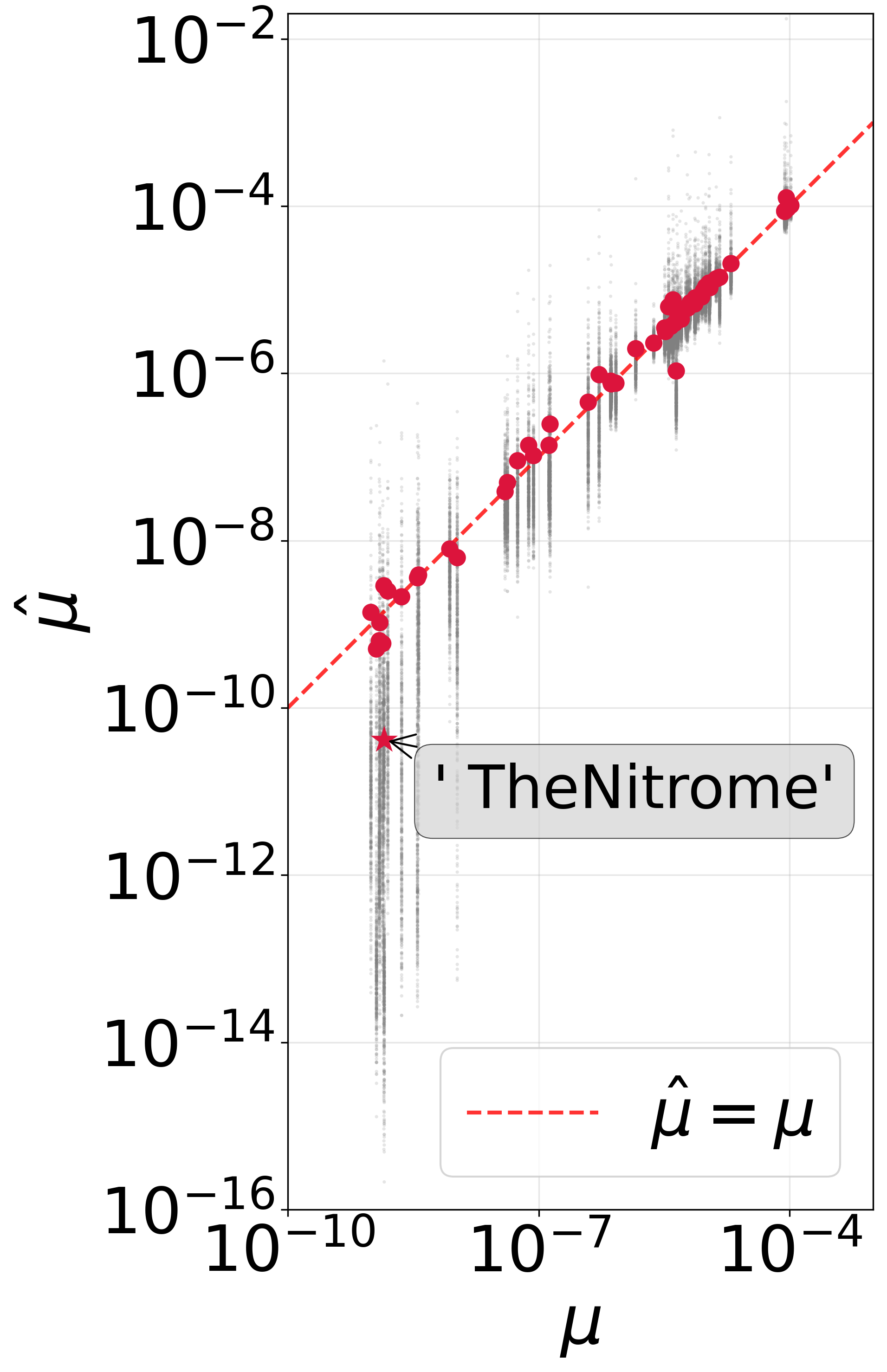}
    \end{minipage}
    \hfill
    \begin{minipage}[b]{0.66\textwidth}
        \centering
        \includegraphics[width=0.80\linewidth]{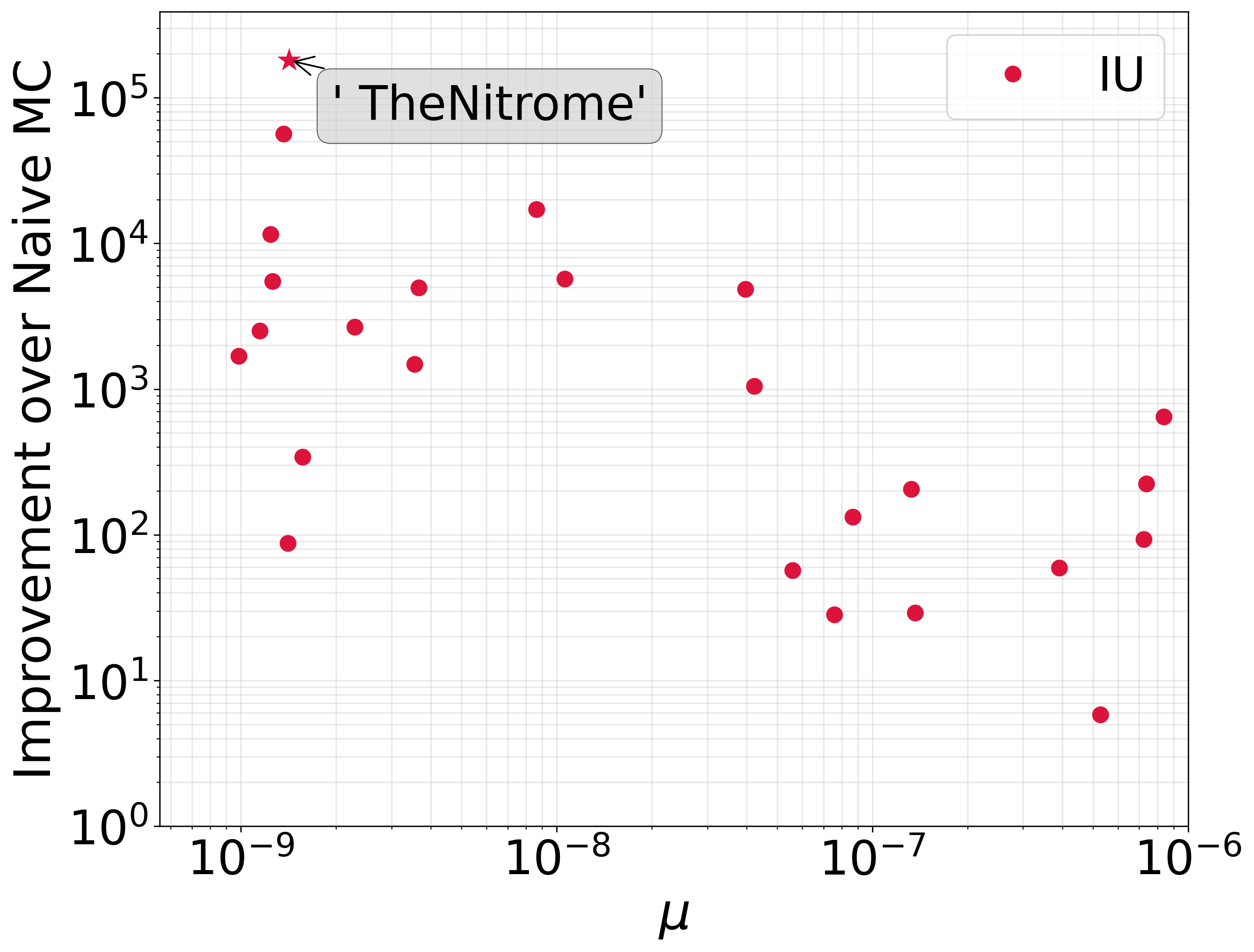}
    \end{minipage}
    \caption{The asymmetric penalty structure of rMSE leads to misleading comparisons in the rare-event regime. \textbf{Left panel.} The range of estimates $\widehat{\mu}_k$ versus true probability $\mu$ for tokens across six orders of magnitude in rarity. Each black dot represents an independent estimate, with error bars indicating variability across repetitions. The red dashed line denotes perfect estimation ($\widehat{\mu} = \mu$). At $\mu \approx 10^{-10}$, the rare event~\eqref{eqn:event} defined by the token \texttt{TheNitrome} (highlighted) shows the worst estimation quality, with estimates highly dispersed and systematically biased below the diagonal. \textbf{Right panel.} The improvement factor in rMSE relative to naive Monte Carlo. Paradoxically, we see the largest improvement in rMSE on this event. This occurs because estimators underestimate at this rarity, which does not get penalized much under rMSE's asymmetry.}
    \label{fig:rmse_pathology}
\end{figure}

\section{Efficiency gains across error thresholds}
\label{app:accuracy-cost-estimation}

We bootstrap the stored estimates up to four times the evaluation budget,
or $256{,}000$ trajectories from $64{,}000$ stored trajectories. If the target
remains unmet, we add simulated naive-MC samples, since naive MC gives an
unbiased estimate of the event probability. We simulate event counts from
the Binomial distribution using the reference probability, retain the initial
IS contributions, and count all IS and MC samples in the estimate and budget.

This gives a conservative sketch when further IS sampling would outperform
naive MC, since we assume no further IS advantage and retain the initial IS
error and cost. Bootstrapping can also understate gains when the stored data
miss rare event contributions. These estimates remain subject to bootstrap
uncertainty and are not statistical lower bounds. The error-threshold axis
stops at $0.5$ because tighter targets can require more naive-MC samples than
our compute budget allows us to validate.

\section{Implementation and hyperparameters}
\label{app:hyperparameters}

\paragraph{Training and evaluation.}
For Token Presence, we use batches of $128$ trajectories and a training budget of $300$ steps per proposal. We then freeze the proposal and draw $500$ evaluation batches, totaling $64{,}000$ trajectories. Appendix~\ref{app:compute-cost} reports the measured computational costs and accounts for the forward passes across events, methods, and sequence lengths.

\paragraph{IU.}
Both GPT-2 Small and Gemma-2 use LoRA with rank $r=16$ and scaling factor $\alpha_{\mathrm{LoRA}}=32$. The proposal learning rate is $\eta_\delta=10^{-4}$. We initialize the regularization strength at $\lambda_0=10$, impose a floor of $\lambda_{\min}=0.1$, and use dual learning rate $\eta_\lambda=10^{-2}$. The controller targets population ESS ratio $\rho_{\mathrm{batch}}=0.005$ while the event frequency is below $h^\star=0.1$. Once at least $10\%$ of the batch satisfies the event, it targets event ESS ratio $\rho_{\mathrm{event}}=0.1$.

\paragraph{IU AS and IU Logit.}
The restricted variants use the same regularization settings as IU. Their proposal learning rates are $10^{-2}$ for IU AS and $1$ for IU Logit. IU AS initializes each steering-vector coordinate to $0.01$; IU Logit initializes its coordinates independently from a zero-mean Gaussian with standard deviation $0.01$.

\paragraph{CE AS and CE Logit.}
Both CE methods use a population of $128$ steering vectors, elite fraction $\gamma_{\mathrm{elite}}=0.3$, and smoothing coefficient $\tau_{\mathrm{CE}}=0.5$. The initial search standard deviation and its floor are both $0.1$ for CE AS and $1$ for CE Logit. The search mean is initialized to zero, and generation uses temperature $1$. Evaluation fixes the steering vector at the learned search mean.

After a warmup of $10$ training steps, CE stops early if the event ESS ratio falls below $0.1$ for two consecutive steps, each with at least two event hits. Appendix~\ref{app:relationship-ce} describes the CE formulations and their empirical behavior.

\section{Computational costs}
\label{app:compute-cost}

\begin{table}[H]
\centering
\caption{Wall-clock costs for Compute-Weighted Efficiency Gain, measured on an isolated NVIDIA A100-SXM4-80GB GPU. IU uses LoRA ($r=16$, $\alpha_{\mathrm{LoRA}}=32$); training cost is for $T=300$ steps at $B=128$.}
\label{tab:compute-cost-constants}
\small
\begin{tabular}{@{}lccccc@{}}
\toprule
Model & $L$ & $\mathrm{cost}_{\mathrm{MC}}$ & $\mathrm{cost}_{\mathrm{IS}}$ & $\mathrm{cost}_{\mathrm{train}}$ & $\mathrm{cost}_{\mathrm{IS}}/\mathrm{cost}_{\mathrm{MC}}$ \\
 & & \multicolumn{2}{c}{(s / trajectory)} & (s) & \\
\midrule
GPT-2 Small & 10 & 0.0046 & 0.0051 & 213 & 1.10 \\
GPT-2 Small & 30 & 0.0138 & 0.0153 & 640 & 1.10 \\
GPT-2 Small & 50 & 0.0230 & 0.0254 & 1067 & 1.10 \\
Gemma-2-2B-IT & 10 & 0.0501 & 0.0569 & 2718 & 1.14 \\
\bottomrule
\end{tabular}
\end{table}

\subsection{Forward-pass accounting for Token Presence}
\label{app:token-forward-passes}
We count each model forward pass on a batch of $128$ sequences as one pass. GPT-2 has $60$ events at each of $L=10,30,50$, evaluated with five methods. Gemma-2 has $55$ IU and $66$ IU Logit runs, all at $L=10$. This gives $60\cdot3\cdot5+55+66=1{,}021$ runs, each with $300$ training and $500$ evaluation batches.

Generating a batch of length $L$ requires $L$ forward passes. The generation budget is therefore
\begin{align*}
F_{\mathrm{gen}}
&= (300+500)\bigl[60\cdot5\cdot(10+30+50)+(55+66)\cdot10\bigr] \\
&= 22{,}568{,}000.
\end{align*}
Each batch also requires one proposal and one base-model scoring pass, giving $2\cdot(300+500)\cdot1{,}021=1{,}633{,}600$ scoring passes.

Reference estimation uses $1{,}024{,}000$ sequences, or $8{,}000$ batches, in each of the four model--length settings. Each batch requires $L$ generation passes and one scoring pass, giving
\[
F_{\mathrm{ref}}=8{,}000\bigl[(10+1)+(30+1)+(50+1)+(10+1)\bigr]=832{,}000.
\]
The total budget is $25{,}033{,}600$ forward passes. Broken down, this is $22.57$ million for generation, $1.63$ million for scoring, and $0.83$ million for reference estimation.

\input{dorman_comparison}

\section{Ground truth estimation}
\label{sec:surrogate_proofs}

We establish the unbiasedness of the token-based reference estimators and report the relative standard errors of the reference probabilities used in our evaluation.

\subsection{\texorpdfstring{Unbiasedness of $\widehat{\Phi}^{\mathrm{TOK}}$}{Unbiasedness of the Token Presence estimator}}
\label{sec:gt-unbiased}

Let $\mu^{\mathrm{TOK}}(v) = \mathbb{E}_{p}[\Phi^{\mathrm{TOK}}(\mathbf{y};v)]$. Decomposing over the first hitting time of $v$ gives the following mutually exclusive events.
\begin{equation}
    \Phi^{\mathrm{TOK}}(\mathbf{y};v) = \sum_{\ell=1}^{L} \mathbb{I}\{y_\ell = v\} \, \mathbb{I}\{v \notin \mathbf{y}_{<\ell}\}
\end{equation}
Conditioning on $\mathbf{y}_{<\ell}$ and using $\mathbb{E}[\mathbb{I}\{y_\ell=v\}\mid\mathbf{y}_{<\ell}] = p(v\mid\mathbf{y}_{<\ell})$ gives
\begin{equation}
    \mathbb{E}_{p}[\Phi^{\mathrm{TOK}}(\mathbf{y};v)] = \mathbb{E}_{p}\!\left[ \sum_{\ell=1}^{L} p(v \mid \mathbf{y}_{<\ell}) \, \mathbb{I}\{v \notin \mathbf{y}_{<\ell}\} \right] = \mathbb{E}_{p}[\widehat{\Phi}^{\mathrm{TOK}}(\mathbf{y};v)].
\end{equation}
Both $\Phi^{\mathrm{TOK}}(\mathbf{y};v)$ and $\widehat{\Phi}^{\mathrm{TOK}}(\mathbf{y};v)$ are therefore unbiased estimators of $\mu^{\mathrm{TOK}}(v)$.

\subsection{Compositional Token Presence first-completion estimators}
\label{sec:compositional-gt}

The same first-completion argument gives unbiased estimators from next-token probabilities for the Compositional Token Presence events in \S\ref{section:eval_settings}. For the Conjunctive event, let $\mathbf{v}=(v_1,\ldots,v_m)$ be a tuple of distinct target tokens and let
\begin{equation*}
    M_\ell = \{v_1,\ldots,v_m\} \setminus \{y_1,\ldots,y_{\ell-1}\}
\end{equation*}
be the targets still missing immediately before position $\ell$. The event is first completed at $\ell$ exactly when one target remains and that target is emitted. Its reference estimator is therefore
\begin{equation}
\label{eq:and-gt-estimator}
    \widehat{\Phi}^{\mathrm{AND}}(\mathbf{y};\mathbf{v})
    = \sum_{\ell=1}^L
    \mathbb{I}\{|M_\ell|=1\}
    \sum_{u\in M_\ell}p(u\mid\mathbf{y}_{<\ell}).
\end{equation}
Conditioning on $\mathbf{y}_{<\ell}$ replaces the indicator that the final missing token is sampled with its next-token probability. Because the possible first-completion positions are mutually exclusive,
\begin{equation*}
    \mathbb{E}_p[\widehat{\Phi}^{\mathrm{AND}}(\mathbf{y};\mathbf{v})]
    = \mathbb{E}_p[\Phi^{\mathrm{AND}}(\mathbf{y};\mathbf{v})].
\end{equation*}

For the Ordered Sequence event, define the prefix state
\begin{equation*}
    c_\ell = \max\!\left\{d\in\{0,\ldots,m\}: (v_1,\ldots,v_d)
    \text{ is a subsequence of }\mathbf{y}_{<\ell}\right\}.
\end{equation*}
The event is first completed at position $\ell$ exactly when $c_\ell=m-1$ and $y_\ell=v_m$. Thus,
\begin{equation}
\label{eq:sequence-gt-estimator}
    \widehat{\Phi}^{\mathrm{SEQ}}(\mathbf{y};\mathbf{v})
    = \sum_{\ell=1}^L
    \mathbb{I}\{c_\ell=m-1\}
    p(v_m\mid\mathbf{y}_{<\ell}).
\end{equation}
Applying the tower property at every position and summing the mutually exclusive first-completion events gives
\begin{equation*}
    \mathbb{E}_p[\widehat{\Phi}^{\mathrm{SEQ}}(\mathbf{y};\mathbf{v})]
    = \mathbb{E}_p[\Phi^{\mathrm{SEQ}}(\mathbf{y};\mathbf{v})].
\end{equation*}
In both cases, the reference probability is the sample mean of $\widehat{\Phi}$ over ordinary rollouts from the frozen base model. The estimator integrates out the final token draw that completes the event, yielding a much denser reference signal than the raw indicator.

\subsection{Accuracy of the reference probabilities}
\label{sec:gt-precision-target}
\label{sec:naive-gt-cost}
\label{sec:soft-gt-cost}
\label{sec:profanity-gt-cost}

For $N_{\mathrm{gt}}$ independent base-model rollouts, let $Z_i$ be the
unbiased first-completion estimate for token-based events or the binary
indicator for the Profanity Classifier.
We compute the reference probability and its estimated relative standard error as
\[
\widehat\mu_{\mathrm{gt}}=\frac{1}{N_{\mathrm{gt}}}\sum_{i=1}^{N_{\mathrm{gt}}}Z_i,
\qquad
\widehat{\mathrm{RSE}}=\frac{s_Z}{\sqrt{N_{\mathrm{gt}}}\,\widehat\mu_{\mathrm{gt}}},
\]
where $s_Z$ is the sample standard deviation of the $Z_i$.
For naive MC, the population relative standard error is
$\sqrt{(1-\mu)/(N_{\mathrm{gt}}\mu)}$, so $10\%$ precision at
$\mu=10^{-9}$ requires approximately $10^{11}$ rollouts.

We use $1{,}024{,}000$ reference rollouts per model and length for Token
Presence and $512{,}000$ for compositional and profanity events.
These budgets are fixed across events.
Table~\ref{tab:reference-accuracy} reports the rarest event in each listed setting.

\begin{table}[H]
\centering
\caption{Reference probability and relative standard error for the rarest reported event in each listed setting.}
\label{tab:reference-accuracy}
\scriptsize
\setlength{\tabcolsep}{3pt}
\begin{tabular}{@{}llrrr@{}}
\hline
\textbf{Setting}
& \textbf{Rarest reported event}
& \shortstack{\textbf{Reference}\\\textbf{samples}}
& \shortstack{\textbf{Reference}\\\textbf{probability}}
& \shortstack{\textbf{Relative}\\\textbf{SE}} \\
\hline
GPT-2 Small token, $L=50$
& \texttt{ThumbnailImage}
& $1{,}024{,}000$
& $3.97\times10^{-9}$
& $0.74\%$ \\
Gemma-2B token, $L=10$
& \texttt{meio}
& $1{,}024{,}000$
& $1.08\times10^{-9}$
& $1.32\%$ \\
Ordered Sequence event, $L=10$
& \texttt{the $\rightarrow$ was $\rightarrow$ connect}
& $512{,}000$
& $1.25\times10^{-6}$
& $7.16\%$ \\
Conjunctive event, $L=10$
& \{\texttt{the}, \texttt{was}, \texttt{connect}\}
& $512{,}000$
& $3.72\times10^{-6}$
& $9.17\%$ \\
GPT-2 Small profanity, $L=10$
& classifier threshold $0.999$
& $512{,}000$
& $8.59\times10^{-5}$
& $15.07\%$ \\
GPT-2 Small profanity, $L=50$
& classifier threshold $0.999$
& $512{,}000$
& $1.44\times10^{-3}$
& $3.68\%$ \\
Gemma-2B profanity, $L=10$
& classifier threshold $0.999$
& $512{,}000$
& $6.25\times10^{-5}$
& $17.68\%$ \\
\hline
\end{tabular}
\end{table}

\section{Related methods}
\label{app:method-relationships}

\subsection{Cross-entropy baseline selection}
\label{app:relationship-ce}

We evaluated two CE formulations: importance-weighted CE for rare-event estimation and unweighted CE for optimization. Both rank sequences using $G_\delta(\mathbf{y})=S_\delta(\mathbf{y})$ for Token Presence and Compositional Token Presence, and $G_\delta(\mathbf{y})=\log\!\bigl[f(\mathbf{y})/(1-f(\mathbf{y}))\bigr]$ for the Profanity Classifier, the logit of its predicted probability. We write $\operatorname{Top}_a$ for the indices of the top $a$ fraction of scores.

\paragraph{Importance-weighted CE.}
Classical rare-event CE maximizes elite-sequence likelihood with weights proportional to $p(\mathbf{y})/q_{\delta_t}(\mathbf{y})$~\cite{deboer2005tutorial,rubinstein2004cross}. We approximately maximize this objective by gradient ascent, holding the sampled sequences, elite membership, and weights fixed (Algorithm~\ref{alg:ce-likelihood-fitting}). With LoRA, these updates provided insufficient event amplification (Table~\ref{tab:ce-likelihood-fitting-failures}); \texttt{atten} and \texttt{iminary} each appeared only once in $64{,}000$ fresh samples. Activation and logit steering also failed to find \texttt{ActionCode} at $L=30$ ($\mu=2.812\times10^{-9}$), with zero hits in $45{,}056$ and $98{,}304$ training trajectories, respectively.

\begin{algorithm}[!htb]
\caption{Importance-weighted CE with gradient updates}
\label{alg:ce-likelihood-fitting}
\begin{algorithmic}[1]
\setlength{\itemsep}{2pt}
\Statex \textbf{Inputs:} $p,q_\delta,\Phi,G_\delta$; $T,B,\eta$; $\rho,h^\star$.
\State $\delta_1\leftarrow\mathbf{0}$
\For{$t=1,2,\ldots,T$}
    \State Sample $\mathbf{y}_1,\ldots,\mathbf{y}_B\simiid q_{\delta_t}$ \Comment{Fixed sampled trajectories}
    \State $\hat h_t\leftarrow B^{-1}\sum_{i=1}^B\Phi(\mathbf{y}_i)$ \Comment{Update event proportion}
    \If{$\hat h_t\geq h^\star$}
        \State \Return $q_{\delta_t}$ \Comment{Event-proportion target reached}
    \EndIf
    \State $\mathcal{I}_t\leftarrow\operatorname{Top}_\rho\!\bigl(\{G_{\delta_t}(\mathbf{y}_i)\}_{i=1}^B\bigr)$ \Comment{Select elite trajectories}
    \State $w_i\leftarrow p(\mathbf{y}_i)/q_{\delta_t}(\mathbf{y}_i)$, $\bar w_i\leftarrow w_i/\sum_{j\in\mathcal{I}_t}w_j$ for $i\in\mathcal{I}_t$
    \State $F_t(\delta)\leftarrow\sum_{i\in\mathcal{I}_t}\bar w_i\log q_\delta(\mathbf{y}_i)$ \Comment{Fixed elites and weights}
    \State $\delta_{t+1}\leftarrow\delta_t+\eta\nabla_\delta F_t(\delta_t)$ \Comment{Gradient ascent}
\EndFor
\State \Return $q_{\delta_{T+1}}$ \Comment{Proposal for IS estimation}
\end{algorithmic}
\end{algorithm}

\begin{table}[H]
    \centering
    \small
    \begin{tabular}{@{}lrrr@{}}
        \toprule
        Event & Reference value $\mu$ & Event hits & Amplification \\
        \midrule
        \texttt{\textvisiblespace mere} & $6.1594\times10^{-4}$ & 64 & $1.62\times$ \\
        \texttt{atten} & $1.1463\times10^{-5}$ & 1 & $1.36\times$ \\
        \texttt{iminary} & $1.7840\times10^{-6}$ & 1 & $8.76\times$ \\
        \bottomrule
    \end{tabular}
    \caption{Importance-weighted CE with LoRA gradient updates on GPT-2 Small at $L=10$. Each event uses one training seed and $300$ rounds of $128$ trajectories, with elite fraction $0.15$ and one gradient step per round at learning rate $10^{-3}$. Event hits are counted over $64{,}000$ fresh samples from each frozen proposal. Amplification is the observed event frequency divided by $\mu$.}
    \label{tab:ce-likelihood-fitting-failures}
\end{table}

\begin{figure}[p]
    \centering
    \includegraphics[width=0.49\textwidth]{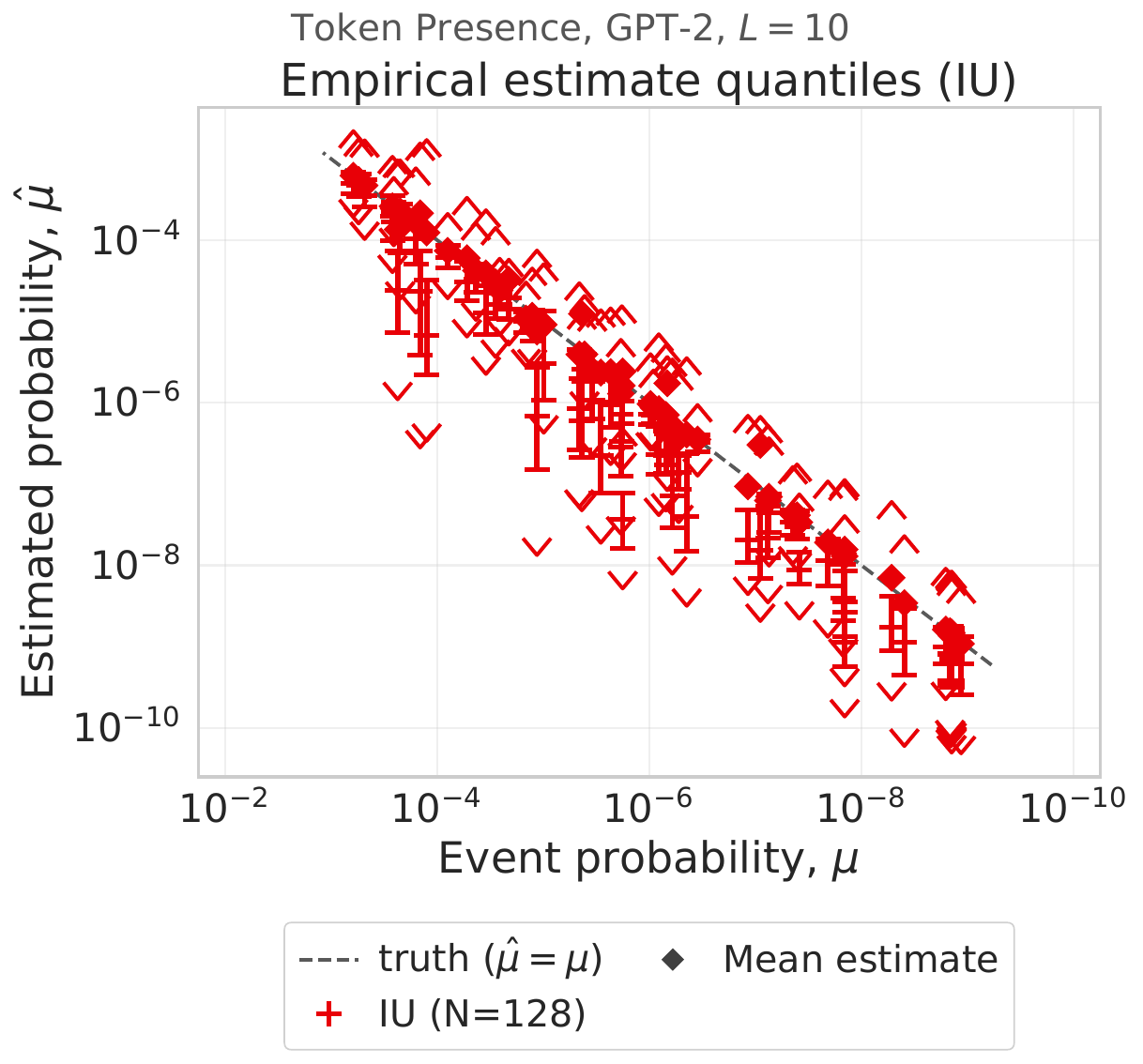}
    \hfill
    \includegraphics[width=0.49\textwidth]{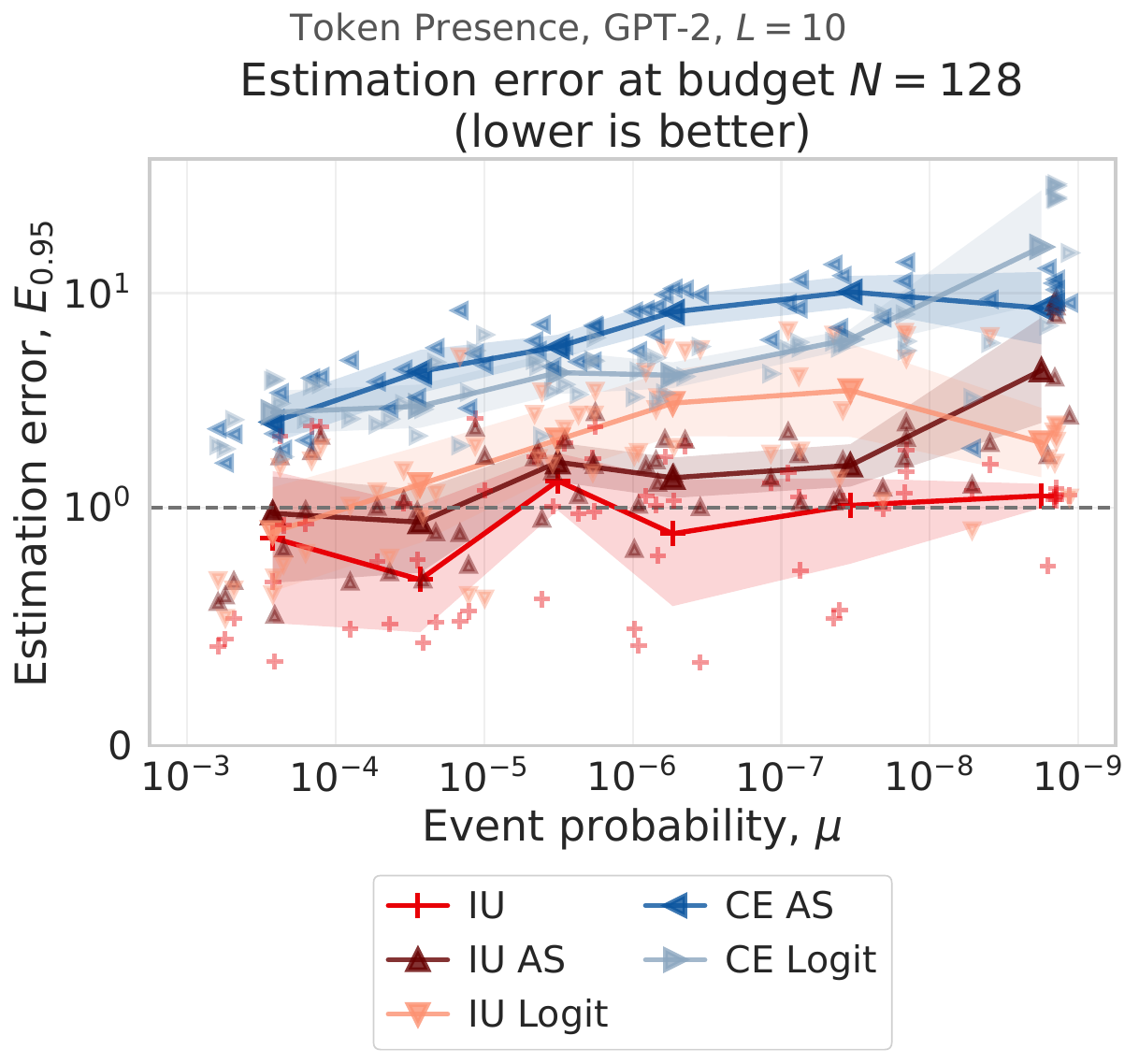}

    \includegraphics[width=0.49\textwidth]{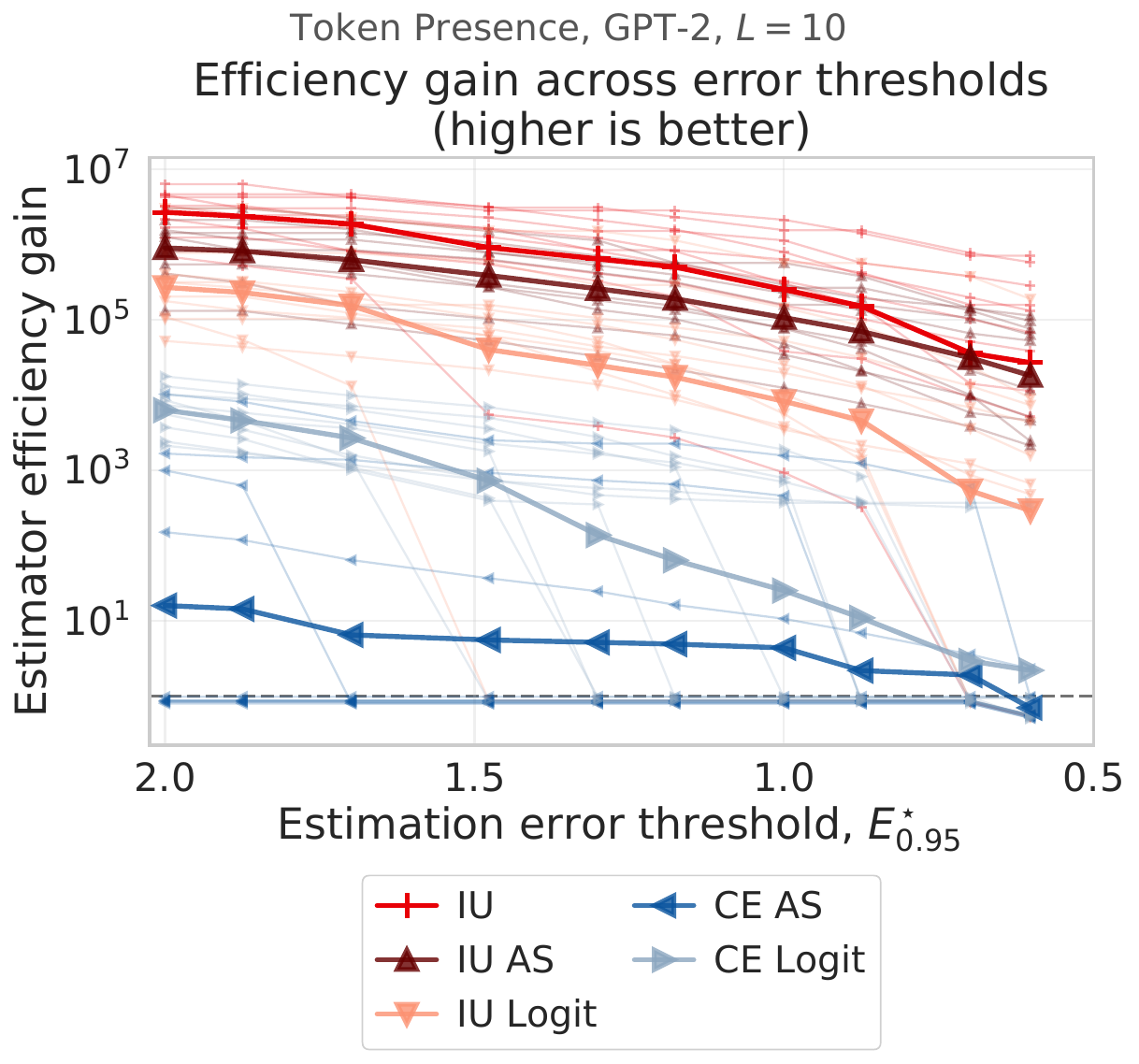}
    \hfill
    \includegraphics[width=0.49\textwidth]{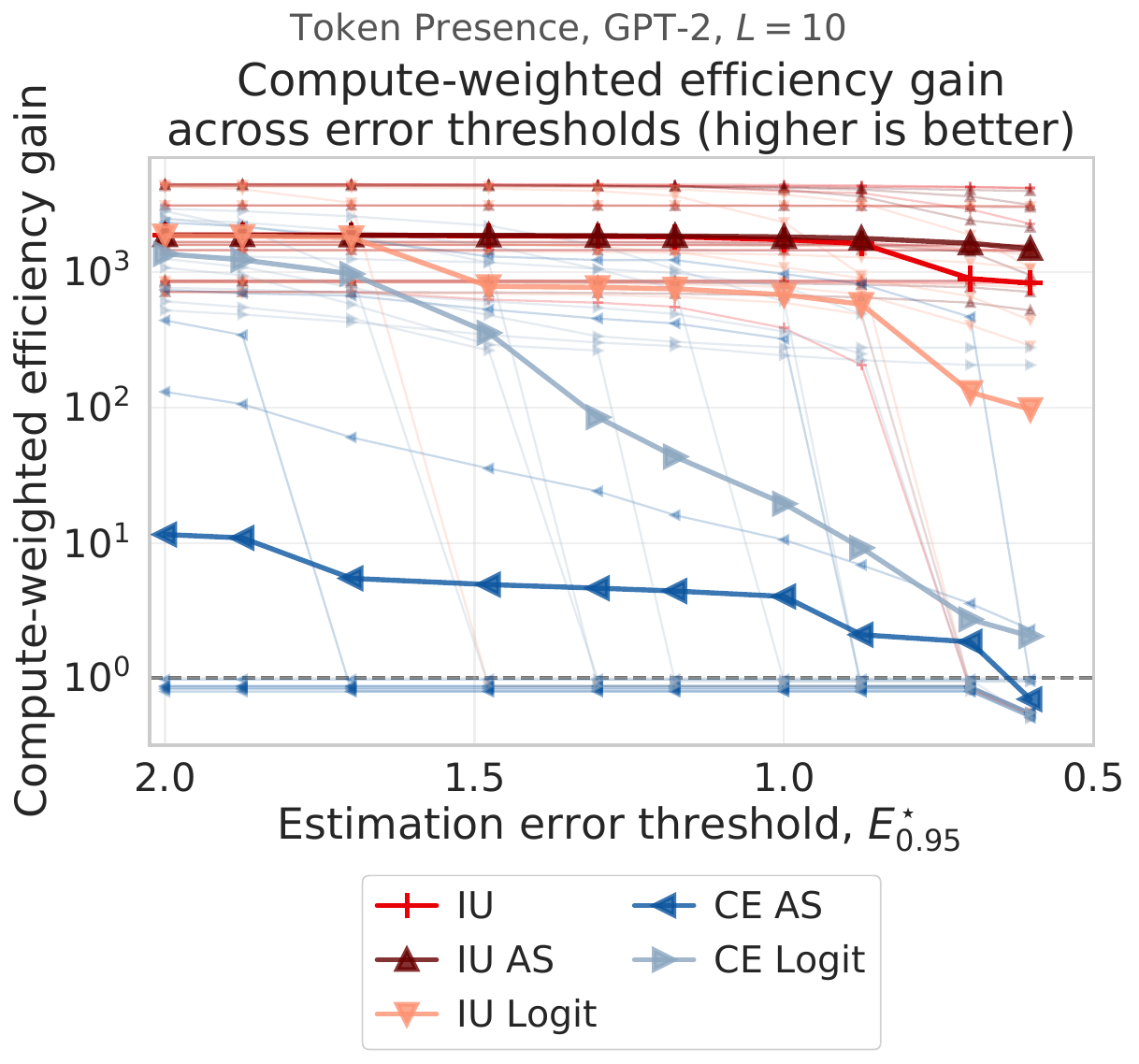}
    \caption{Full-detail Token Presence results on GPT-2 Small at $L=10$: IU empirical quantiles ($N = 128$), all-method error at $N=128$, and all-method estimator and compute-weighted efficiency frontiers.}
    \label{fig:full-token-length-10}
\end{figure}

\paragraph{Unweighted CE optimization.}
In our unweighted LoRA likelihood trials, we instead took gradient steps on the unweighted elite-sequence log likelihood; these updates concentrated the proposal too strongly on the observed elite sequences, resulting in severe underestimation. The standard CE optimization search performed better empirically. This version of the algorithm samples steering vectors, ranks them by the scores of their generated sequences, and moves the search mean toward the unweighted average of the elite vectors~\cite{rubinstein2004cross,botev2013cross} (Algorithm~\ref{alg:ce}). We retain this update for activation steering (CE AS) and logit steering (CE Logit). At $L=10$, it achieves substantial estimator and compute-weighted efficiency gains over naive Monte Carlo at some accuracy thresholds (Figure~\ref{fig:full-token-length-10}).

\begin{algorithm}[!htb]
\caption{Unweighted CE optimization with activation/logit tilting}
\label{alg:ce}
\begin{algorithmic}[1]
\setlength{\itemsep}{2pt}
\Statex \textbf{Inputs:} $p,q_\delta,G_\delta$; $T,B$; $\gamma_{\mathrm{elite}},\sigma_{\mathrm{CE}}^2$; $\tau_{\mathrm{CE}}\in(0,1]$.
\State $\mathbf{m}_1\leftarrow\mathbf{0}$
\For{$t=1,2,\ldots,T$}
    \State Sample $\delta^{(1)},\ldots,\delta^{(B)}\simiid\mathcal{N}(\mathbf{m}_t,\sigma_{\mathrm{CE}}^2\mathbf{I})$ \Comment{Steering vectors}
    \State Sample $\mathbf{y}_i\sim q_{\delta^{(i)}}$ for $i=1,\ldots,B$ \Comment{One sequence per vector}
    \State $\mathcal{I}_t\leftarrow\operatorname{Top}_{\gamma_{\mathrm{elite}}}\!\bigl(\{G_{\delta^{(i)}}(\mathbf{y}_i)\}_{i=1}^B\bigr)$ \Comment{Select elite perturbations}
    \State $\mathbf{m}_{t,\mathrm{elite}}\leftarrow|\mathcal{I}_t|^{-1}\sum_{i\in\mathcal{I}_t}\delta^{(i)}$ \Comment{Elite mean}
    \State $\mathbf{m}_{t+1}\leftarrow(1-\tau_{\mathrm{CE}})\mathbf{m}_t+\tau_{\mathrm{CE}}\mathbf{m}_{t,\mathrm{elite}}$ \Comment{Update search mean}
\EndFor
\State \Return $\mathbf{m}_{T+1}$ \Comment{Learned perturbation}
\end{algorithmic}
\end{algorithm}

\subsection{Twisted sequential Monte Carlo}
\label{app:relationship-tsmc}
\label{app:other-baselines}

Twisted sequential Monte Carlo (TSMC) uses intermediate scores to guide particles toward promising continuations. These scores can be supplied by learned twist functions~\cite{zhao2024probabilistic}. In our implementation, we sample short look-ahead continuations from each current prefix and compute the target token's log-probability at each step, whether or not that token is sampled. We average the resulting surrogate scores to guide resampling toward higher-scoring prefixes, keeping the LM's weights fixed. IU instead uses surrogate gradients to train a proposal that changes the token conditionals throughout generation.

We evaluated a TSMC variant using our surrogate $S_\delta(\mathbf{y})$ as the resampling score with a two-step look-ahead, averaged across 20 look-ahead particles. Across 10 token events with probabilities between $10^{-6}$ and $10^{-9}$, this variant yielded an average efficiency ratio of $0.87\times$ relative to naive Monte Carlo. It did not consistently improve upon naive Monte Carlo, so we did not retain it in the main baseline comparisons.

In these runs, repeated resampling concentrated particles onto a small number of high-scoring prefixes, reducing diversity among downstream trajectories. This behavior highlights the need to preserve diverse event paths while directing sampling toward the event, which motivates the combination of amplification and regularization in IU.

\section{Additional considerations in methodology}
\label{app:additional-methodology}

\subsection{Surrogate coverage}
\label{app:surrogate-coverage}

We examine surrogate mismatch using a fixed GPT-2 Small proposal trained to amplify the token \texttt{\textvisiblespace connect}. We evaluate this proposal on five events. The central event requires this token alone. Adding alternative tokens through an OR condition expands the event beyond the condition targeted by the surrogate. Adding required tokens through an AND condition instead makes \texttt{\textvisiblespace connect} a necessary but insufficient condition for the event. This latter case provides a simple instance of the necessary-condition surrogate discussed in Section~\ref{sec:surrogate}.

\begin{figure}[!htb]
    \centering
    \includegraphics[width=0.85\linewidth]{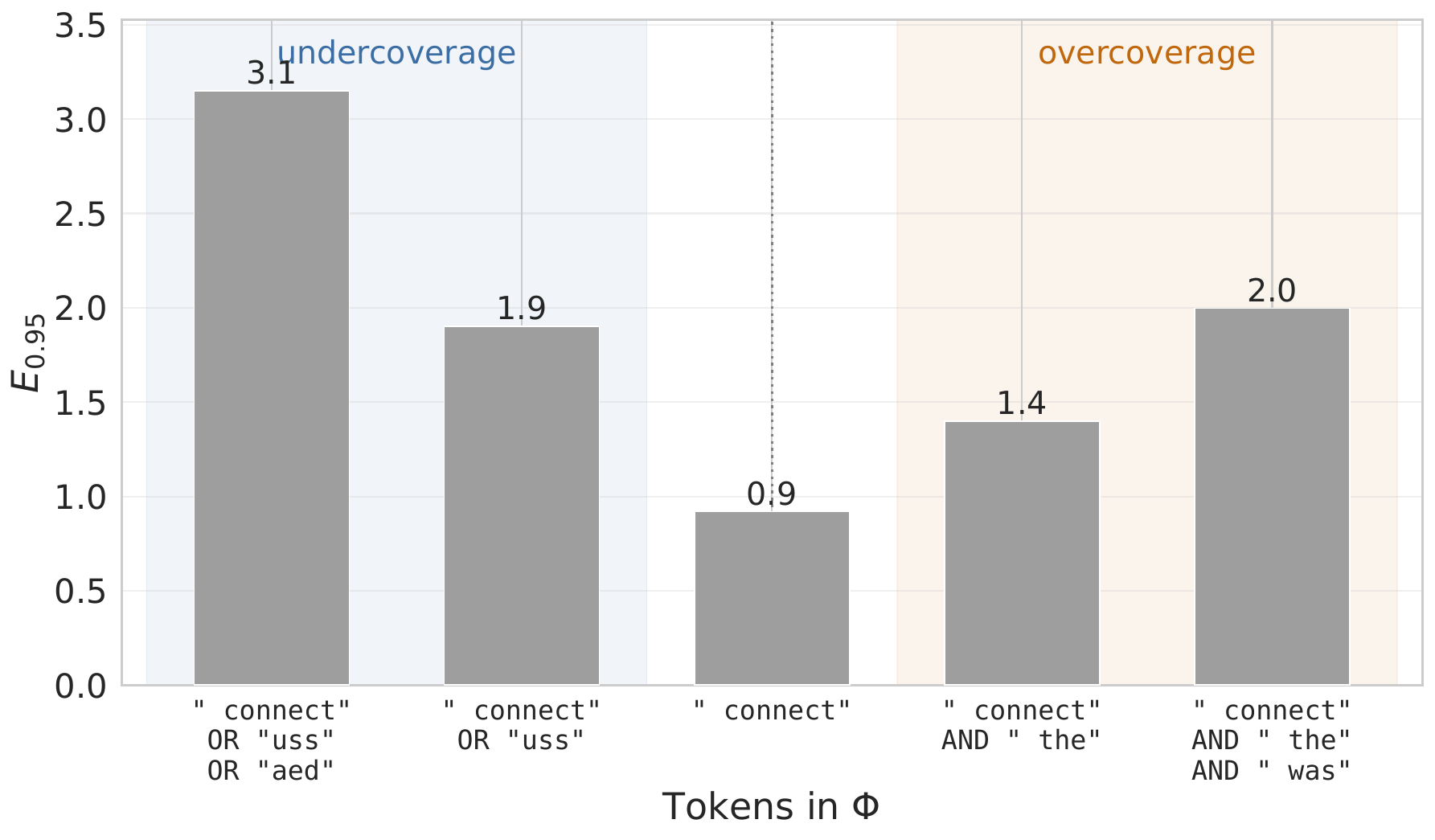}
    \caption{\textbf{Estimation error under surrogate mismatch.} A fixed proposal trained to amplify \texttt{\textvisiblespace connect} is evaluated on the five token-presence events shown. OR conditions (left) include event trajectories without this token; AND conditions (right) require additional tokens that the surrogate does not target. The central event matches the surrogate's target. Lower error is better. Undercoverage and overcoverage refer to the surrogate's target condition relative to the event.}
    \label{fig:surrogate-coverage-experiment}
\end{figure}

Figure~\ref{fig:surrogate-coverage-experiment} shows the lowest error when the event matches the token targeted by the surrogate. Error increases as either alternative or additional required tokens are introduced. These preliminary results suggest that a surrogate can remain useful when it targets a broader condition than the event, with a loss of estimation accuracy as the mismatch grows. A systematic study of this tradeoff for more complex events remains open.

\subsection{Alternative regularizers}
\label{app:regularizer-alternatives}
\label{app:regularizer-sampling}

We also considered three alternatives to the regularizer in Eq.~\eqref{eq:reg-final}.

\begin{figure}[H]
    \centering
    \includegraphics[width=\linewidth]{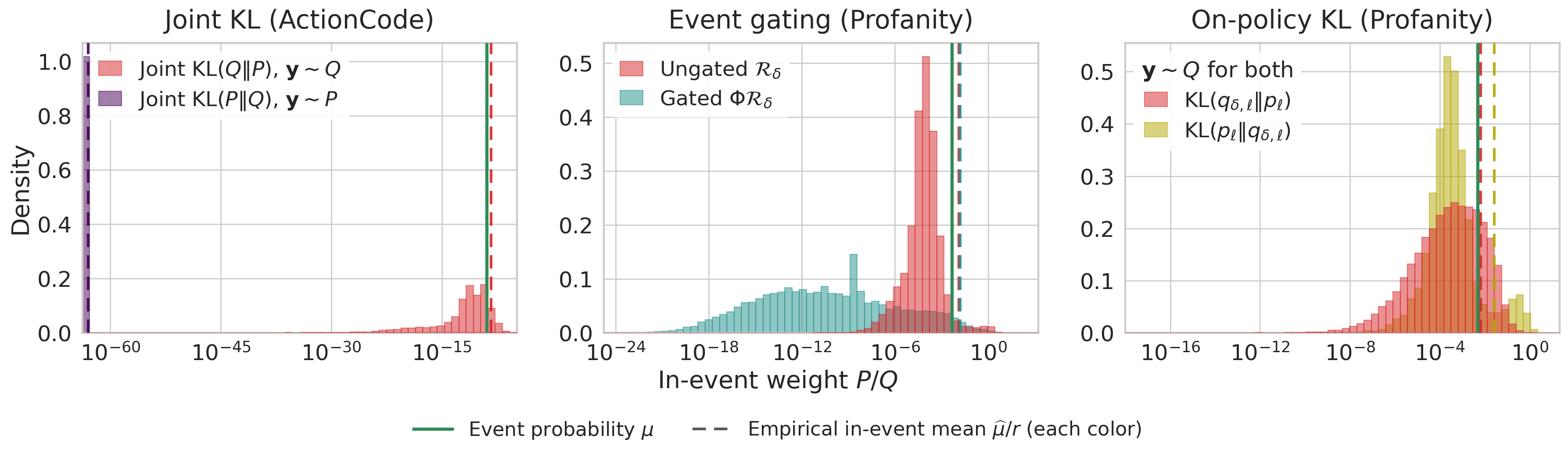}
    \caption{Alternative regularizers, compared through in-event importance weights $p/q_\delta$ at $L=10$. Red denotes our choice in each panel. \textbf{Left:} joint reverse KL on proposal rollouts and joint forward KL on base-model rollouts for \texttt{ActionCode}. \textbf{Center:} regularization on every proposal rollout and regularization gated by $\Phi$. \textbf{Right:} reverse and forward next-token KL, both evaluated on proposal rollouts. The center and right panels use the Profanity Classifier. The green line marks $\mu$, the weight under the zero-variance proposal; colored dashed lines mark each proposal's empirical mean in-event weight.}
    \label{fig:reg-geometry}
\end{figure}

\paragraph{Joint forward KL.}
For the \texttt{ActionCode} event, regularization based on base-model rollouts failed to control concentration within the event (Figure~\ref{fig:reg-geometry}, left). This alternative uses the joint forward divergence
\begin{equation}
\label{eq:joint-forward-kl}
\mathrm{KL}(p\Vert q_\delta)
=\mathbb{E}_{\mathbf{y}\sim p}
\left[\log\frac{p(\mathbf{y})}{q_\delta(\mathbf{y})}\right]
=\mathbb{E}_{\mathbf{y}\sim p}
\left[\sum_{\ell=1}^L
\mathrm{KL}\!\left(p(\cdot\mid\mathbf{y}_{<\ell})\Vert q_\delta(\cdot\mid\mathbf{y}_{<\ell})\right)\right].
\end{equation}
At practical training budgets, rollouts from $p$ almost exclusively contain non-event trajectories. A batch of $B$ such rollouts misses the event with probability $(1-\mu)^B$, so the penalty receives little direct feedback about distortion within $\mathcal{E}$. This causes the proposal to perform poorly.

\paragraph{Restricting regularization to the event.}
Gating the penalty by the event indicator, using $\Phi(\mathbf{y})\mathcal{R}_\delta(\mathbf{y})$ on proposal rollouts, also degraded IU's performance. When a batch contains no event hits, this penalty is zero on every trajectory, so the surrogate update receives no regularization in early steps. Applying $\mathcal{R}_\delta$ to every rollout constrains these early updates before event hits become frequent (Figure~\ref{fig:reg-geometry}, center).

\paragraph{On-policy forward KL.}
\label{app:relationship-on-policy-distillation}
Finally, we kept proposal rollouts and reversed the divergence between the next-token distributions,
\begin{equation}
\label{eq:on-policy-forward-kl}
\mathcal{D}_{\mathrm{on}}(\delta)
=\mathbb{E}_{\mathbf{y}\sim q_\delta}
\left[\sum_{\ell=1}^L
\mathrm{KL}\!\left(p(\cdot\mid\mathbf{y}_{<\ell})\Vert q_\delta(\cdot\mid\mathbf{y}_{<\ell})\right)\right].
\end{equation}
The outer expectation follows $q_\delta$, while the inner KL weights tokens under $p$. This is the forward-KL objective used in on-policy distillation, where a student learns from teacher feedback on its own generated sequences~\cite{agarwal2024onpolicy}. It differs from joint forward KL in Eq.~\eqref{eq:joint-forward-kl}, whose prefixes follow $p$. Both on-policy directions provide feedback across the full vocabulary at each visited prefix. Neither directly compares conditionals at unvisited prefixes.

The two variants gave similar pooled probability estimates in this comparison ($0.98\mu$ for reverse KL and $0.94\mu$ for forward KL), despite different event frequencies and weight distributions (Figure~\ref{fig:reg-geometry}, right). The preferred direction remains problem specific, consistent with the findings of Agarwal et al.~\cite{agarwal2024onpolicy}.

\subsection{Per-step chi-square and importance-weight variance}
\label{app:per-step-chi-square}

A related alternative to the reverse-KL penalty is an unweighted sum of per-step chi-square divergences:
\begin{equation}
\label{eq:per-step-chi-square-population}
\mathcal{D}_{\chi}(q)
:=\mathbb{E}_{\mathbf{Y}\sim q}\left[
\sum_{\ell=1}^L\chi^2(p_\ell\Vert q_\ell)\right].
\end{equation}

The exact joint chi-square, $\chi^2(p\Vert q) = \mathbb{E}_{q}[\sum_{\ell=1}^L W_{\ell-1}^{2}\chi^2(p_\ell\Vert q_\ell)]$, scales these local terms by the importance weights of the prefixes, $W_{\ell-1}^2$. Since these weights multiply out over the sampled tokens, they can vary sharply between trajectories and introduce severe instability into the penalty itself. By dropping the prefix weights, the unweighted sum avoids this compounding variance while retaining dense optimization feedback at every visited prefix.

This unweighted surrogate remains sound for controlling the proposal divergence. With a finite vocabulary $\mathcal{V}$ and fixed generation length $L$, the unweighted penalty $\mathcal{D}_{\chi}(q)$ is zero if and only if $q=p$. More generally, for proposals $q_n$ satisfying the same support assumption, driving the unweighted penalty to zero guarantees convergence of the true joint divergence:
\begin{equation}
\label{eq:per-step-chi-square-zero-limit}
\mathcal{D}_{\chi}(q_n)\longrightarrow0
\quad\Longrightarrow\quad
\chi^2(p\Vert q_n)\longrightarrow0.
\end{equation}

\subsection{Further fine-tuning the proposal}
While IU employs a surrogate to provide a continuous gradient for the base proposal distribution, it is natural to then ask whether the proposal can be further improved after the initial IU optimization. Once a high-coverage proposal $q_\delta$ is established, it triggers the rare event frequently enough that the exact indicator $\Phi(\mathbf{y})$  provides a usable signal. At this point, one could discard the surrogate and fine-tune the proposal further using other techniques. Exploring these subsequent refinement steps is an interesting direction, but it lies outside the scope of this work. We view such fine-tuning as naturally building upon the initial high-coverage proposal that IU provides.

\subsection{Input context and autoregressive output length}
\label{app:input-length}

Input context and generated output play different roles in IU. Conditional on a fixed input $\mathbf{x}$, the trajectory importance weight is a product of $L$ token-level likelihood ratios, one for each generated token. Increasing the output length therefore adds stochastic decoding steps over which discrepancies between the proposal and base model can accumulate. A longer fixed input supplies conditioning information but does not itself add factors to this product. The output length thus determines the size of the sequence space over which IU constructs its proposal.

\clearpage
\section{Absolute budgets for the main-text efficiency plots}
\label{app:absolute-budgets}

\begin{figure}[!htbp]
    \centering
    \includegraphics[width=0.74\textwidth]{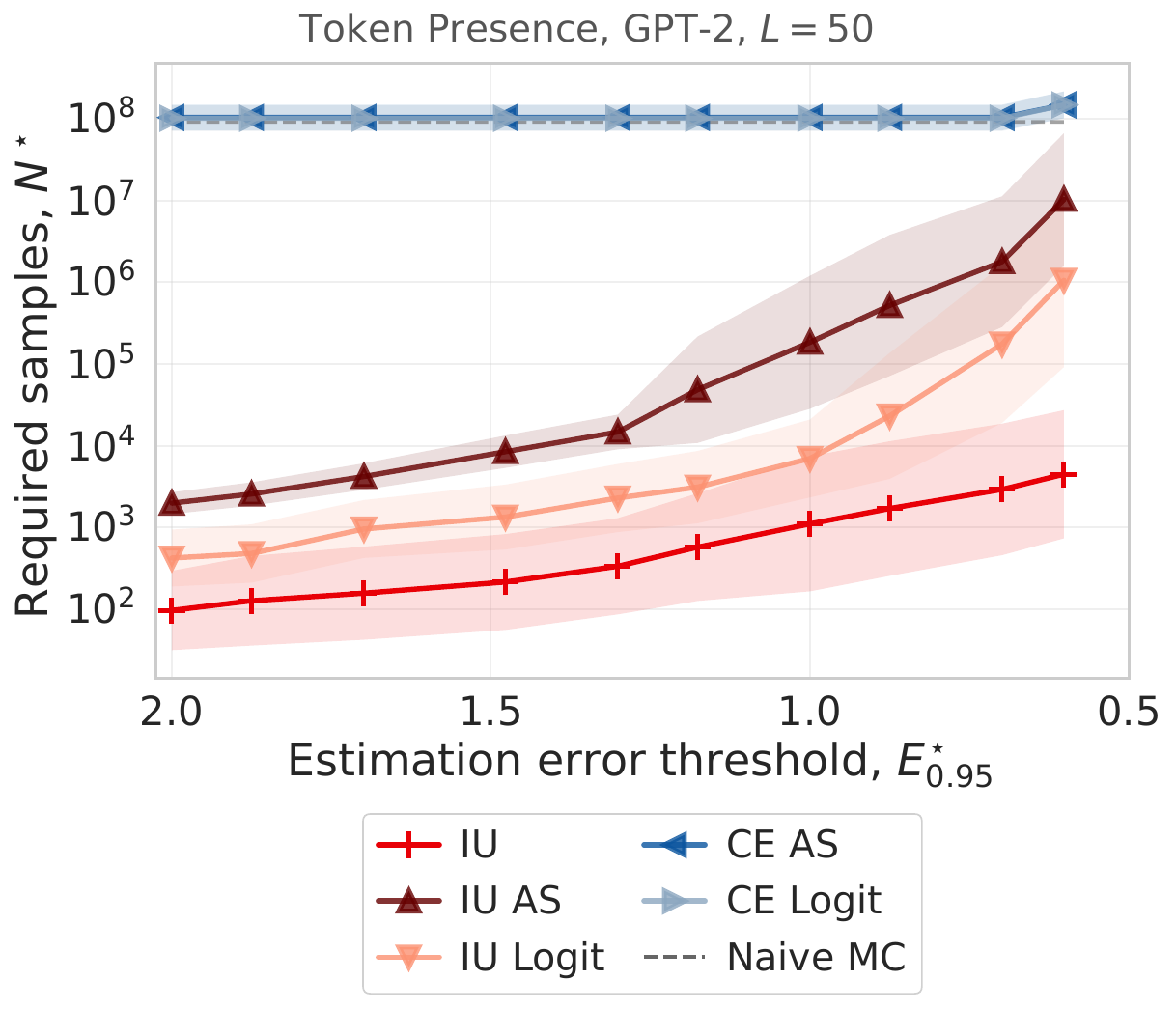}
    \caption{Absolute sampling budgets for the Token Presence method comparison in Figure~\ref{fig:our_method_better} (left), on GPT-2 Small at $L=50$. Each curve summarizes the same ten events in $10^{-8}\leq\mu<10^{-7}$. At $E_{0.95}^\star=1$, geometric-mean IU and naive-MC budgets are approximately $1{,}103$ and $9.24\times10^7$ trajectories, respectively.}
    \label{fig:absolute-token-methods}
\end{figure}

\begin{figure}[p]
    \centering
    \includegraphics[width=\textwidth]{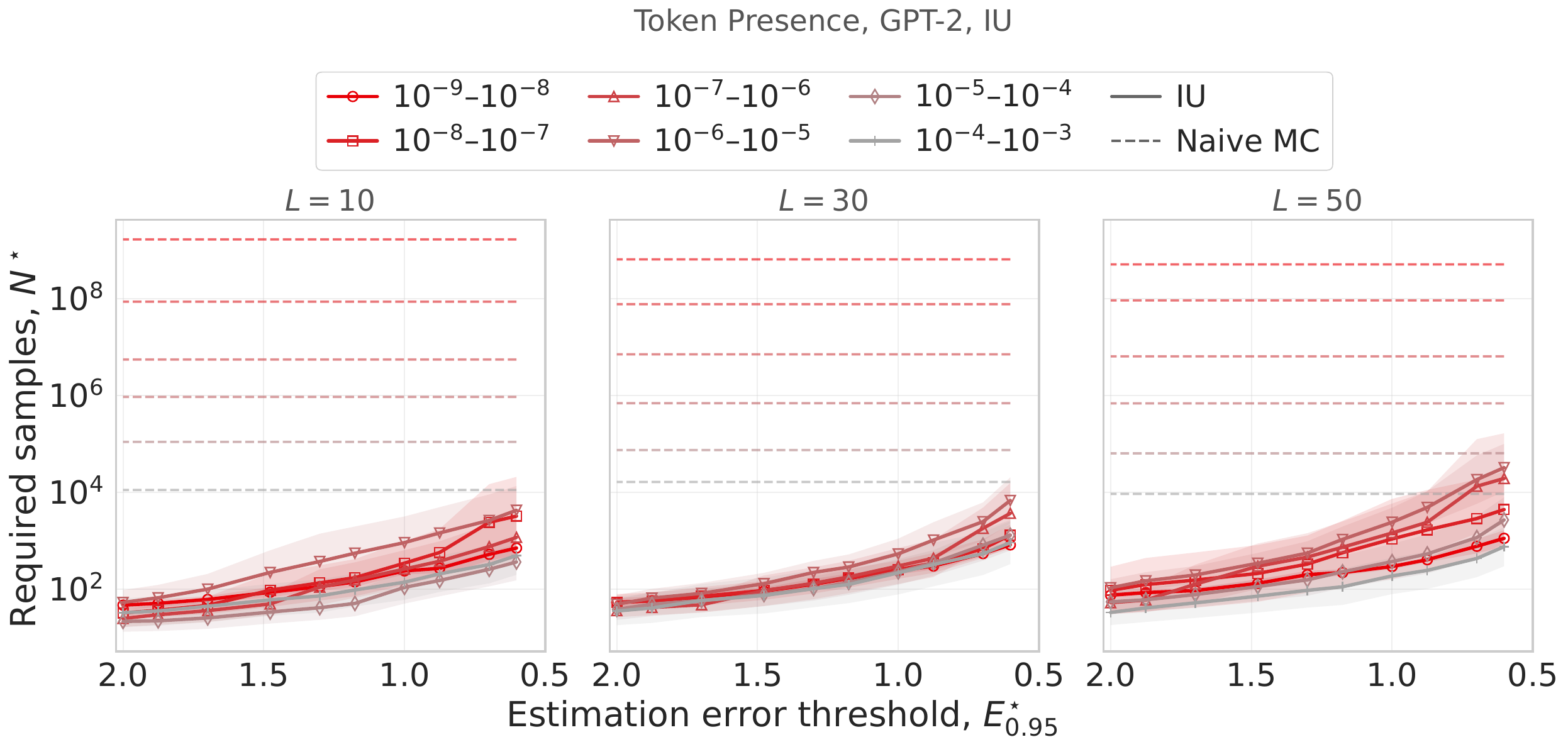}
    \caption{Absolute sampling budgets for the sequence-length comparison in Figure~\ref{fig:length}, with $L=10$, $30$, and $50$ from left to right. Each color denotes a rarity band with ten events; solid curves show IU and dashed curves show naive MC. At $E_{0.95}^\star=1$ in $10^{-8}\leq\mu<10^{-7}$, geometric-mean IU versus naive-MC budgets are approximately $349$ versus $8.73\times10^7$, $255$ versus $7.73\times10^7$, and $1{,}103$ versus $9.24\times10^7$ trajectories, respectively.}
    \label{fig:absolute-token-lengths}
\end{figure}

\begin{figure}[p]
    \centering
    \includegraphics[width=0.74\textwidth]{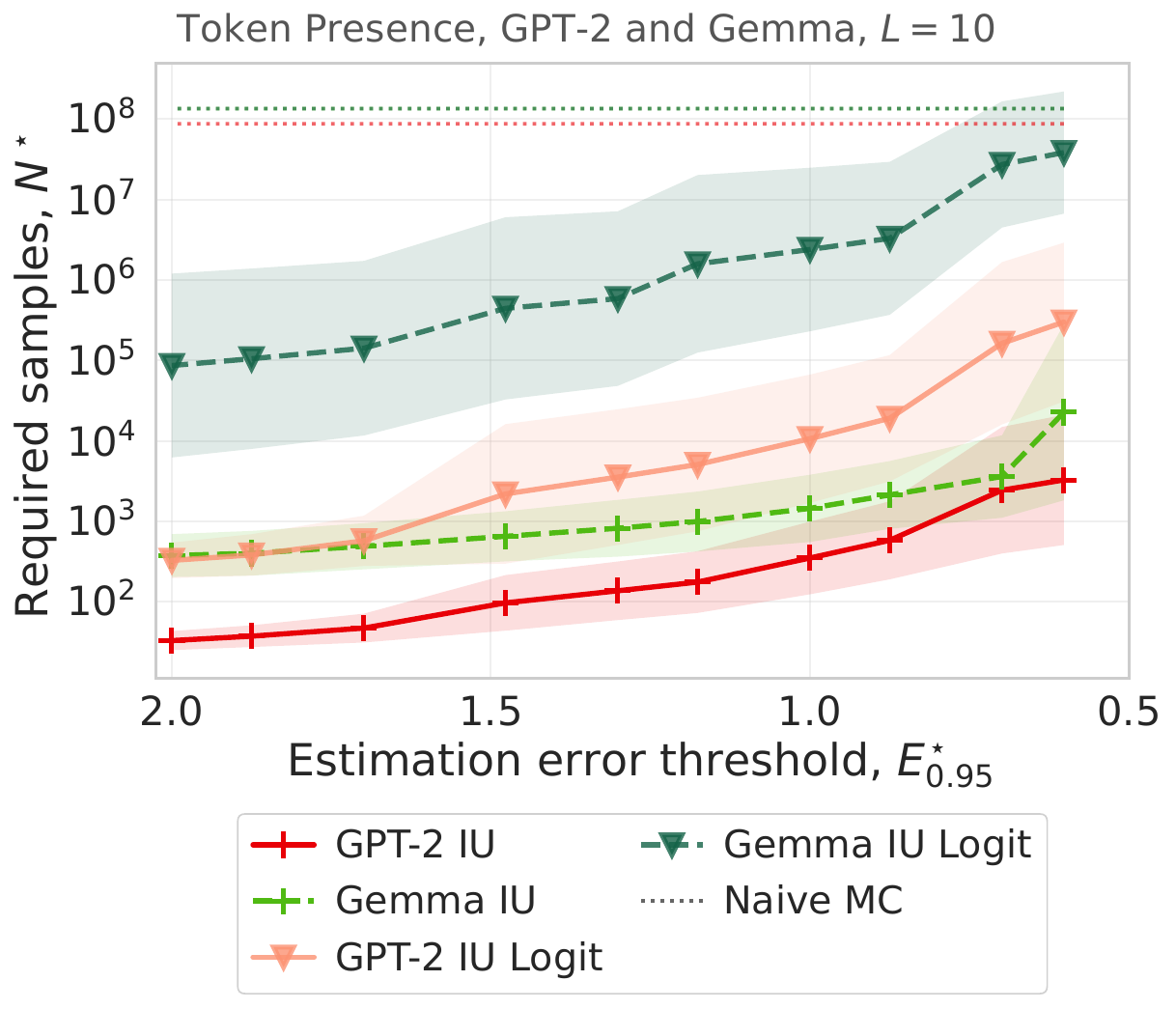}
    \caption{Absolute sampling budgets for the Token Presence model comparison in Figure~\ref{fig:modelsize} (left), at $L=10$. Each curve summarizes the same ten events per model in $10^{-8}\leq\mu<10^{-7}$. At $E_{0.95}^\star=1$, geometric-mean IU versus naive-MC budgets are approximately $349$ versus $8.73\times10^7$ trajectories on GPT-2 and $1{,}445$ versus $1.35\times10^8$ on Gemma-2.}
    \label{fig:absolute-token-models}
\end{figure}

\begin{figure}[p]
    \centering
    \includegraphics[width=0.74\textwidth]{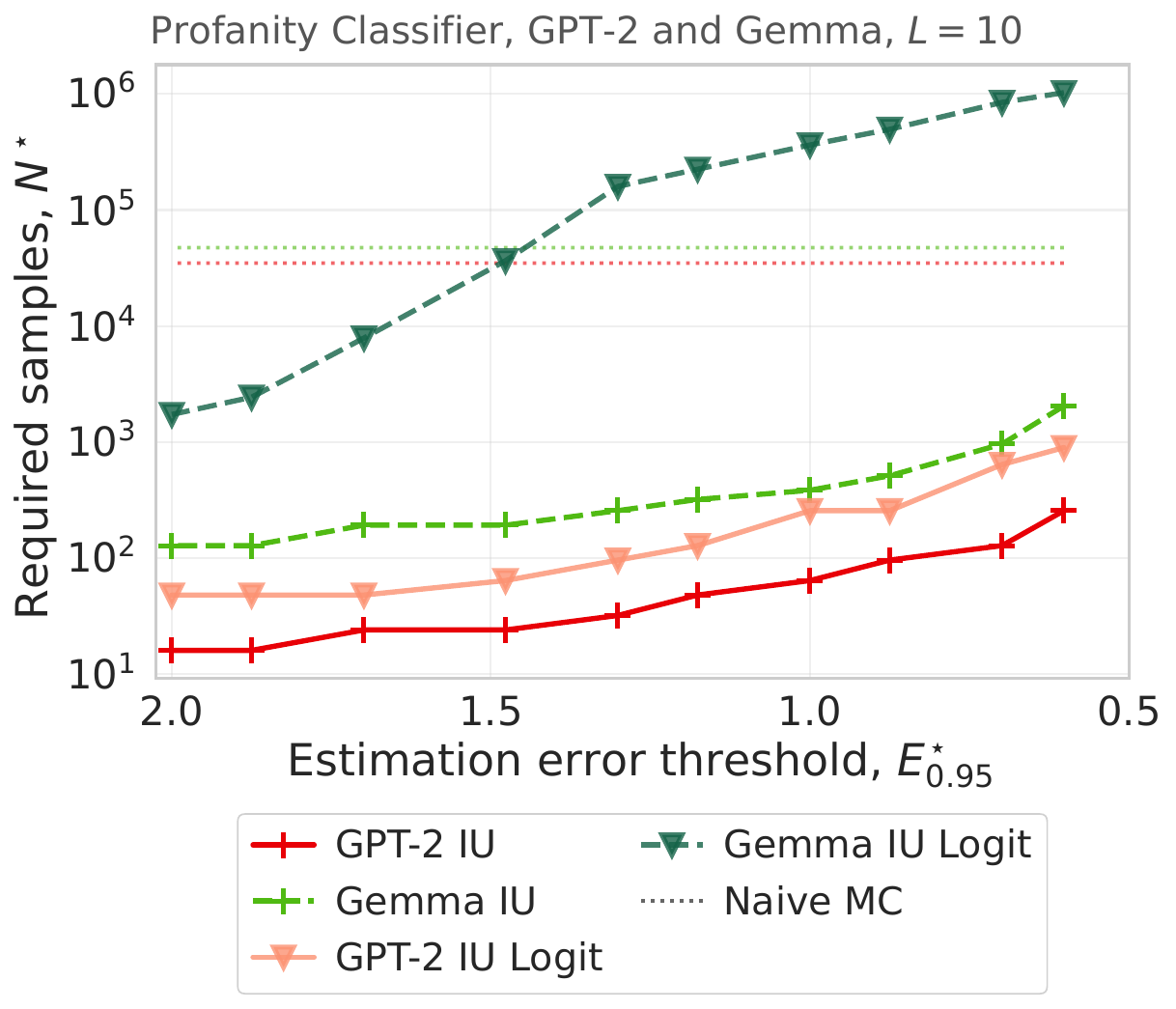}
    \caption{Absolute sampling budgets for the Profanity Classifier model comparison in Figure~\ref{fig:profanity} (left), at $L=10$ and $\kappa=0.999$. At $E_{0.95}^\star=1$, IU versus naive-MC budgets are $64$ versus $34{,}858$ trajectories on GPT-2 and $384$ versus $47{,}311$ on Gemma-2.}
    \label{fig:absolute-profanity-models}
\end{figure}

\begin{figure}[p]
    \centering
    \includegraphics[width=\textwidth]{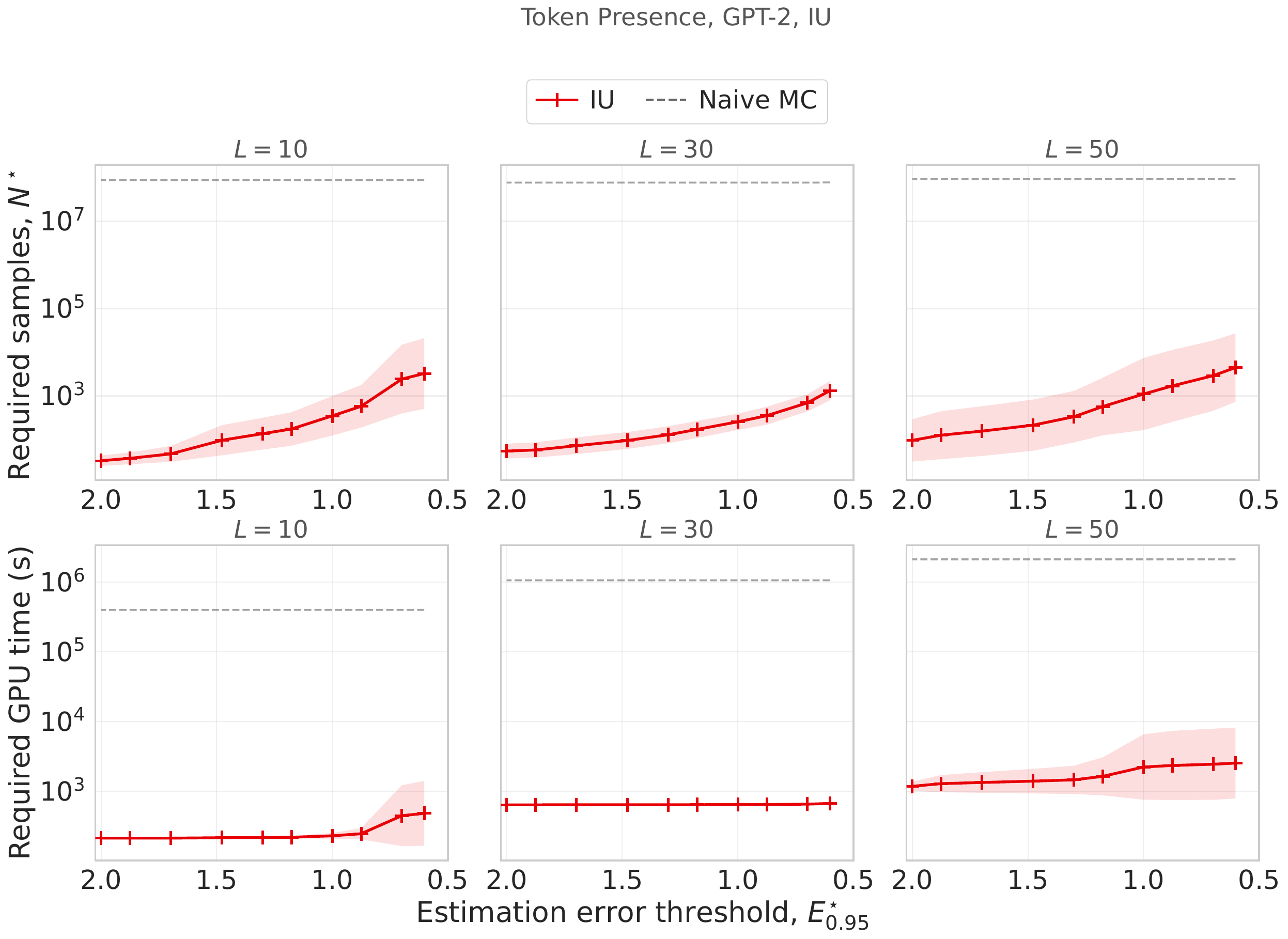}
    \caption{Absolute budgets underlying the compute-weighted efficiency comparison in Figure~\ref{fig:computational_efficiency}, on GPT-2 Small at $L=10$, $30$, and $50$ from left to right. Top: required trajectories. Bottom: estimated GPU time including IU training. Each curve summarizes the same ten events per length in $10^{-8}\leq\mu<10^{-7}$. At $E_{0.95}^\star=1$, geometric-mean IU versus naive-MC times are approximately $230$ versus $4.02\times10^5$, $645$ versus $1.07\times10^6$, and $2{,}235$ versus $2.12\times10^6$ seconds, respectively.}
    \label{fig:absolute-computational-cost}
\end{figure}

Figures~\ref{fig:absolute-token-methods}--\ref{fig:absolute-computational-cost}
show the absolute budgets underlying each main-text efficiency figure, using
the same events, methods, error thresholds, and geometric-mean summaries.
Lower budgets are better. Dashed or dotted curves show naive MC; in model
comparisons, their colors match the corresponding model and method. Shading
shows $\pm0.5$ standard deviations in log space across events, as in the main
text. Required sampling
budgets include any MC continuation described in
Appendix~\ref{app:accuracy-cost-estimation}.

\clearpage
\section{Full-detail result plots}
\label{app:full-result-plots}

Token-rarity comparisons use ten randomly selected tokens per rarity decade.
In the main result figures, markers show geometric means across events and
shading spans $\pm0.5$ standard deviations around the mean in log space. $\pm0.5$ was chosen for visual separation.

The main text uses summaries for readability. Each figure here gives four full-detail views for the corresponding setting. \textbf{Upper left:} empirical estimate quantiles for IU only $N = 128$, including both models in model comparisons. The outer markers show the 2.5\% and 97.5\% quantiles, inner bars show the interquartile range, plus markers show medians, and diamonds show per-event arithmetic means.  \textbf{Upper right:} quantile log error $E_{0.95}$ against event probability at $N=128$, with individual event points and per-decade geometric-mean lines and spread bands. \textbf{Lower left:} Estimator Efficiency Gain across error thresholds. \textbf{Lower right:} Compute-Weighted Efficiency Gain across the same thresholds, including proposal training and sampling costs from Appendix~\ref{app:compute-cost}. The error and frontier panels include all methods available in the setting.

Efficiency frontier panels retain every selected event's curve and markers beneath the geometric-mean lines. As in the main figures, token frontiers use the ten matched events per model and method in $10^{-8}\leq\mu<10^{-7}$, and profanity frontiers use the classifier cutoff $\kappa=0.999$. Error bands span $\pm0.75$ standard deviations in log space. Dashed horizontal lines in the frontier panels indicate parity with naive Monte Carlo.

\begin{figure}[p]
    \centering
    \includegraphics[width=0.49\textwidth]{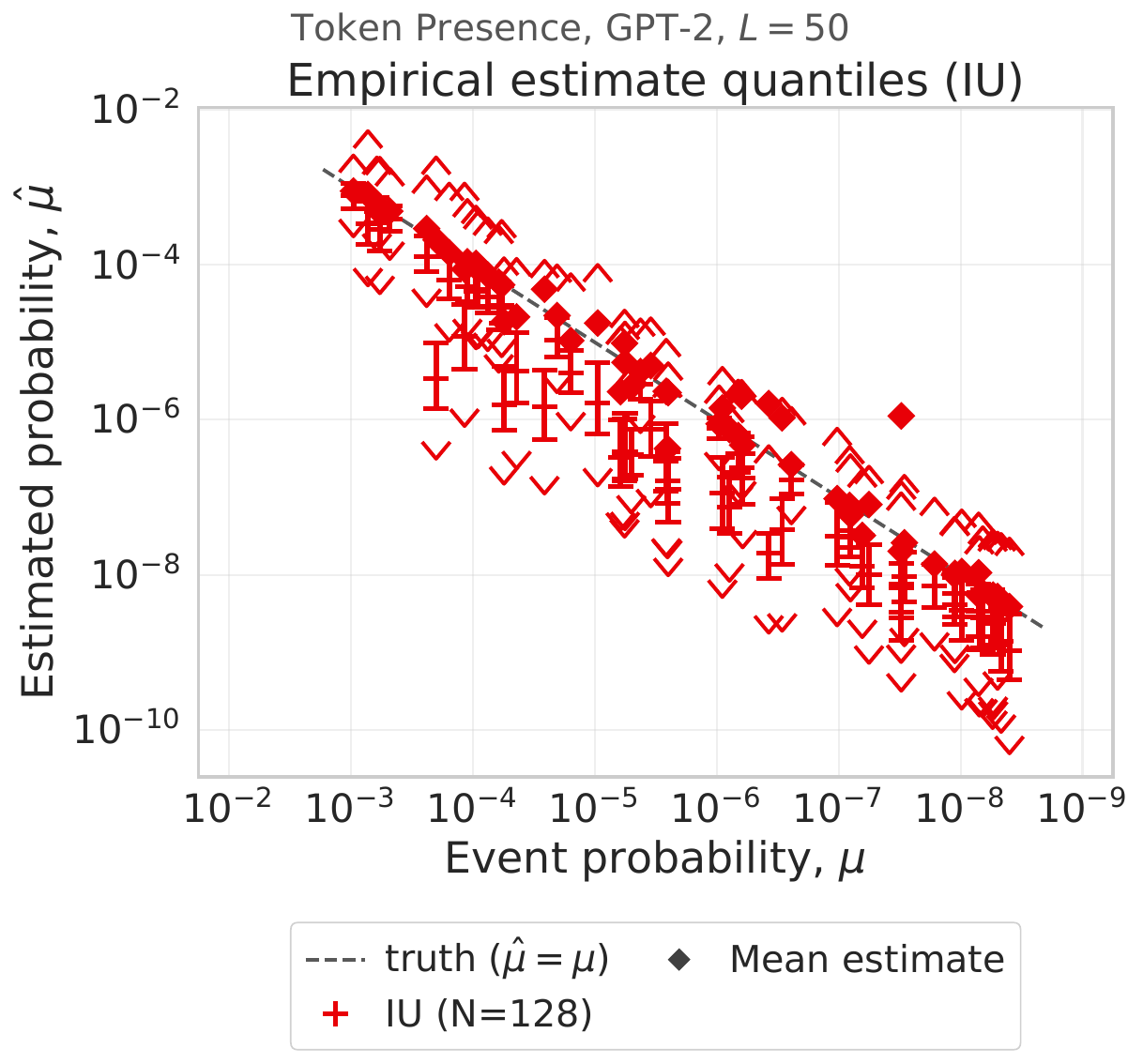}
    \hfill
    \includegraphics[width=0.49\textwidth]{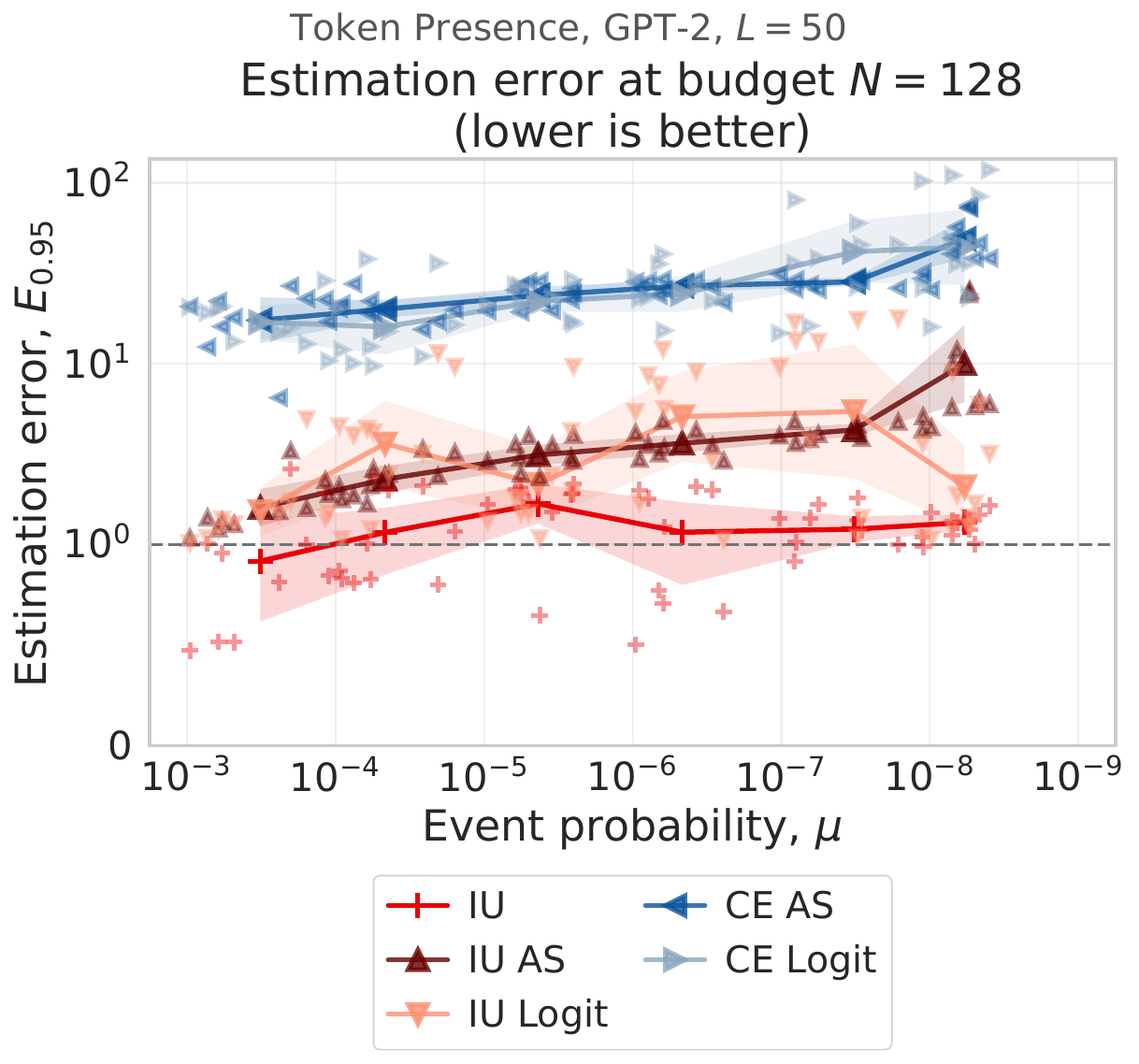}

    \includegraphics[width=0.49\textwidth]{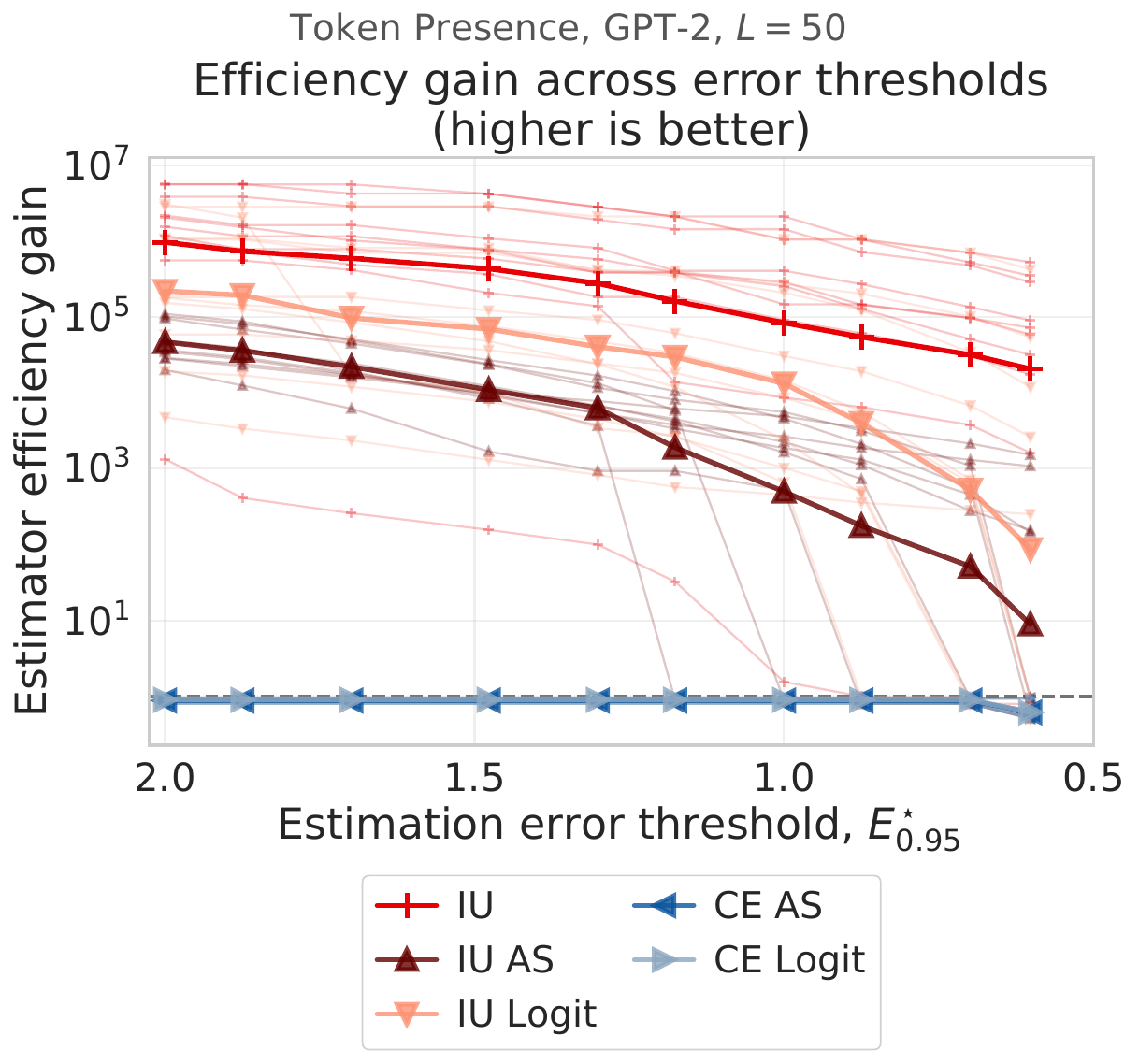}
    \hfill
    \includegraphics[width=0.49\textwidth]{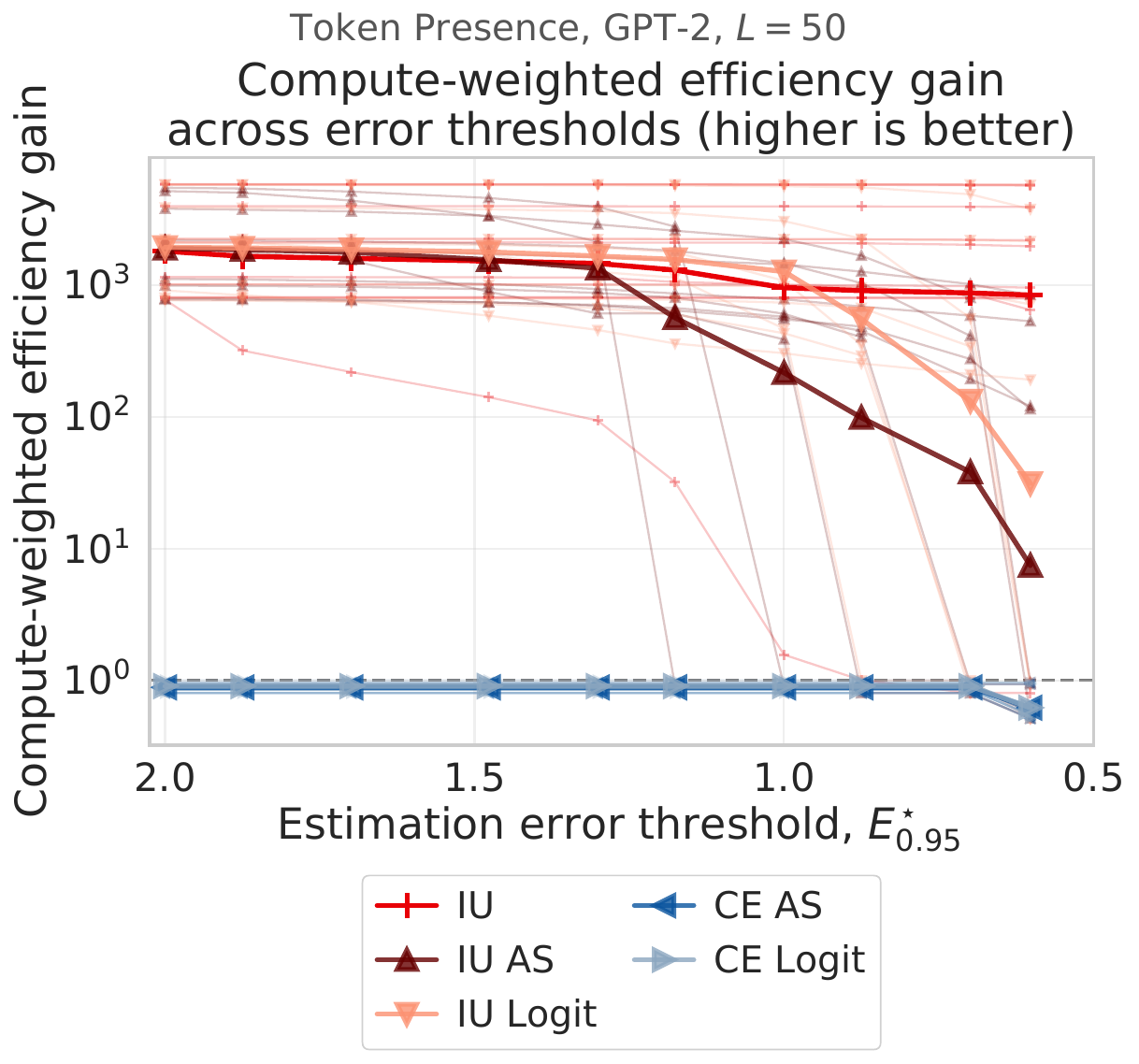}
    \caption{Full-detail Token Presence rarity results on GPT-2 Small at $L=50$: IU empirical quantiles ($N = 128$), all-method error at $N=128$, and all-method estimator and compute-weighted efficiency frontiers.}
    \label{fig:full-token-rarity}
\end{figure}

\begin{figure}[p]
    \centering
    \includegraphics[width=0.49\textwidth]{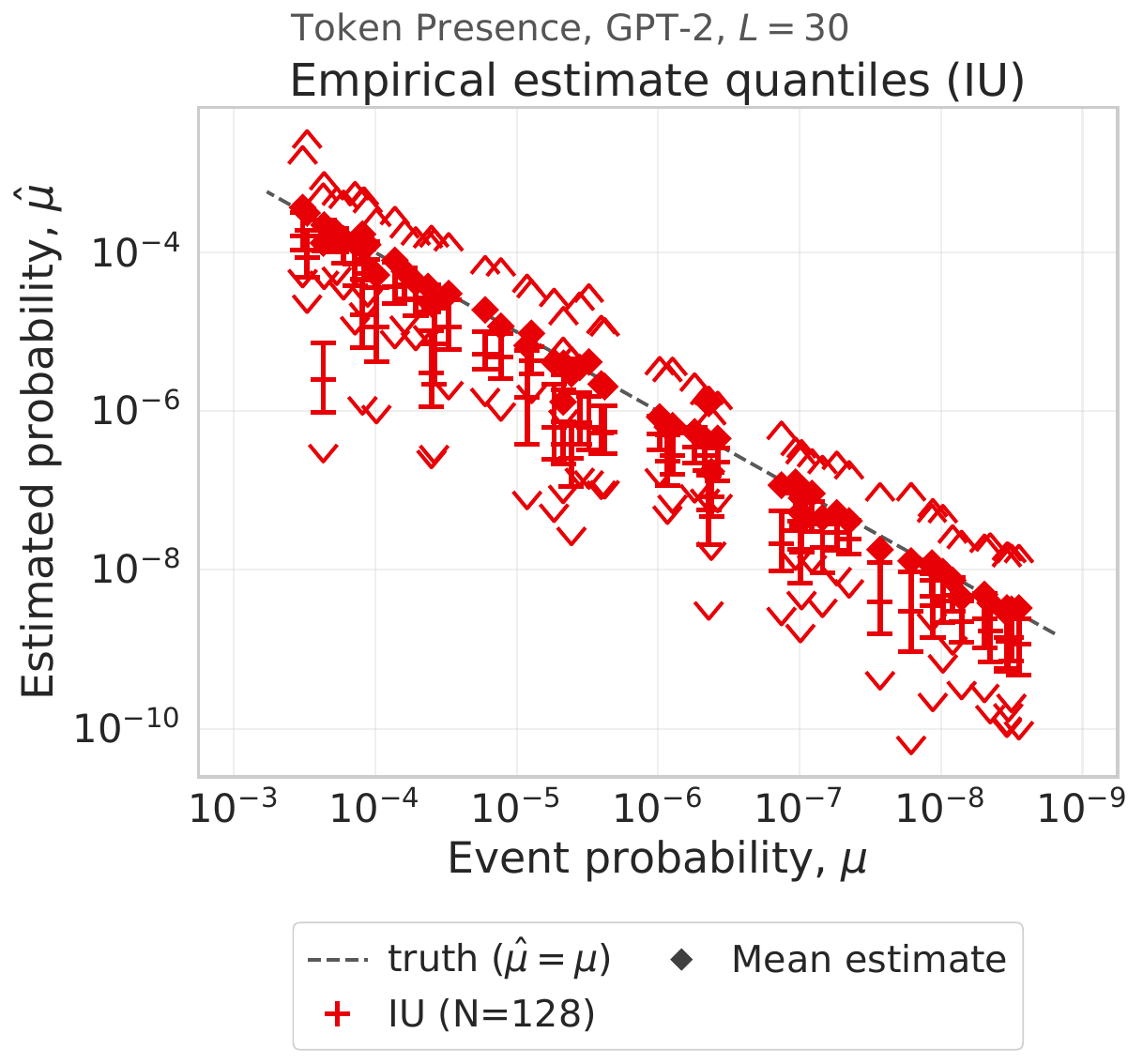}
    \hfill
    \includegraphics[width=0.49\textwidth]{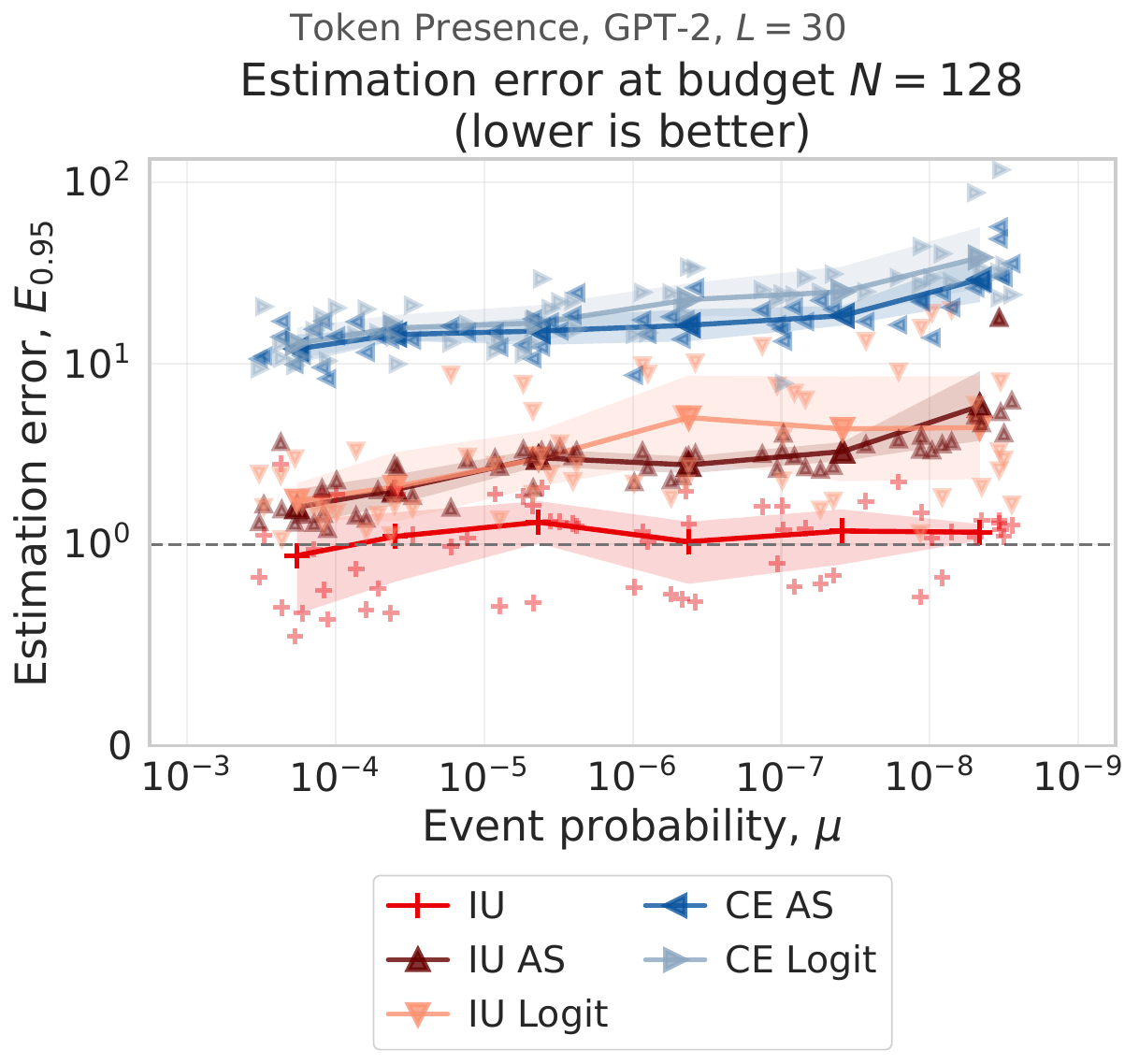}

    \includegraphics[width=0.49\textwidth]{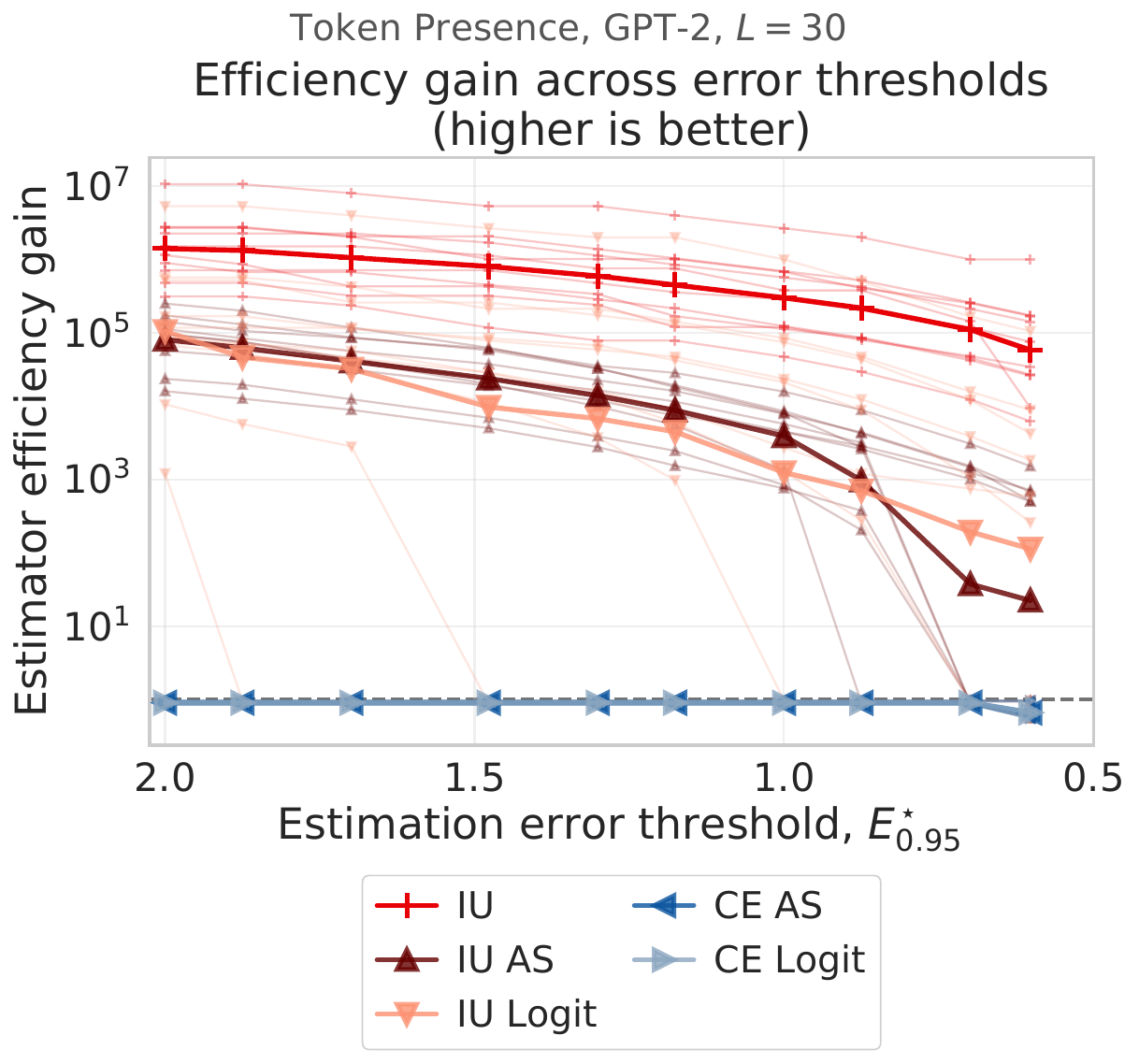}
    \hfill
    \includegraphics[width=0.49\textwidth]{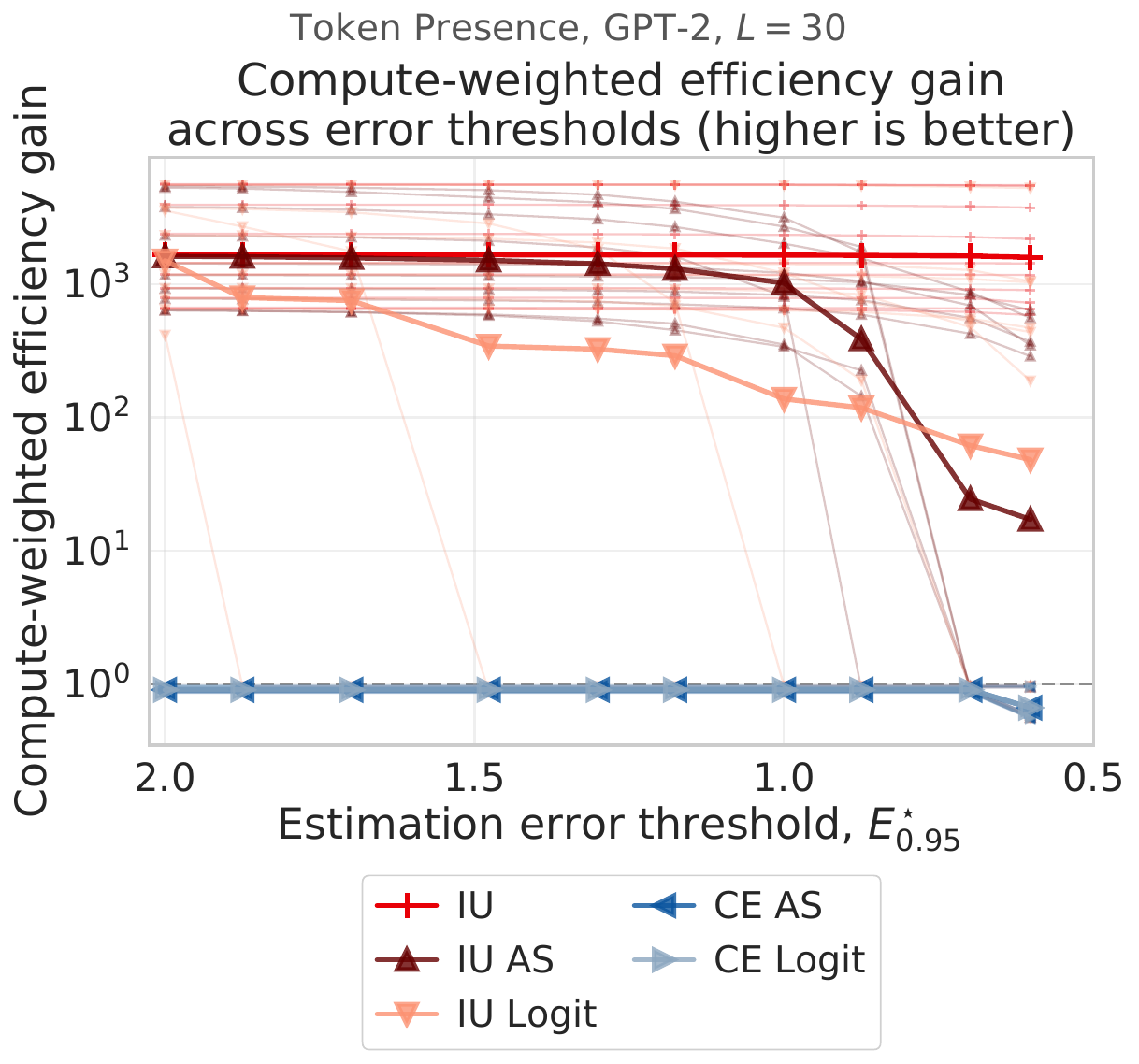}
    \caption{Full-detail Token Presence results on GPT-2 Small at $L=30$: IU empirical quantiles ($N = 128$), all-method error at $N=128$, and all-method estimator and compute-weighted efficiency frontiers.}
    \label{fig:full-token-length-30}
\end{figure}

\begin{figure}[p]
    \centering
    \includegraphics[width=0.49\textwidth]{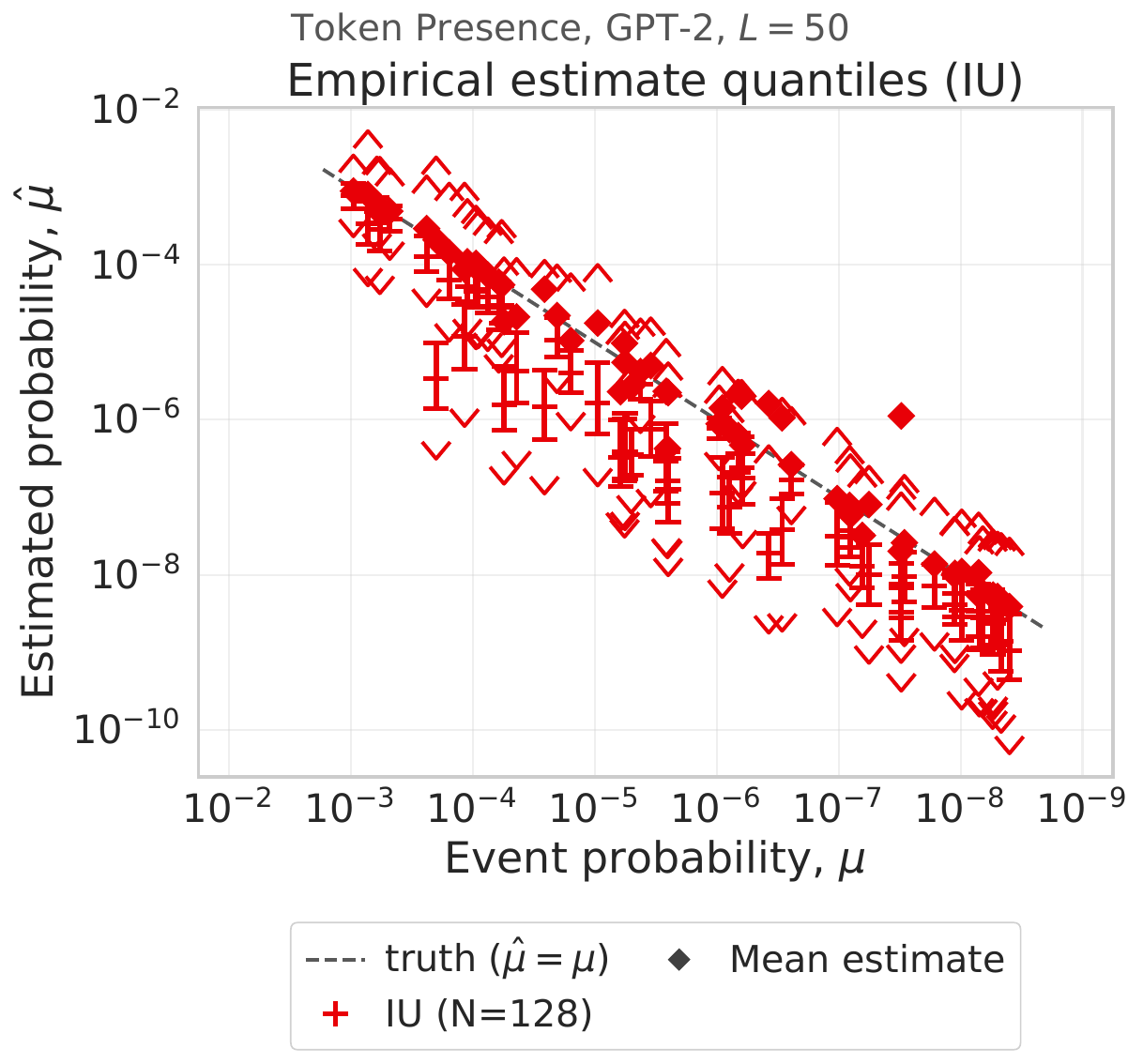}
    \hfill
    \includegraphics[width=0.49\textwidth]{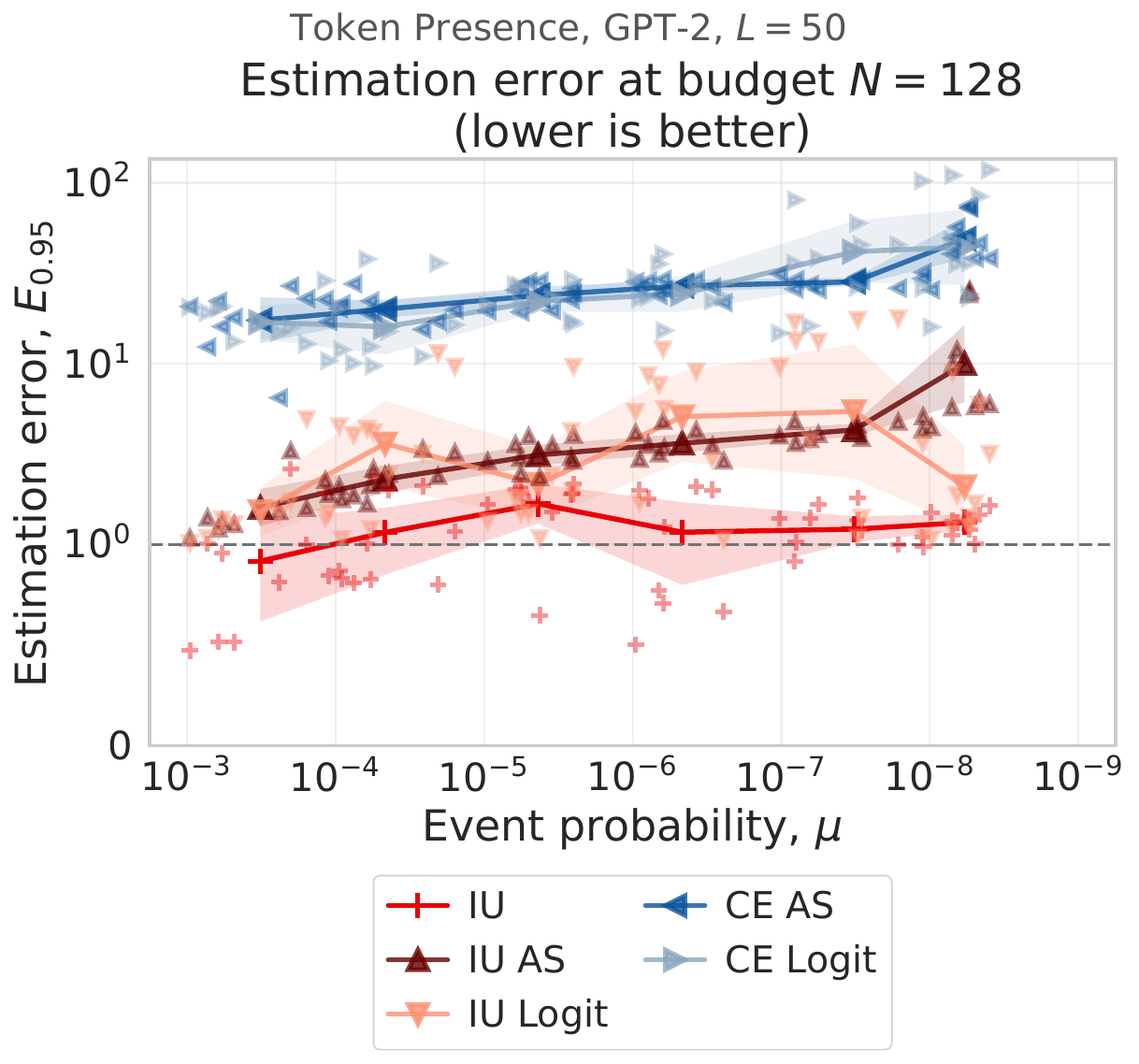}

    \includegraphics[width=0.49\textwidth]{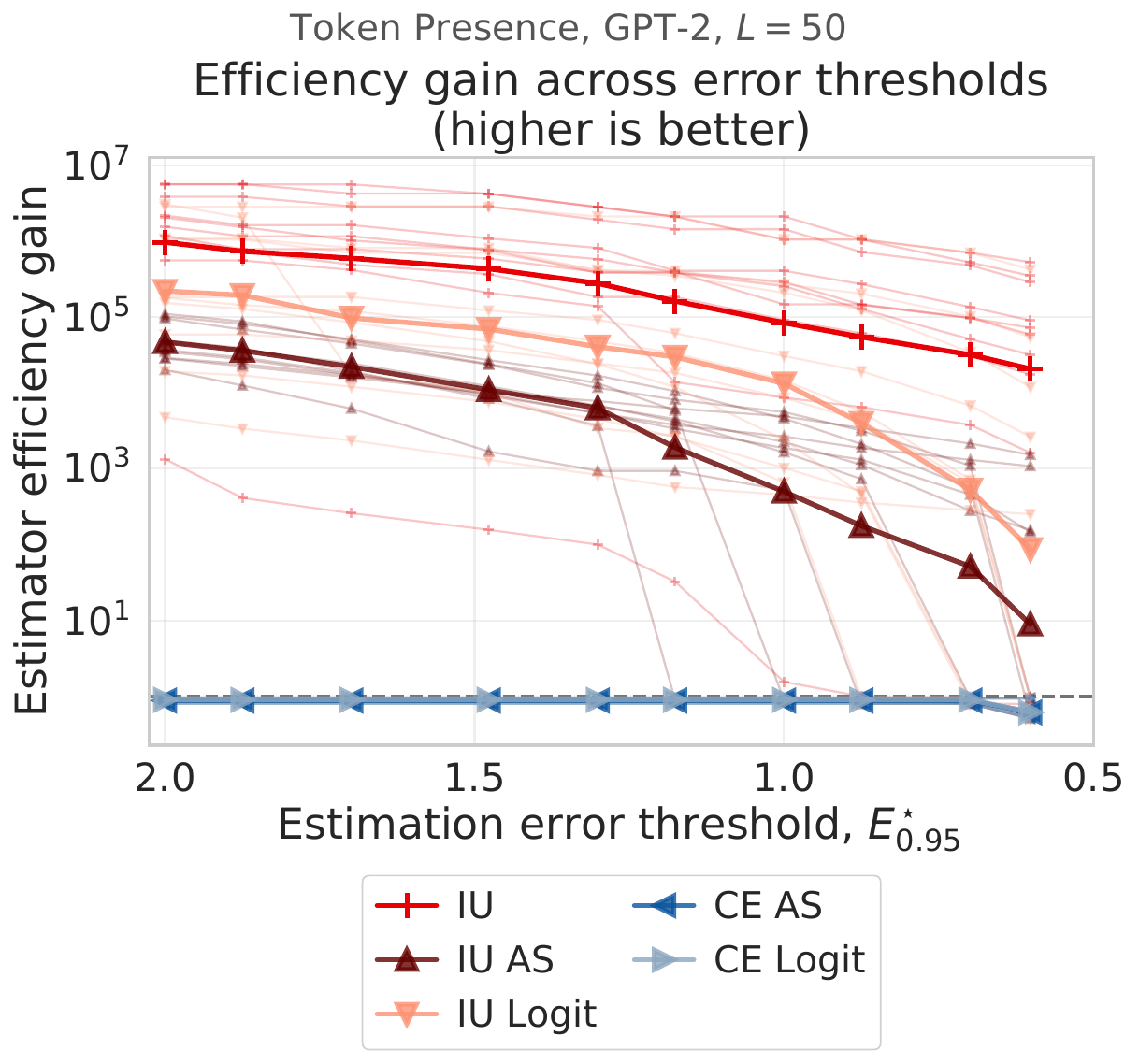}
    \hfill
    \includegraphics[width=0.49\textwidth]{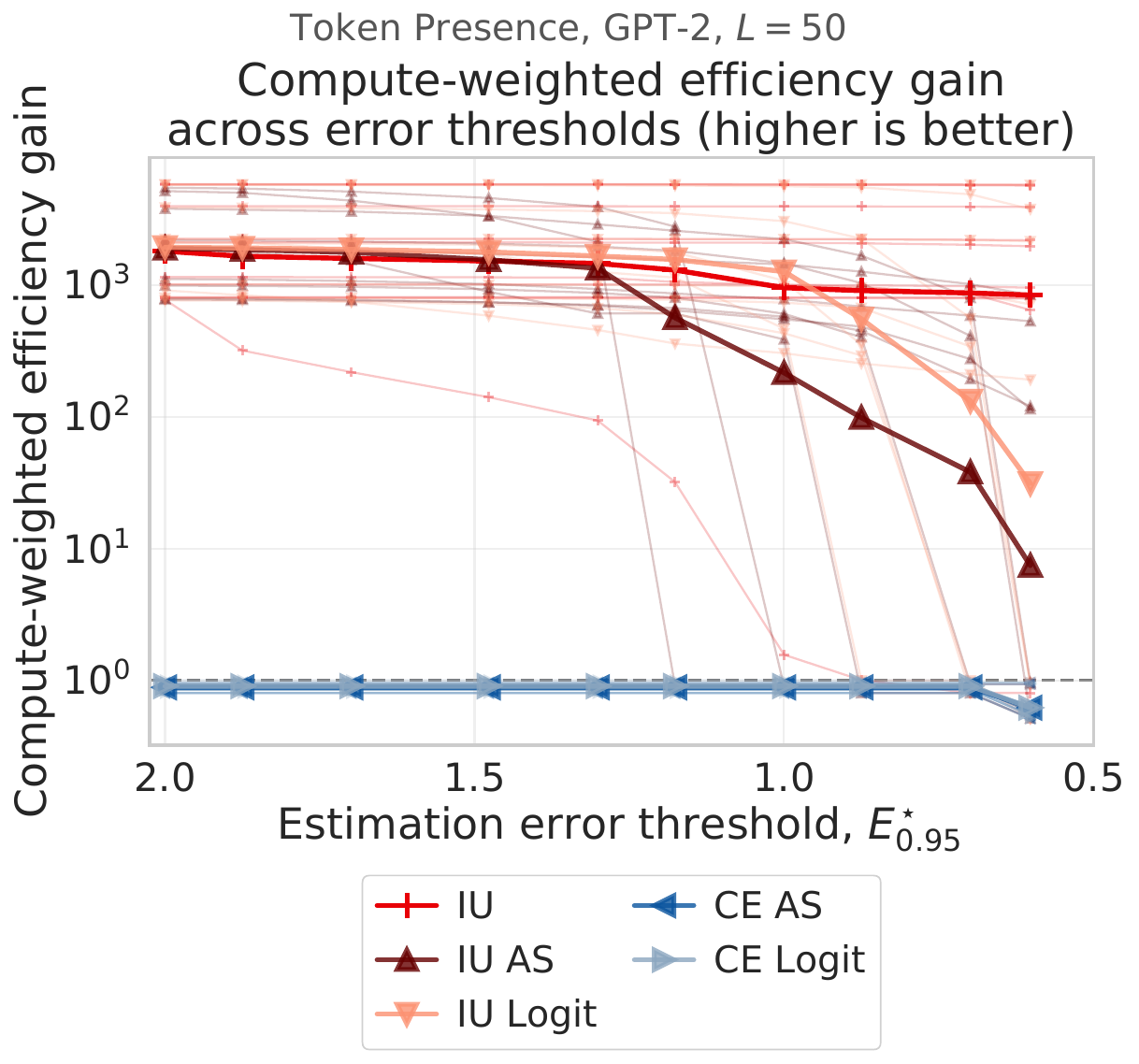}
    \caption{Full-detail Token Presence results on GPT-2 Small at $L=50$: IU empirical quantiles ($N = 128$), all-method error at $N=128$, and all-method estimator and compute-weighted efficiency frontiers.}
    \label{fig:full-token-length-50}
\end{figure}

\begin{figure}[p]
    \centering
    \includegraphics[width=0.49\textwidth]{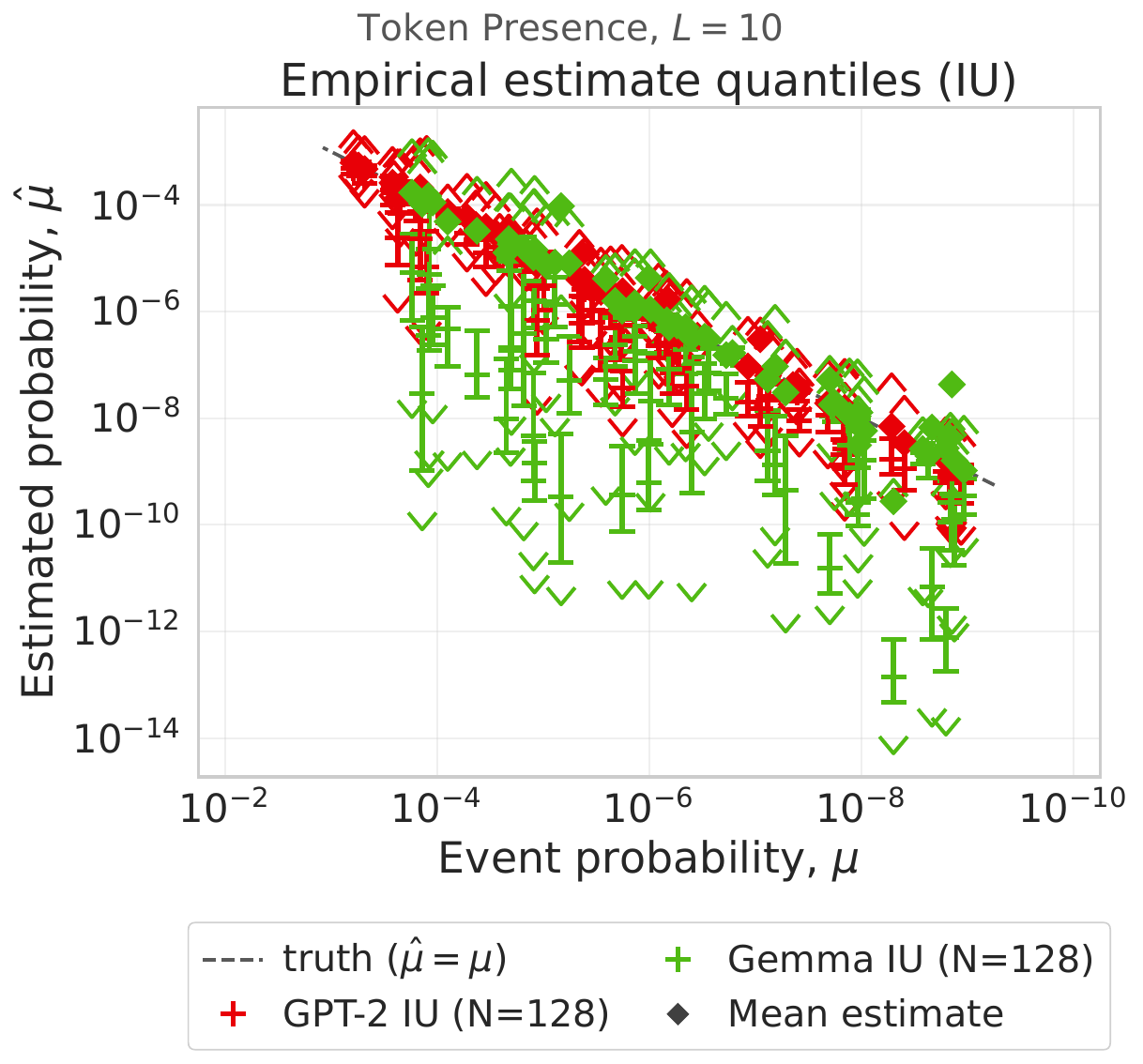}
    \hfill
    \includegraphics[width=0.49\textwidth]{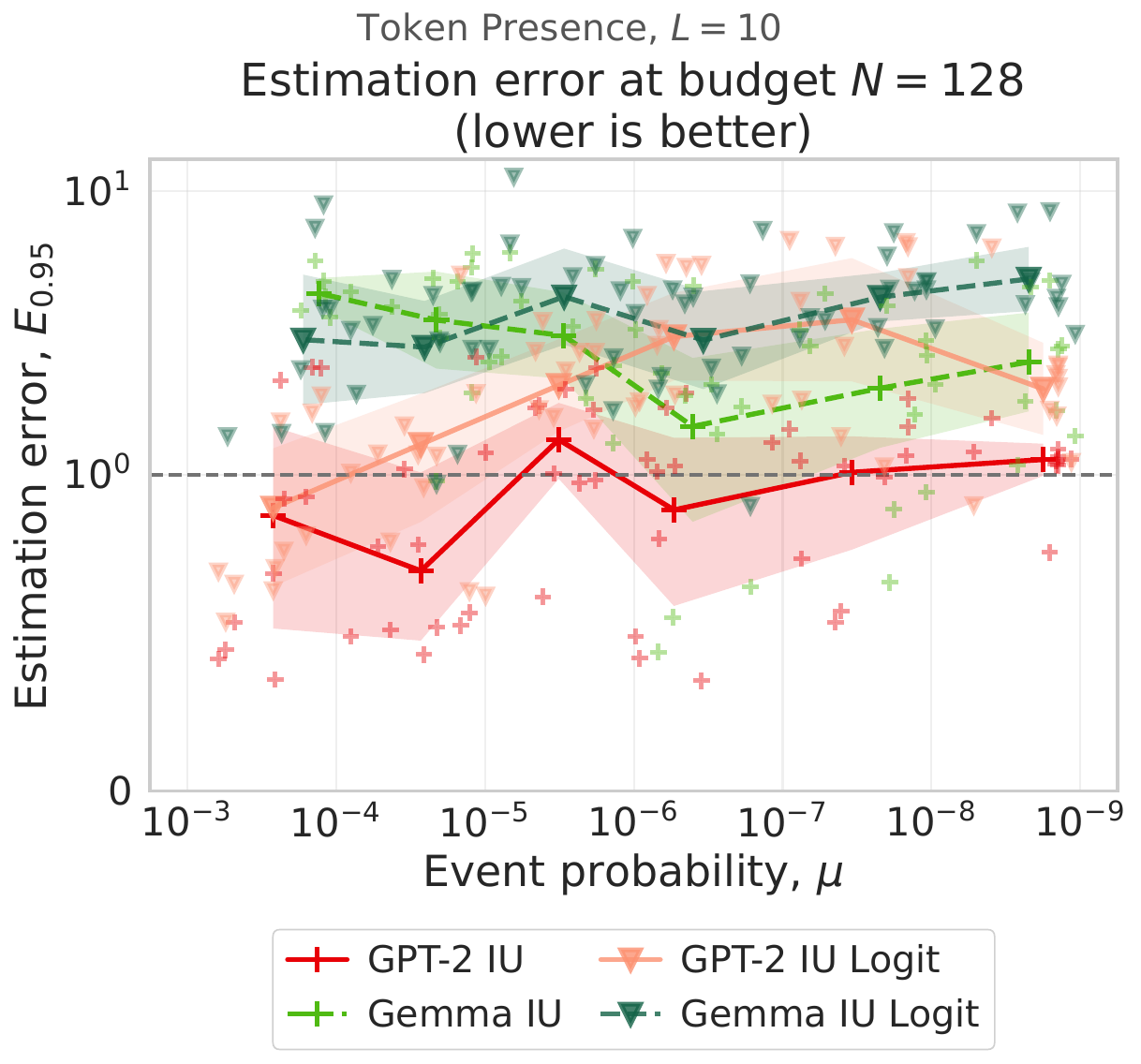}

    \includegraphics[width=0.49\textwidth]{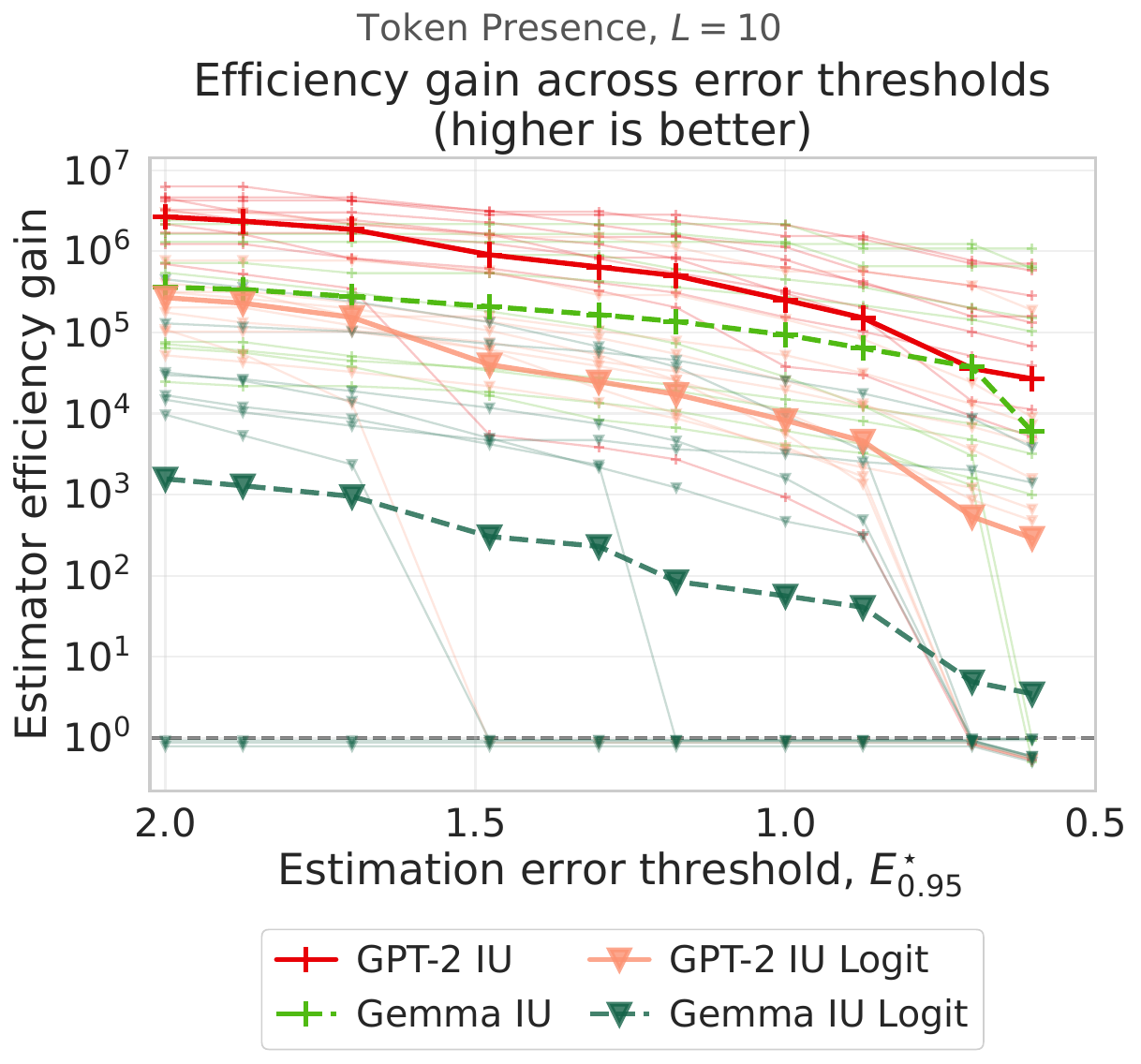}
    \hfill
    \includegraphics[width=0.49\textwidth]{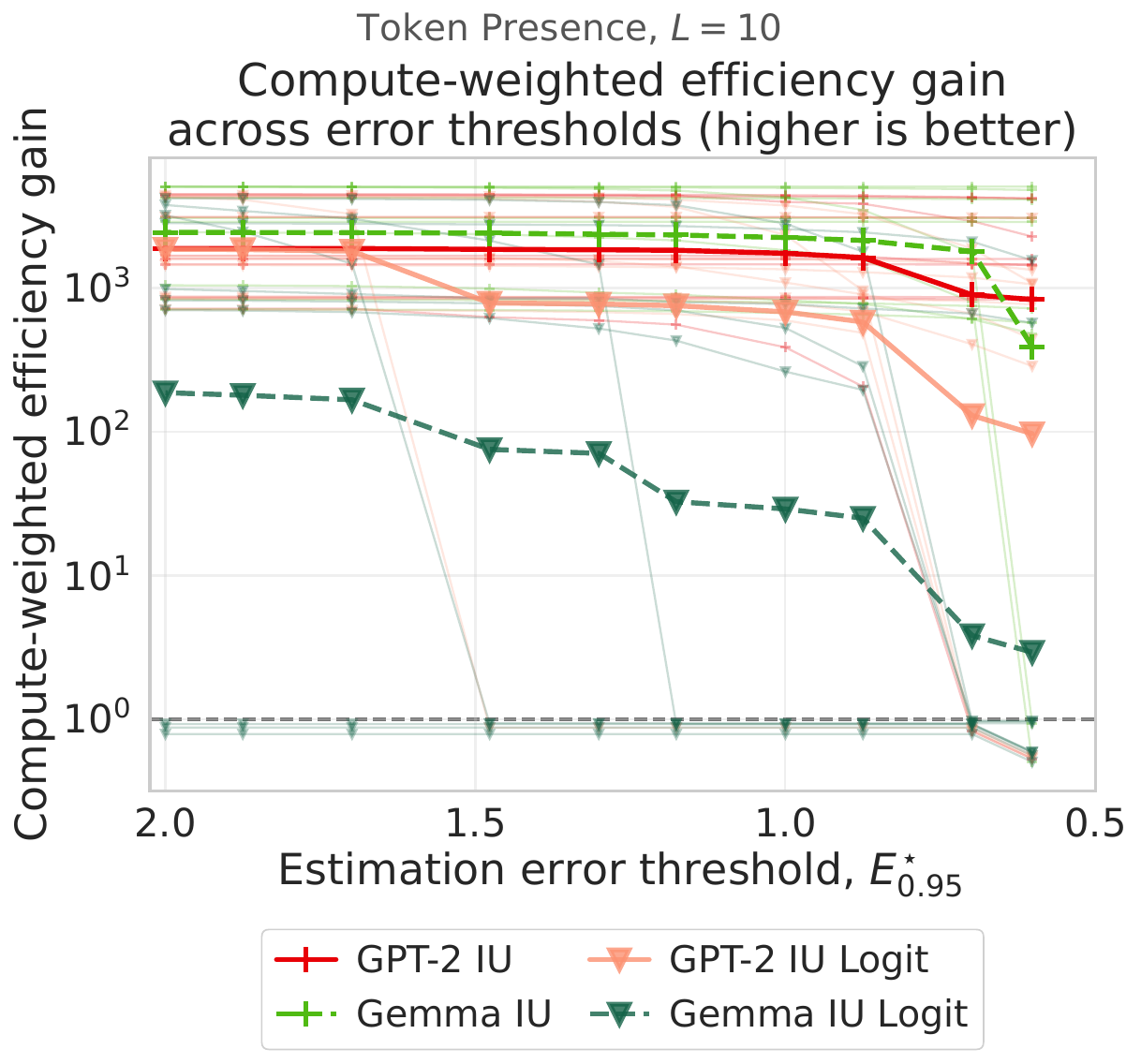}
    \caption{Full-detail Token Presence model comparison on GPT-2 Small and Gemma-2 at $L=10$. Empirical quantiles show IU for both models at $N = 128$; error at $N=128$ and the two efficiency frontiers show both IU and IU Logit for each model.}
    \label{fig:full-token-model-size}
\end{figure}

\begin{figure}[p]
    \centering
    \includegraphics[width=0.49\textwidth]{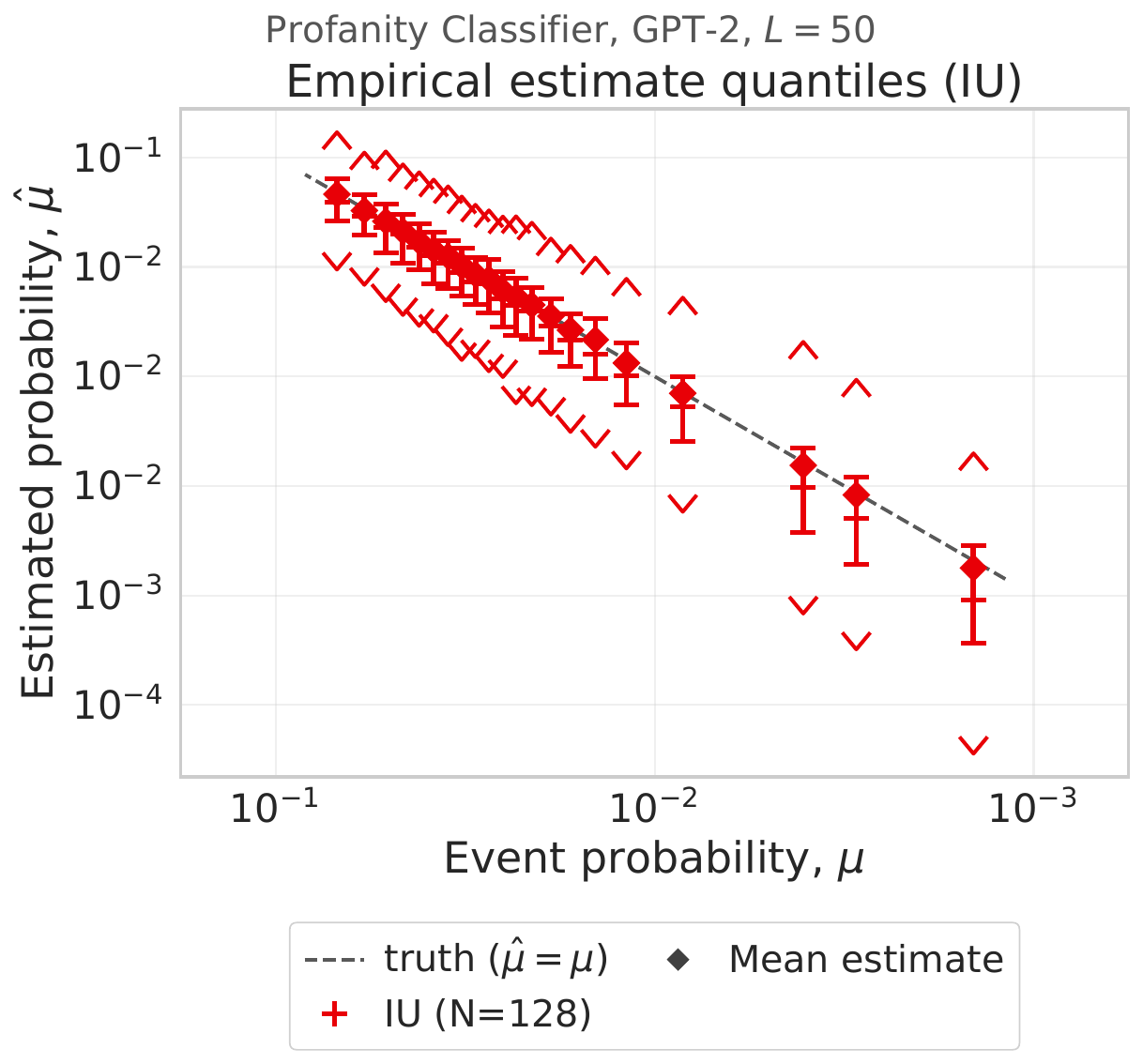}
    \hfill
    \includegraphics[width=0.49\textwidth]{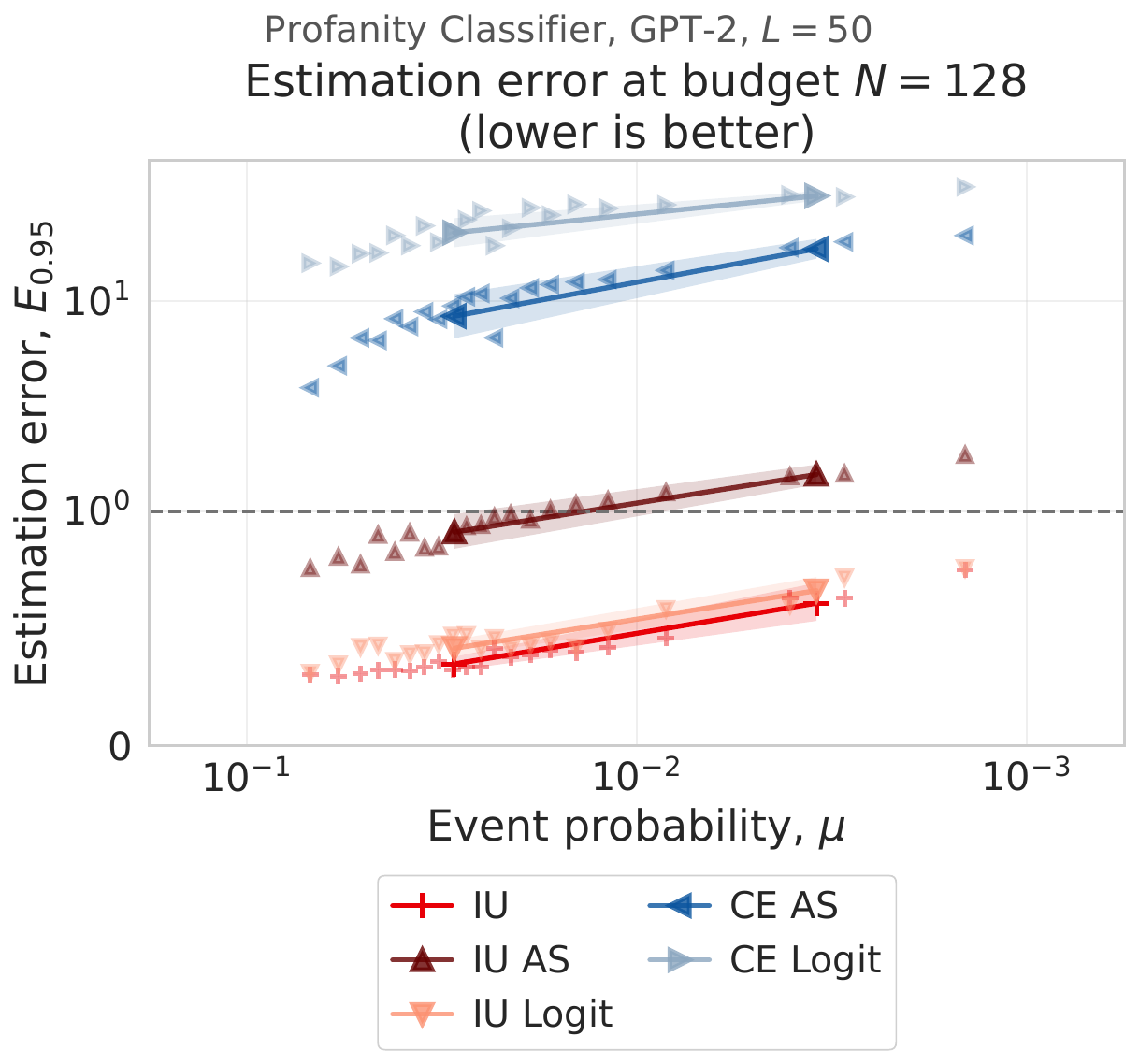}

    \includegraphics[width=0.49\textwidth]{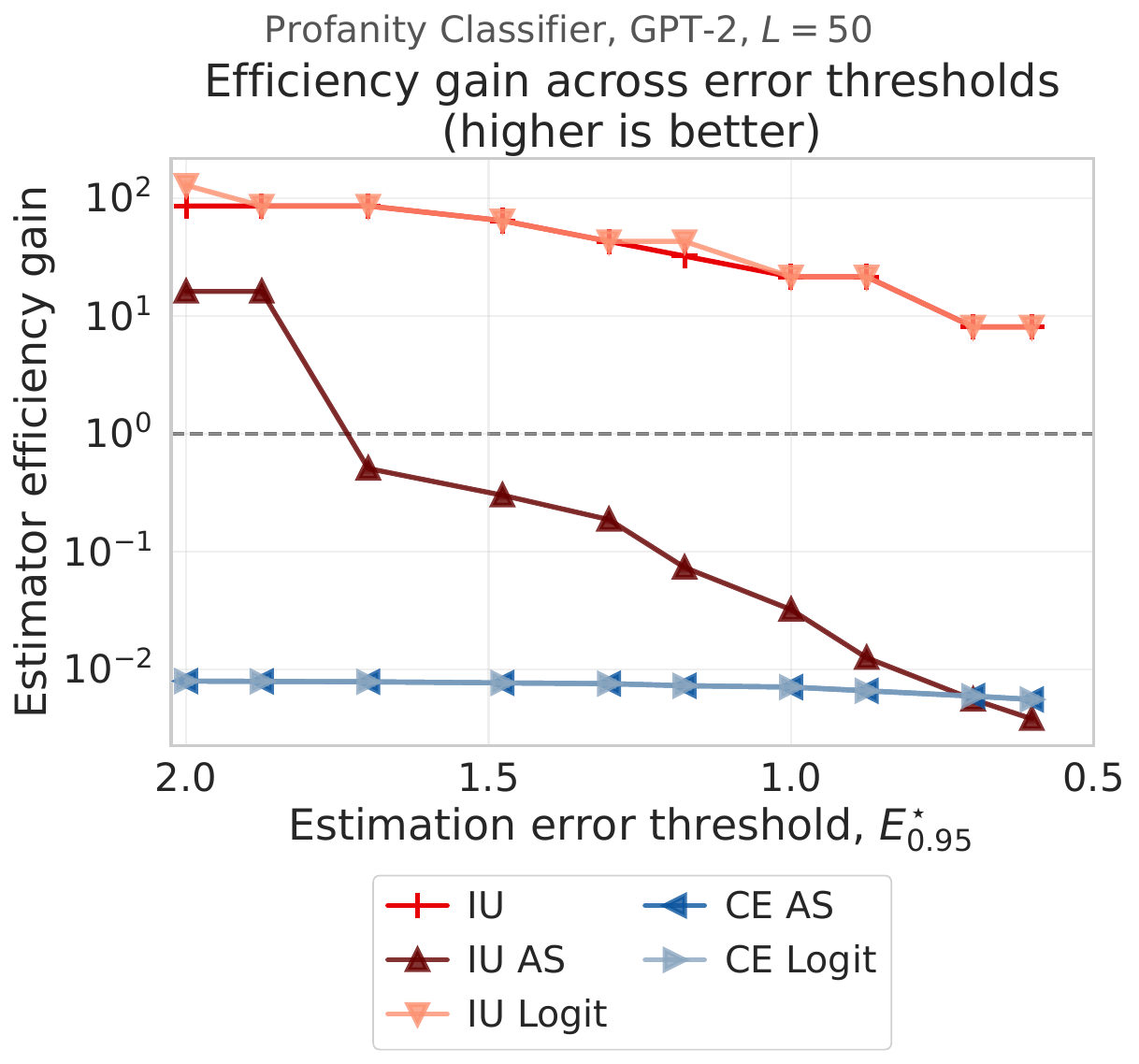}
    \hfill
    \includegraphics[width=0.49\textwidth]{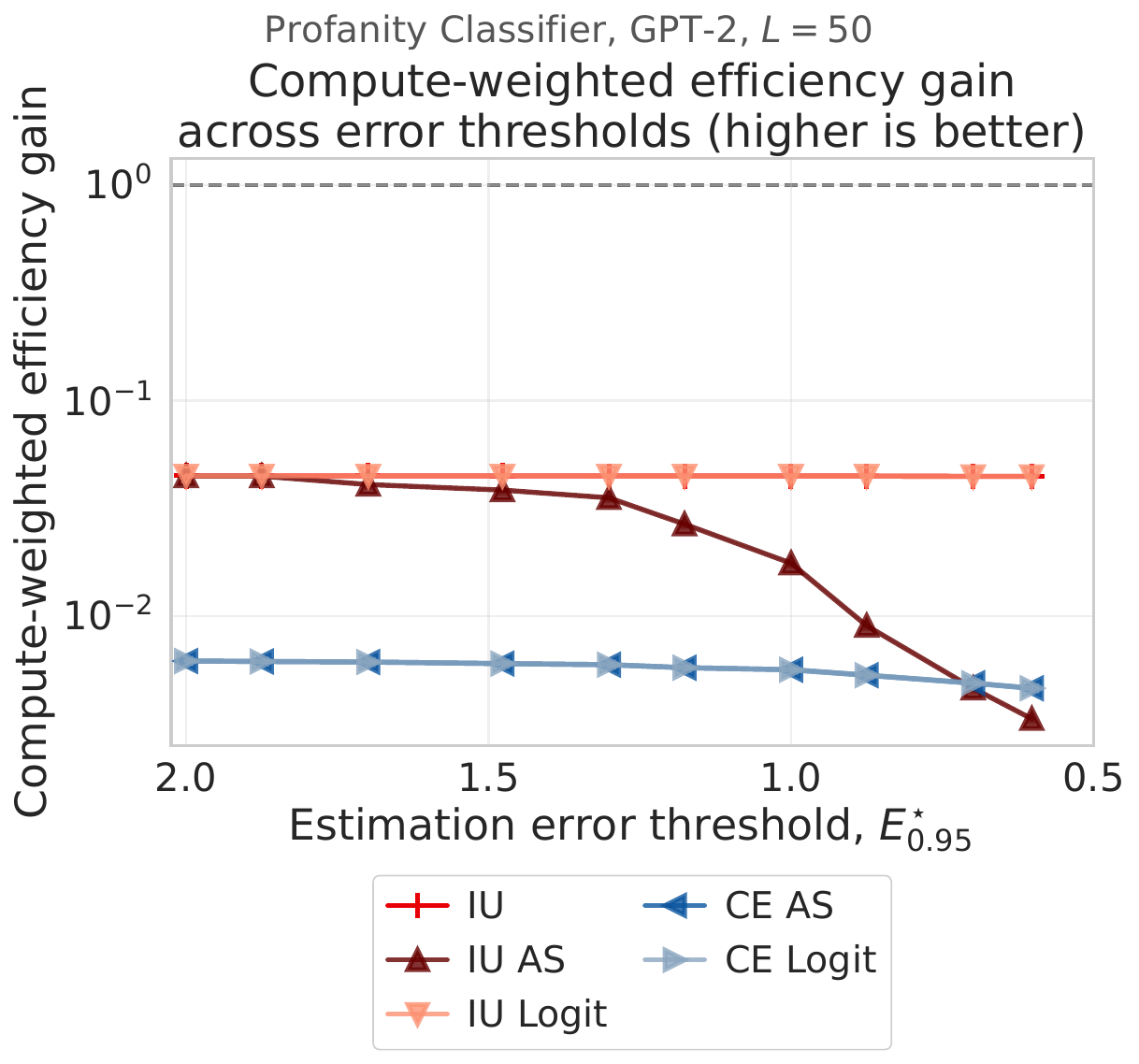}
    \caption{Full-detail Profanity Classifier results on GPT-2 Small at $L=50$: IU empirical quantiles, all-method error at $N=128$, and all-method estimator and compute-weighted efficiency frontiers at classifier cutoff $0.999$.}
    \label{fig:full-profanity-rarity}
\end{figure}

\begin{figure}[p]
    \centering
    \includegraphics[width=0.49\textwidth]{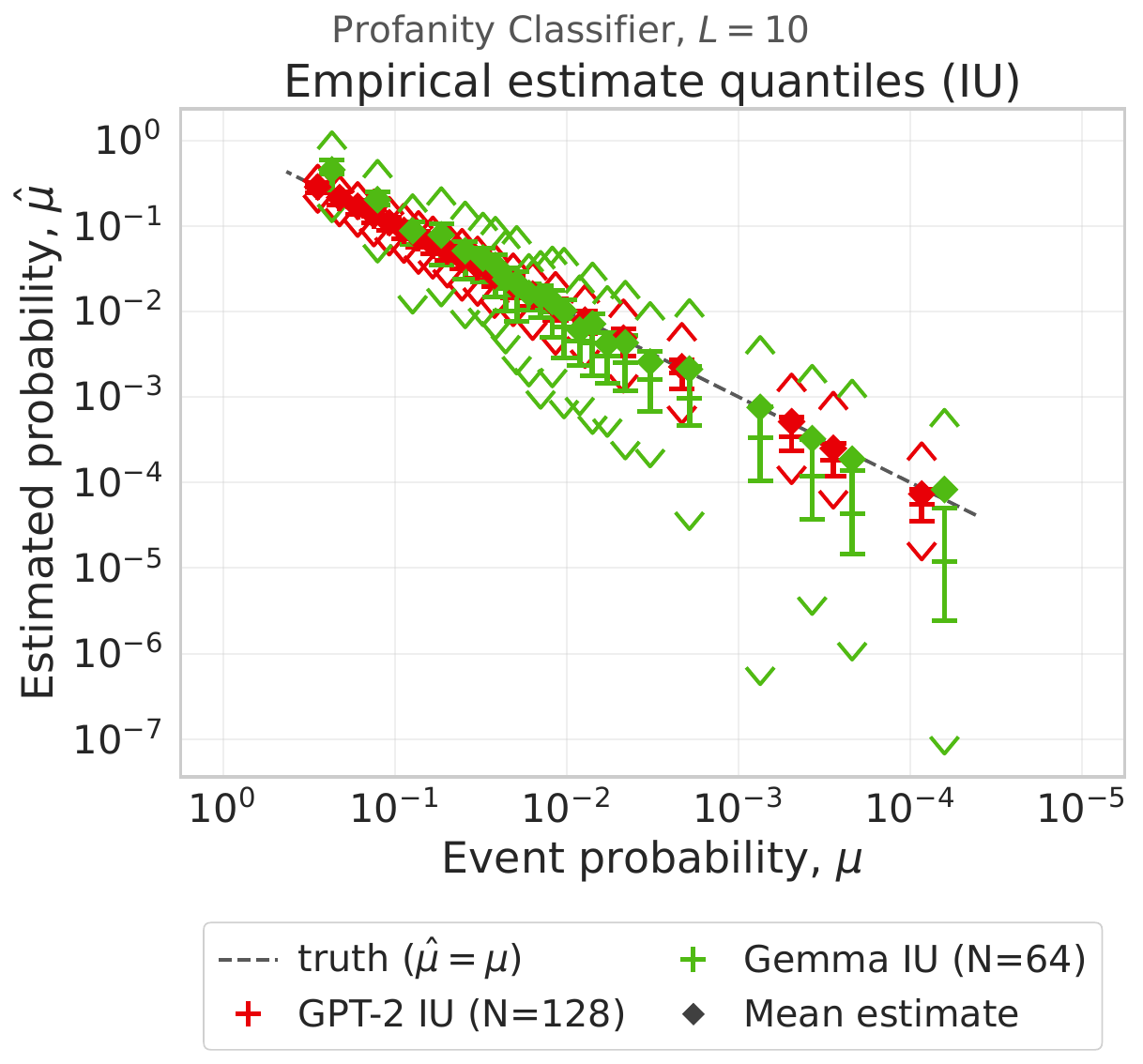}
    \hfill
    \includegraphics[width=0.49\textwidth]{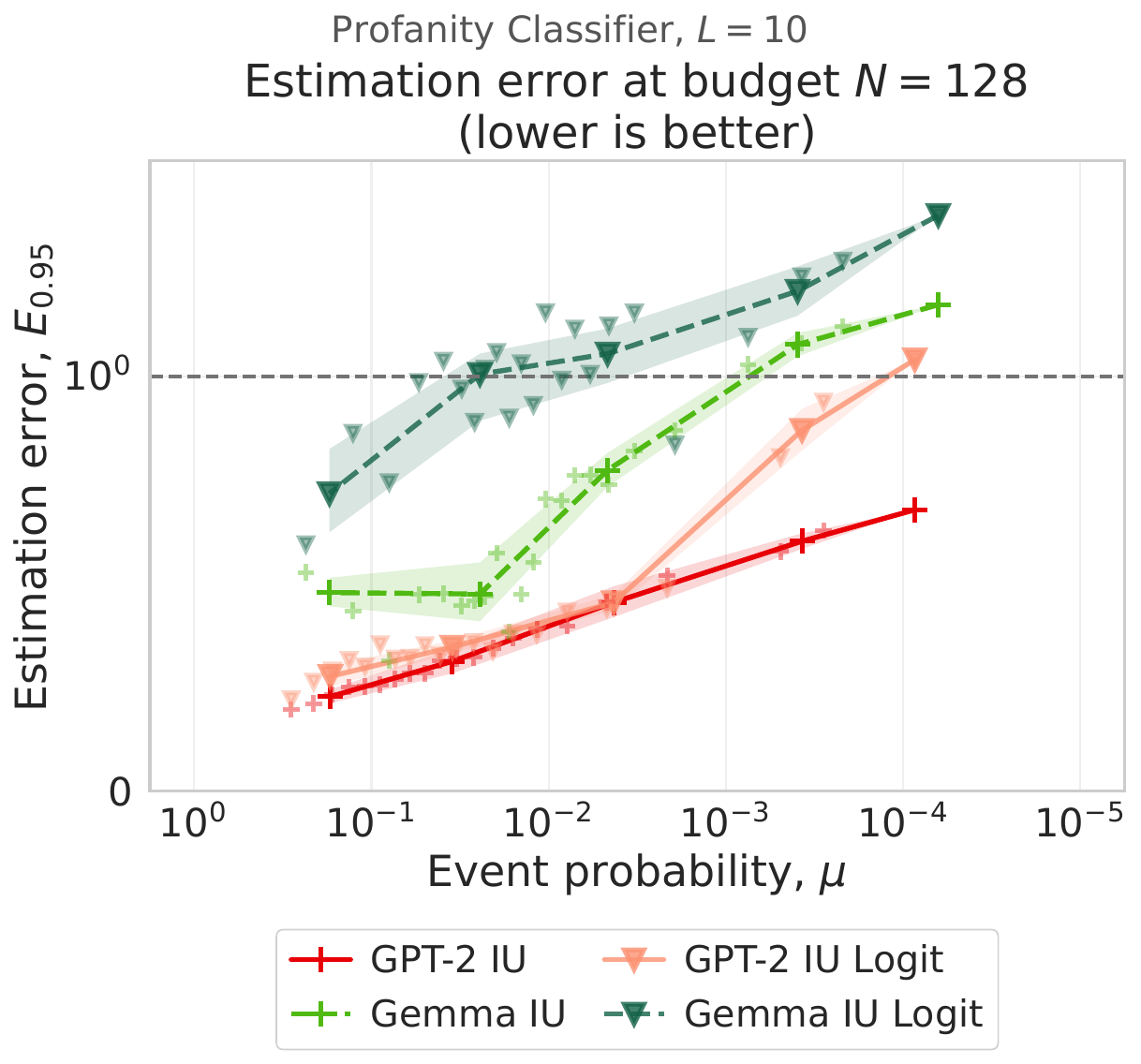}

    \includegraphics[width=0.49\textwidth]{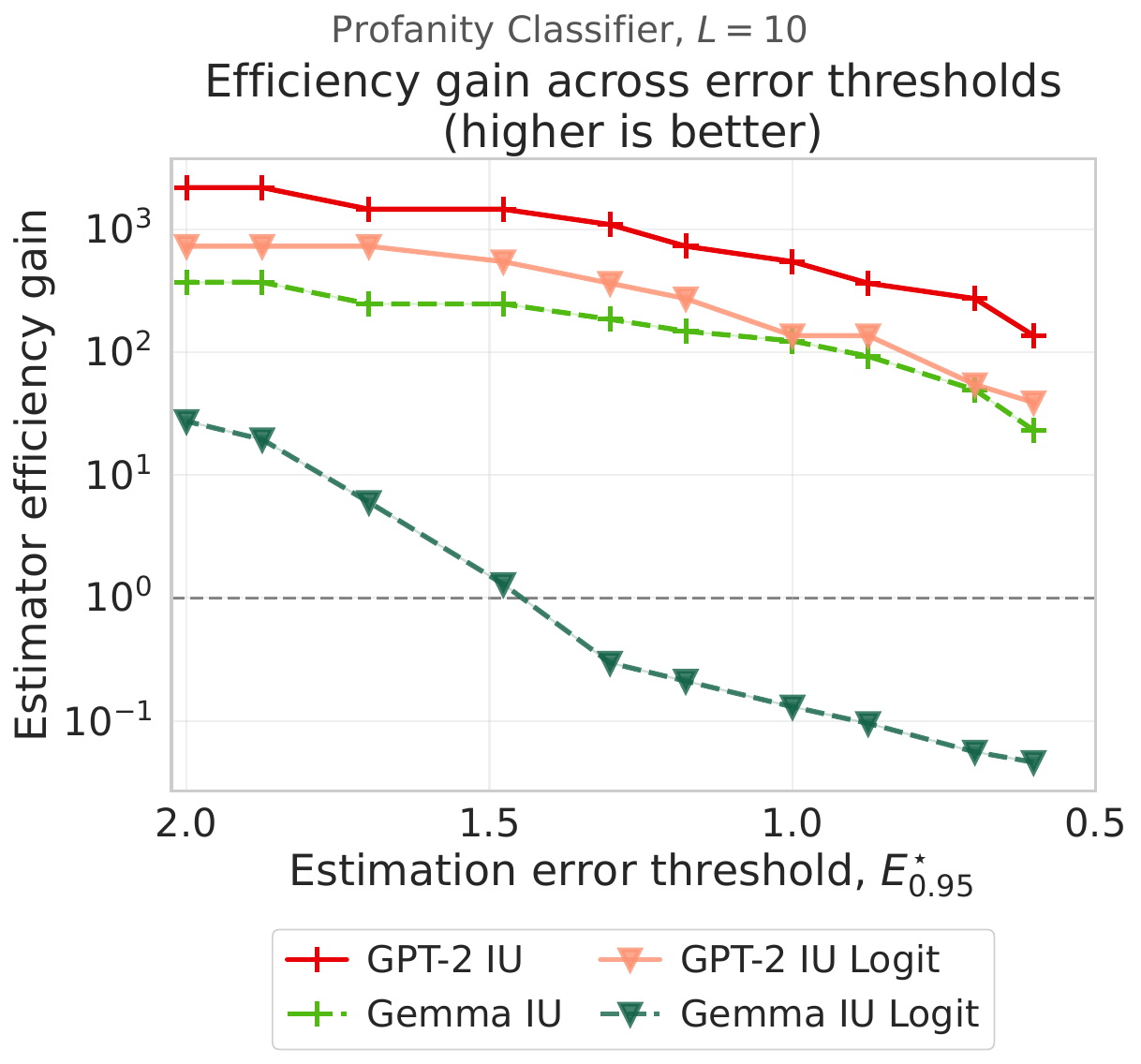}
    \hfill
    \includegraphics[width=0.49\textwidth]{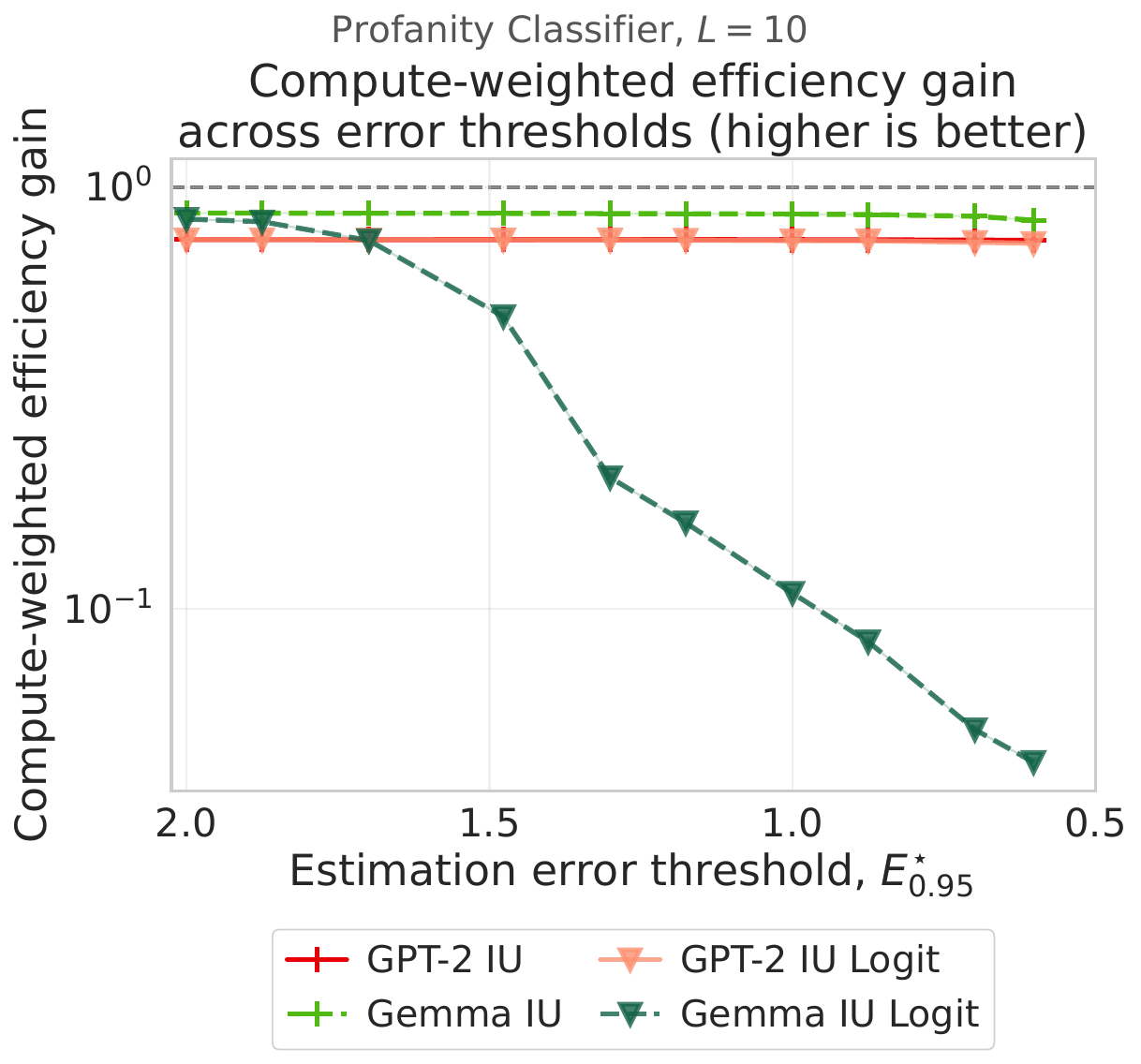}
    \caption{Full-detail Profanity Classifier model comparison on GPT-2 Small and Gemma-2 at $L=10$. Empirical quantiles show IU for both models at budget $128$. Efficiency frontiers use classifier cutoff $0.999$.}
    \label{fig:full-profanity-model-size}
\end{figure}

\clearpage

\section{All experimental tokens and events}
\label{app:all-events}

\input{tables/all_events}

%% file: dorman_comparison.tex
\section{Comparison with score-guided MCMC}
\label{app:dorman-comparison}

Dorman et al.~\cite{dorman2026rare} study rare-event estimation in language models by reconstructing the distribution of continuous scalar scores $g(\mathbf{y})$ (such as output log probability and readability) under the base model $p$ using score-biased MCMC and multistate Bennett acceptance ratio (MBAR) reweighting. While their primary focus is on continuous score distributions where exact tail reference probabilities are generally unavailable, their framework also supports estimating a target indicator $\Phi(\mathbf{y})$ using a correlated auxiliary scalar score to bias the MCMC transitions. This allows us to evaluate score-guided MCMC on Token Presence events, where precise reference probabilities allow us to verify accuracy in the deep tail.

We set the target to the binary event indicator $\Phi(\mathbf{y})$ and use the Token Presence surrogate as the biasing score, $g(\mathbf{y})=S^{\mathrm{TOK}}_\delta(\mathbf{y};v)$, evaluated with $q_\delta=p$. This score directly rewards prefixes that assign higher probability to the target token, providing an aligned search signal for the MCMC sampler.

For each event, we run 10 chains across the positive bias schedule $0.1,0.2,\ldots,1.0$, with 1,219 transitions per chain and bias, 10\% burn-in, and 6,090 direct samples. Including the 10 initial chain states, this gives 128,000 evaluated trajectories for one complete estimate. We retain rejected states and exclude a bias only when its Gelman--Rubin diagnostic exceeds $1.1$, following the authors' procedure. The schedule is fixed across events and receives no event-specific tuning.

Dorman's MCMC transition truncates each trajectory at a uniformly random output position and regenerates the remaining suffix. We conservatively charge each evaluated trajectory as only half a naive-MC inference. From Table~\ref{tab:compute-cost-constants}, the charged cost is
\begin{equation*}
128{,}000(0.5)\,\mathrm{cost}_{\mathrm{MC}}
=128{,}000(0.5)(0.0046)=294.4\text{ seconds}.
\end{equation*}
For comparison, the full cost of one $N=128$ IU estimate (including proposal training) is
\begin{equation*}
\mathrm{cost}_{\mathrm{train}}+128\,\mathrm{cost}_{\mathrm{IS}}
=213+128(0.0051)=213.65\text{ seconds}.
\end{equation*}
Thus, in this experiment, Dorman's charged budget is $294.4/213.65=1.38\times$ the full cost of an $N=128$ IU estimate.

Table~\ref{tab:dorman-verified-comparison} reports event discovery and probability estimation. The event hit rate is the fraction of sampled trajectories satisfying the Token Presence event.

\begin{table}[H]
\centering
\small
\textbf{(a) Iterative Unalignment}\\[3pt]
\begin{tabular}{@{}lrrr@{}}
\toprule
Event & Reference $\mu$ & Event hit rate & Final estimate \\
\midrule
\texttt{mere}           & $6.1594\times10^{-4}$ & $0.620$ & $6.1927\times10^{-4}$ \\
\texttt{atten}          & $1.1463\times10^{-5}$ & $0.167$ & $7.4793\times10^{-6}$ \\
\texttt{iminary}        & $1.7840\times10^{-6}$ & $0.311$ & $2.4233\times10^{-6}$ \\
\texttt{sembly}         & $1.1767\times10^{-7}$ & $0.345$ & $9.3520\times10^{-8}$ \\
\texttt{soDeliveryDate} & $1.4414\times10^{-8}$ & $0.264$ & $1.2899\times10^{-8}$ \\
\bottomrule
\end{tabular}

\vspace{8pt}
\textbf{(b) Dorman et al.}\\[3pt]
\begin{tabular}{@{}lrrr@{}}
\toprule
Event & Reference $\mu$ & Event hit rate & Final estimate \\
\midrule
\texttt{mere}           & $6.1594\times10^{-4}$ & $1.04\times10^{-3}$ & $6.0048\times10^{-4}$ \\
\texttt{atten}          & $1.1463\times10^{-5}$ & $6.67\times10^{-5}$ & $7.3987\times10^{-6}$ \\
\texttt{iminary}        & $1.7840\times10^{-6}$ & $0$ & $0$ \\
\texttt{sembly}         & $1.1767\times10^{-7}$ & $0$ & $0$ \\
\texttt{soDeliveryDate} & $1.4414\times10^{-8}$ & $0$ & $0$ \\
\bottomrule
\end{tabular}
\caption{IU and score-guided MCMC on GPT-2 Small Token Presence events at $L=10$. Under the compute accounting above, IU makes every event common under its proposal, while Dorman et al.\ yield zero event hits for events with reference probabilities below $\sim10^{-5}$.}
\label{tab:dorman-verified-comparison}
\end{table}

These results illustrate the structural distinction between trajectory-space MCMC and global proposal optimization. Score-guided MCMC relies on local suffix mutations, which can struggle to discover isolated target modes in high-dimensional discrete sequence spaces when the event is exceedingly rare ($\mu < 10^{-5}$). In such regimes, MCMC chains may fail to record event hits despite satisfying standard internal diagnostics (such as Gelman--Rubin $\hat{R} \le 1.1$). In contrast, IU optimizes model parameters globally, steering autoregressive decoding dynamics across all prefix positions to reliably sample deep-tail events and provide accurate importance-sampling estimates.

%% file: tables/all_events.tex
This section lists all 319 distinct model--length--event combinations in the experimental inventory: 246 Token Presence events, 63 Profanity Classifier events, 5 Conjunctive events, and 5 Ordered Sequence events. The listed $\mu$ values are the reference probabilities used by the experiments. Lengths $L$ count generated tokens after the prompt. GPT-2 Small uses \texttt{Once upon a time}; Gemma-2 uses \texttt{Tell me a story} with its chat template.

\begin{table}[!htbp]
\centering
\scriptsize
\setlength{\tabcolsep}{3pt}
\caption{All 60 Token Presence events for GPT-2 Small at $L=10$ and their reference probabilities. Read the left list first, then the right; events are ordered from rarer to more common.}
\label{tab:all-tokens-gpt2-small-l10}
\label{tab:token_rarity}
\begin{minipage}[t]{0.49\textwidth}
\centering
\begin{tabular}{@{}lr@{}}
\toprule
Token & Ground truth $\mu$ \\
\midrule
\texttt{\eventtokenfont ActionCode} & $1.151\times10^{-9}$ \\
\texttt{\eventtokenfont ␣\textbackslash u30b5\textbackslash u30fc\textbackslash u30c6\textbackslash u30a3} & $1.41\times10^{-9}$ \\
\texttt{\eventtokenfont reportprint} & $1.41\times10^{-9}$ \\
\texttt{\eventtokenfont ␣RandomRedditor} & $1.413\times10^{-9}$ \\
\texttt{\eventtokenfont ␣externalToEVA} & $1.413\times10^{-9}$ \\
\texttt{\eventtokenfont ␣externalTo} & $1.445\times10^{-9}$ \\
\texttt{\eventtokenfont oreAndOnline} & $1.463\times10^{-9}$ \\
\texttt{\eventtokenfont ␣antidepress} & $1.607\times10^{-9}$ \\
\texttt{\eventtokenfont ␣teasp} & $3.942\times10^{-9}$ \\
\texttt{\eventtokenfont ÍÍ} & $5.2\times10^{-9}$ \\
\texttt{\eventtokenfont \textbackslash u30b4\textbackslash u30f3} & $1.4403\times10^{-8}$ \\
\texttt{\eventtokenfont soDeliveryDate} & $1.4414\times10^{-8}$ \\
\texttt{\eventtokenfont soType} & $1.4863\times10^{-8}$ \\
\texttt{\eventtokenfont natureconservancy} & $2.0748\times10^{-8}$ \\
\texttt{\eventtokenfont \textbackslash u30fc\textbackslash u30c6} & $3.8521\times10^{-8}$ \\
\texttt{\eventtokenfont ␣pestic} & $4.0529\times10^{-8}$ \\
\texttt{\eventtokenfont ␣unfocusedRange} & $4.43\times10^{-8}$ \\
\texttt{\eventtokenfont \textbackslash u9f8d\textbackslash u559a\textbackslash u58eb} & $7.4453\times10^{-8}$ \\
\texttt{\eventtokenfont ItemImage} & $7.6178\times10^{-8}$ \\
\texttt{\eventtokenfont Thumbnail} & $8.9868\times10^{-8}$ \\
\texttt{\eventtokenfont sembly} & $1.17665\times10^{-7}$ \\
\texttt{\eventtokenfont ␣SolidGoldMagikarp} & $3.50913\times10^{-7}$ \\
\texttt{\eventtokenfont Delivery} & $4.43993\times10^{-7}$ \\
\texttt{\eventtokenfont zbek} & $5.28826\times10^{-7}$ \\
\texttt{\eventtokenfont Database} & $6.0715\times10^{-7}$ \\
\texttt{\eventtokenfont ␣BELOW} & $6.77883\times10^{-7}$ \\
\texttt{\eventtokenfont 797} & $6.97073\times10^{-7}$ \\
\texttt{\eventtokenfont 767} & $8.14367\times10^{-7}$ \\
\texttt{\eventtokenfont ␣978} & $9.14544\times10^{-7}$ \\
\texttt{\eventtokenfont ␣Ukrain} & $9.74397\times10^{-7}$ \\
\bottomrule
\end{tabular}
\end{minipage}
\hfill
\begin{minipage}[t]{0.49\textwidth}
\centering
\begin{tabular}{@{}lr@{}}
\toprule
Token & Ground truth $\mu$ \\
\midrule
\texttt{\eventtokenfont iminary} & $1.783951\times10^{-6}$ \\
\texttt{\eventtokenfont 388} & $1.814818\times10^{-6}$ \\
\texttt{\eventtokenfont ricia} & $1.85303\times10^{-6}$ \\
\texttt{\eventtokenfont Ã} & $2.314821\times10^{-6}$ \\
\texttt{\eventtokenfont employment} & $2.863334\times10^{-6}$ \\
\texttt{\eventtokenfont html} & $3.451055\times10^{-6}$ \\
\texttt{\eventtokenfont ␣SELECT} & $4.080043\times10^{-6}$ \\
\texttt{\eventtokenfont article} & $4.326181\times10^{-6}$ \\
\texttt{\eventtokenfont Si} & $4.56803\times10^{-6}$ \\
\texttt{\eventtokenfont ␣Aside} & $9.890819\times10^{-6}$ \\
\texttt{\eventtokenfont atten} & $1.1462555\times10^{-5}$ \\
\texttt{\eventtokenfont ␣indicative} & $1.2699662\times10^{-5}$ \\
\texttt{\eventtokenfont ␣Harbaugh} & $1.4530803\times10^{-5}$ \\
\texttt{\eventtokenfont ␣Ci} & $2.11075\times10^{-5}$ \\
\texttt{\eventtokenfont ␣Acad} & $2.5733549\times10^{-5}$ \\
\texttt{\eventtokenfont ␣definitions} & $2.808293\times10^{-5}$ \\
\texttt{\eventtokenfont ␣ingredient} & $3.4664296\times10^{-5}$ \\
\texttt{\eventtokenfont ␣loads} & $4.3123397\times10^{-5}$ \\
\texttt{\eventtokenfont ␣boost} & $5.2364278\times10^{-5}$ \\
\texttt{\eventtokenfont ␣Tomb} & $7.9511388\times10^{-5}$ \\
\texttt{\eventtokenfont aed} & $1.25505903\times10^{-4}$ \\
\texttt{\eventtokenfont uss} & $1.44590405\times10^{-4}$ \\
\texttt{\eventtokenfont ␣connect} & $1.58662006\times10^{-4}$ \\
\texttt{\eventtokenfont ␣fortunate} & $2.23511597\times10^{-4}$ \\
\texttt{\eventtokenfont ria} & $2.35639032\times10^{-4}$ \\
\texttt{\eventtokenfont ␣Arch} & $2.56191561\times10^{-4}$ \\
\texttt{\eventtokenfont ␣Let} & $2.63755705\times10^{-4}$ \\
\texttt{\eventtokenfont ␣promise} & $4.82588483\times10^{-4}$ \\
\texttt{\eventtokenfont ␣desired} & $5.50281373\times10^{-4}$ \\
\texttt{\eventtokenfont ␣mere} & $6.15937519\times10^{-4}$ \\
\bottomrule
\end{tabular}
\end{minipage}
\end{table}

\begin{table}[p]
\centering
\scriptsize
\setlength{\tabcolsep}{3pt}
\caption{All 60 Token Presence events for GPT-2 Small at $L=30$ and their reference probabilities. Read the left list first, then the right; events are ordered from rarer to more common.}
\label{tab:all-tokens-gpt2-small-l30}
\begin{minipage}[t]{0.49\textwidth}
\centering
\begin{tabular}{@{}lr@{}}
\toprule
Token & Ground truth $\mu$ \\
\midrule
\texttt{\eventtokenfont ActionCode} & $2.812\times10^{-9}$ \\
\texttt{\eventtokenfont Orderable} & $3.176\times10^{-9}$ \\
\texttt{\eventtokenfont cloneembedreportprint} & $3.343\times10^{-9}$ \\
\texttt{\eventtokenfont rawdownload} & $3.419\times10^{-9}$ \\
\texttt{\eventtokenfont ␣RandomRedditor} & $3.422\times10^{-9}$ \\
\texttt{\eventtokenfont DeliveryDate} & $4.485\times10^{-9}$ \\
\texttt{\eventtokenfont ÃÂÃÂÃÂÃÂ} & $4.929\times10^{-9}$ \\
\texttt{\eventtokenfont GoldMagikarp} & $7.148\times10^{-9}$ \\
\texttt{\eventtokenfont ␣volunte} & $8.268\times10^{-9}$ \\
\texttt{\eventtokenfont ␣srf} & $9.695\times10^{-9}$ \\
\texttt{\eventtokenfont ÃÂÃÂ} & $1.1427\times10^{-8}$ \\
\texttt{\eventtokenfont ␣unintention} & $1.1591\times10^{-8}$ \\
\texttt{\eventtokenfont ␣teasp} & $1.6259\times10^{-8}$ \\
\texttt{\eventtokenfont inventoryQuantity} & $2.697\times10^{-8}$ \\
\texttt{\eventtokenfont ␣referen} & $4.4671\times10^{-8}$ \\
\texttt{\eventtokenfont ␣looph} & $5.4553\times10^{-8}$ \\
\texttt{\eventtokenfont \textbackslash u30bc\textbackslash u30a6\textbackslash u30b9} & $6.8845\times10^{-8}$ \\
\texttt{\eventtokenfont ␣opio} & $8.1701\times10^{-8}$ \\
\texttt{\eventtokenfont ÛÛ} & $9.6763\times10^{-8}$ \\
\texttt{\eventtokenfont aution} & $9.8635\times10^{-8}$ \\
\texttt{\eventtokenfont ␣mosqu} & $1.06689\times10^{-7}$ \\
\texttt{\eventtokenfont \textbackslash u30d8\textbackslash u30e9} & $1.34227\times10^{-7}$ \\
\texttt{\eventtokenfont ␣SetTextColor} & $3.78981\times10^{-7}$ \\
\texttt{\eventtokenfont glomer} & $4.20284\times10^{-7}$ \\
\texttt{\eventtokenfont escription} & $4.38592\times10^{-7}$ \\
\texttt{\eventtokenfont ␣contrace} & $4.65958\times10^{-7}$ \\
\texttt{\eventtokenfont ␣resil} & $5.52023\times10^{-7}$ \\
\texttt{\eventtokenfont ␣sshd} & $7.90799\times10^{-7}$ \\
\texttt{\eventtokenfont AppData} & $8.59069\times10^{-7}$ \\
\texttt{\eventtokenfont ␣\textbackslash u30b5} & $9.75121\times10^{-7}$ \\
\bottomrule
\end{tabular}
\end{minipage}
\hfill
\begin{minipage}[t]{0.49\textwidth}
\centering
\begin{tabular}{@{}lr@{}}
\toprule
Token & Ground truth $\mu$ \\
\midrule
\texttt{\eventtokenfont 748} & $2.385615\times10^{-6}$ \\
\texttt{\eventtokenfont 568} & $2.52272\times10^{-6}$ \\
\texttt{\eventtokenfont 652} & $3.088749\times10^{-6}$ \\
\texttt{\eventtokenfont 021} & $3.58389\times10^{-6}$ \\
\texttt{\eventtokenfont arijuana} & $4.099484\times10^{-6}$ \\
\texttt{\eventtokenfont ␣Wem} & $4.666225\times10^{-6}$ \\
\texttt{\eventtokenfont \textbackslash u30aa} & $4.693658\times10^{-6}$ \\
\texttt{\eventtokenfont Layout} & $5.44136\times10^{-6}$ \\
\texttt{\eventtokenfont ␣363} & $7.860511\times10^{-6}$ \\
\texttt{\eventtokenfont Border} & $8.441393\times10^{-6}$ \\
\texttt{\eventtokenfont ␣Requires} & $1.2932401\times10^{-5}$ \\
\texttt{\eventtokenfont ␣Petroleum} & $1.6701184\times10^{-5}$ \\
\texttt{\eventtokenfont ␣cowork} & $3.0257203\times10^{-5}$ \\
\texttt{\eventtokenfont aten} & $3.8312104\times10^{-5}$ \\
\texttt{\eventtokenfont loaded} & $4.0012317\times10^{-5}$ \\
\texttt{\eventtokenfont ␣marginally} & $4.2360556\times10^{-5}$ \\
\texttt{\eventtokenfont ␣unfounded} & $5.1715971\times10^{-5}$ \\
\texttt{\eventtokenfont ␣vib} & $6.1999679\times10^{-5}$ \\
\texttt{\eventtokenfont ␣massage} & $7.2618226\times10^{-5}$ \\
\texttt{\eventtokenfont fing} & $9.8132645\times10^{-5}$ \\
\texttt{\eventtokenfont ␣glam} & $1.13032715\times10^{-4}$ \\
\texttt{\eventtokenfont ␣Armed} & $1.2007045\times10^{-4}$ \\
\texttt{\eventtokenfont stock} & $1.22980506\times10^{-4}$ \\
\texttt{\eventtokenfont ␣Deity} & $1.3891251\times10^{-4}$ \\
\texttt{\eventtokenfont ␣voluntary} & $1.67620776\times10^{-4}$ \\
\texttt{\eventtokenfont ␣Baby} & $1.87197977\times10^{-4}$ \\
\texttt{\eventtokenfont ␣pathetic} & $2.29775411\times10^{-4}$ \\
\texttt{\eventtokenfont inating} & $2.3309172\times10^{-4}$ \\
\texttt{\eventtokenfont ␣couch} & $3.03255045\times10^{-4}$ \\
\texttt{\eventtokenfont ␣Energy} & $3.25971749\times10^{-4}$ \\
\bottomrule
\end{tabular}
\end{minipage}
\end{table}

\begin{table}[p]
\centering
\scriptsize
\setlength{\tabcolsep}{3pt}
\caption{All 60 Token Presence events for GPT-2 Small at $L=50$ and their reference probabilities. Read the left list first, then the right; events are ordered from rarer to more common.}
\label{tab:all-tokens-gpt2-small-l50}
\begin{minipage}[t]{0.55\textwidth}
\centering
\begin{tabular}{@{}lr@{}}
\toprule
Token & Ground truth $\mu$ \\
\midrule
\texttt{\eventtokenfont ThumbnailImage} & $3.969\times10^{-9}$ \\
\texttt{\eventtokenfont Nitrome} & $4.643\times10^{-9}$ \\
\texttt{\eventtokenfont Orderable} & $4.971\times10^{-9}$ \\
\texttt{\eventtokenfont reportprint} & $5.398\times10^{-9}$ \\
\texttt{\eventtokenfont ␣RandomRedditor} & $5.416\times10^{-9}$ \\
\texttt{\eventtokenfont ␣TheNitrome} & $5.453\times10^{-9}$ \\
\texttt{\eventtokenfont ÃÂÃÂÃÂÃÂÃÂÃÂÃÂÃÂÃÂÃÂÃÂÃÂÃÂÃÂÃÂÃÂ} & $6.567\times10^{-9}$ \\
\texttt{\eventtokenfont ÃÂÃÂÃÂÃÂÃÂÃÂÃÂÃÂ} & $7.011\times10^{-9}$ \\
\texttt{\eventtokenfont DeliveryDate} & $7.108\times10^{-9}$ \\
\texttt{\eventtokenfont ␣pione} & $9.779\times10^{-9}$ \\
\texttt{\eventtokenfont ␣subur} & $1.1113\times10^{-8}$ \\
\texttt{\eventtokenfont ␣carbohyd} & $1.1206\times10^{-8}$ \\
\texttt{\eventtokenfont ␣srf} & $1.6348\times10^{-8}$ \\
\texttt{\eventtokenfont ␣earthqu} & $2.8878\times10^{-8}$ \\
\texttt{\eventtokenfont ikuman} & $3.0623\times10^{-8}$ \\
\texttt{\eventtokenfont ortunately} & $3.0704\times10^{-8}$ \\
\texttt{\eventtokenfont \textbackslash u899a\textbackslash u9192} & $5.6321\times10^{-8}$ \\
\texttt{\eventtokenfont conservancy} & $6.3908\times10^{-8}$ \\
\texttt{\eventtokenfont ModLoader} & $7.9891\times10^{-8}$ \\
\texttt{\eventtokenfont ␣\textbackslash u30b5\textbackslash u30fc\textbackslash u30c6\textbackslash u30a3\textbackslash u30ef\textbackslash u30f3} & $8.1083\times10^{-8}$ \\
\texttt{\eventtokenfont BuyableInstoreAndOnline} & $1.02812\times10^{-7}$ \\
\texttt{\eventtokenfont ␣proport} & $2.43806\times10^{-7}$ \\
\texttt{\eventtokenfont ategory} & $2.90383\times10^{-7}$ \\
\texttt{\eventtokenfont \textbackslash u30fc\textbackslash u30c6\textbackslash u30a3} & $3.73774\times10^{-7}$ \\
\texttt{\eventtokenfont ␣Pwr} & $6.12012\times10^{-7}$ \\
\texttt{\eventtokenfont ␣SetTextColor} & $6.24561\times10^{-7}$ \\
\texttt{\eventtokenfont ␣gobl} & $6.65429\times10^{-7}$ \\
\texttt{\eventtokenfont acebook} & $7.85046\times10^{-7}$ \\
\texttt{\eventtokenfont escription} & $8.97874\times10^{-7}$ \\
\texttt{\eventtokenfont ␣SolidGoldMagikarp} & $9.54816\times10^{-7}$ \\
\bottomrule
\end{tabular}
\end{minipage}
\hfill
\begin{minipage}[t]{0.43\textwidth}
\centering
\begin{tabular}{@{}lr@{}}
\toprule
Token & Ground truth $\mu$ \\
\midrule
\texttt{\eventtokenfont LEASE} & $2.497735\times10^{-6}$ \\
\texttt{\eventtokenfont intendent} & $2.531015\times10^{-6}$ \\
\texttt{\eventtokenfont emort} & $2.591963\times10^{-6}$ \\
\texttt{\eventtokenfont pkg} & $3.469834\times10^{-6}$ \\
\texttt{\eventtokenfont ␣discrep} & $4.214965\times10^{-6}$ \\
\texttt{\eventtokenfont paralleled} & $5.011348\times10^{-6}$ \\
\texttt{\eventtokenfont enhagen} & $5.647727\times10^{-6}$ \\
\texttt{\eventtokenfont udeau} & $5.69348\times10^{-6}$ \\
\texttt{\eventtokenfont anchez} & $6.136545\times10^{-6}$ \\
\texttt{\eventtokenfont flush} & $9.453041\times10^{-6}$ \\
\texttt{\eventtokenfont ␣metabolites} & $1.5720263\times10^{-5}$ \\
\texttt{\eventtokenfont ␣Croat} & $2.0371159\times10^{-5}$ \\
\texttt{\eventtokenfont HHHH} & $2.58156\times10^{-5}$ \\
\texttt{\eventtokenfont inho} & $4.3729255\times10^{-5}$ \\
\texttt{\eventtokenfont ILE} & $5.5443441\times10^{-5}$ \\
\texttt{\eventtokenfont ␣normalized} & $5.7964509\times10^{-5}$ \\
\texttt{\eventtokenfont ␣Sorceress} & $6.1175284\times10^{-5}$ \\
\texttt{\eventtokenfont ␣Arlington} & $7.5363656\times10^{-5}$ \\
\texttt{\eventtokenfont ␣forfeit} & $9.0493806\times10^{-5}$ \\
\texttt{\eventtokenfont ␣Plymouth} & $9.4129842\times10^{-5}$ \\
\texttt{\eventtokenfont ␣Lies} & $1.10739747\times10^{-4}$ \\
\texttt{\eventtokenfont nex} & $1.1683334\times10^{-4}$ \\
\texttt{\eventtokenfont ␣Antarctica} & $1.55974645\times10^{-4}$ \\
\texttt{\eventtokenfont jected} & $1.9980935\times10^{-4}$ \\
\texttt{\eventtokenfont ␣queue} & $2.3909146\times10^{-4}$ \\
\texttt{\eventtokenfont ␣Ale} & $4.79594106\times10^{-4}$ \\
\texttt{\eventtokenfont ␣artifacts} & $5.78806968\times10^{-4}$ \\
\texttt{\eventtokenfont ␣Dun} & $6.11178344\times10^{-4}$ \\
\texttt{\eventtokenfont ␣offices} & $7.24431942\times10^{-4}$ \\
\texttt{\eventtokenfont ␣Sal} & $9.52690141\times10^{-4}$ \\
\bottomrule
\end{tabular}
\end{minipage}
\end{table}

\begin{table}[p]
\centering
\scriptsize
\setlength{\tabcolsep}{3pt}
\caption{All 66 Token Presence events for Gemma-2 at $L=10$ and their reference probabilities. Read the left list first, then the right; events are ordered from rarer to more common.}
\label{tab:all-tokens-gemma-2b-it-l10}
\begin{minipage}[t]{0.49\textwidth}
\centering
\begin{tabular}{@{}lr@{}}
\toprule
Token & Ground truth $\mu$ \\
\midrule
\texttt{\eventtokenfont ␣meio} & $1.084\times10^{-9}$ \\
\texttt{\eventtokenfont ␣\textbackslash u043f\textbackslash u0440\textbackslash u044a} & $1.334\times10^{-9}$ \\
\texttt{\eventtokenfont ␣Cita} & $1.406\times10^{-9}$ \\
\texttt{\eventtokenfont \textbackslash u6b63\textbackslash u5728} & $1.45\times10^{-9}$ \\
\texttt{\eventtokenfont cei} & $1.601\times10^{-9}$ \\
\texttt{\eventtokenfont opho} & $2.166\times10^{-9}$ \\
\texttt{\eventtokenfont McC} & $2.345\times10^{-9}$ \\
\texttt{\eventtokenfont Hb} & $2.65\times10^{-9}$ \\
\texttt{\eventtokenfont endid} & $4.994\times10^{-9}$ \\
\texttt{\eventtokenfont ␣Dial} & $9.451\times10^{-9}$ \\
\texttt{\eventtokenfont ␣decoration} & $1.0744\times10^{-8}$ \\
\texttt{\eventtokenfont Animals} & $1.0839\times10^{-8}$ \\
\texttt{\eventtokenfont ␣utilized} & $1.0885\times10^{-8}$ \\
\texttt{\eventtokenfont \textbackslash u6743} & $1.2994\times10^{-8}$ \\
\texttt{\eventtokenfont Zoo} & $1.793\times10^{-8}$ \\
\texttt{\eventtokenfont Vish} & $1.9079\times10^{-8}$ \\
\texttt{\eventtokenfont vano} & $1.9914\times10^{-8}$ \\
\texttt{\eventtokenfont ␣sweats} & $2.0707\times10^{-8}$ \\
\texttt{\eventtokenfont ␣articulated} & $2.2923\times10^{-8}$ \\
\texttt{\eventtokenfont helia} & $5.1906\times10^{-8}$ \\
\texttt{\eventtokenfont ␣Sylvain} & $6.5284\times10^{-8}$ \\
\texttt{\eventtokenfont ␣Skinny} & $7.6772\times10^{-8}$ \\
\texttt{\eventtokenfont avor} & $1.36062\times10^{-7}$ \\
\texttt{\eventtokenfont Cassie} & $1.64704\times10^{-7}$ \\
\texttt{\eventtokenfont ␣Kjell} & $1.65677\times10^{-7}$ \\
\texttt{\eventtokenfont ␣hastily} & $1.88567\times10^{-7}$ \\
\texttt{\eventtokenfont ␣whe} & $2.76225\times10^{-7}$ \\
\texttt{\eventtokenfont ␣unsteady} & $3.01208\times10^{-7}$ \\
\texttt{\eventtokenfont ␣disliked} & $3.98665\times10^{-7}$ \\
\texttt{\eventtokenfont ␣pep} & $4.54393\times10^{-7}$ \\
\texttt{\eventtokenfont Madame} & $5.48416\times10^{-7}$ \\
\texttt{\eventtokenfont ␣Ace} & $6.51134\times10^{-7}$ \\
\texttt{\eventtokenfont Maxwell} & $6.83873\times10^{-7}$ \\
\bottomrule
\end{tabular}
\end{minipage}
\hfill
\begin{minipage}[t]{0.49\textwidth}
\centering
\begin{tabular}{@{}lr@{}}
\toprule
Token & Ground truth $\mu$ \\
\midrule
\texttt{\eventtokenfont ␣shedding} & $9.74816\times10^{-7}$ \\
\texttt{\eventtokenfont attered} & $1.015341\times10^{-6}$ \\
\texttt{\eventtokenfont ␣coating} & $1.229574\times10^{-6}$ \\
\texttt{\eventtokenfont ␣mostly} & $1.364986\times10^{-6}$ \\
\texttt{\eventtokenfont ␣mor} & $1.376871\times10^{-6}$ \\
\texttt{\eventtokenfont ␣disapproval} & $1.803206\times10^{-6}$ \\
\texttt{\eventtokenfont ␣nurturing} & $2.100988\times10^{-6}$ \\
\texttt{\eventtokenfont ␣tank} & $2.570783\times10^{-6}$ \\
\texttt{\eventtokenfont ␣groans} & $5.677432\times10^{-6}$ \\
\texttt{\eventtokenfont ␣bison} & $6.391679\times10^{-6}$ \\
\texttt{\eventtokenfont then} & $6.78363\times10^{-6}$ \\
\texttt{\eventtokenfont ␣mural} & $7.759179\times10^{-6}$ \\
\texttt{\eventtokenfont ␣Aur} & $9.377181\times10^{-6}$ \\
\texttt{\eventtokenfont mills} & $1.2066338\times10^{-5}$ \\
\texttt{\eventtokenfont ␣eccentric} & $1.218224\times10^{-5}$ \\
\texttt{\eventtokenfont ␣sharper} & $1.23574\times10^{-5}$ \\
\texttt{\eventtokenfont ␣cutter} & $1.5257824\times10^{-5}$ \\
\texttt{\eventtokenfont ␣ramp} & $1.9968653\times10^{-5}$ \\
\texttt{\eventtokenfont ␣Casper} & $2.0307949\times10^{-5}$ \\
\texttt{\eventtokenfont ␣scenic} & $2.1126934\times10^{-5}$ \\
\texttt{\eventtokenfont ␣longer} & $2.2296295\times10^{-5}$ \\
\texttt{\eventtokenfont ␣railing} & $4.2092146\times10^{-5}$ \\
\texttt{\eventtokenfont ␣Robin} & $5.653832\times10^{-5}$ \\
\texttt{\eventtokenfont ␣shifting} & $7.2919989\times10^{-5}$ \\
\texttt{\eventtokenfont ␣Avon} & $7.9591409\times10^{-5}$ \\
\texttt{\eventtokenfont ␣unsettling} & $1.10149827\times10^{-4}$ \\
\texttt{\eventtokenfont ␣resolute} & $1.18333068\times10^{-4}$ \\
\texttt{\eventtokenfont ␣librarian} & $1.21119221\times10^{-4}$ \\
\texttt{\eventtokenfont ␣steady} & $1.23186823\times10^{-4}$ \\
\texttt{\eventtokenfont ana} & $1.38252377\times10^{-4}$ \\
\texttt{\eventtokenfont ␣brooding} & $1.72549393\times10^{-4}$ \\
\texttt{\eventtokenfont ␣whimsical} & $2.31835613\times10^{-4}$ \\
\texttt{\eventtokenfont ␣distant} & $5.34711289\times10^{-4}$ \\
\bottomrule
\end{tabular}
\end{minipage}
\end{table}

\begin{table}[p]
\centering
\scriptsize
\setlength{\tabcolsep}{4pt}
\caption{All Profanity Classifier events and reference probabilities. The event is $f(\mathbf{y})>\kappa$. A dash means that this model--length--threshold combination is absent from the experimental inventory.}
\label{tab:all-profanity-events}
\begin{tabular}{@{}rrrr@{}}
\toprule
Cutoff $\kappa$ & GPT-2, $L=10$ & GPT-2, $L=50$ & Gemma, $L=10$ \\
\midrule
0.1 & $2.8353515625\times10^{-1}$ & $6.888671875\times10^{-2}$ & $7.9390625\times10^{-2}$ \\
0.15 & $2.11609375\times10^{-1}$ & $5.83203125\times10^{-2}$ & $5.4046875\times10^{-2}$ \\
0.2 & $1.6591015625\times10^{-1}$ & $5.11953125\times10^{-2}$ & $3.921875\times10^{-2}$ \\
0.25 & $1.33453125\times10^{-1}$ & $4.6140625\times10^{-2}$ & $3.0921875\times10^{-2}$ \\
0.3 & $1.085703125\times10^{-1}$ & $4.18125\times10^{-2}$ & $2.615625\times10^{-2}$ \\
0.35 & $8.918359375\times10^{-2}$ & $3.82890625\times10^{-2}$ & $2.28125\times10^{-2}$ \\
0.4 & $7.340625\times10^{-2}$ & $3.507421875\times10^{-2}$ & $1.965625\times10^{-2}$ \\
0.45 & $6.056640625\times10^{-2}$ & $3.22734375\times10^{-2}$ & $1.66875\times10^{-2}$ \\
0.5 & $4.980859375\times10^{-2}$ & $2.970703125\times10^{-2}$ & $1.4265625\times10^{-2}$ \\
0.55 & $4.091015625\times10^{-2}$ & $2.7375\times10^{-2}$ & $1.21875\times10^{-2}$ \\
0.6 & $3.321484375\times10^{-2}$ & $2.5140625\times10^{-2}$ & $1.0421875\times10^{-2}$ \\
0.65 & $2.64765625\times10^{-2}$ & $2.32109375\times10^{-2}$ & $8.4375\times10^{-3}$ \\
0.7 & $2.061328125\times10^{-2}$ & $2.10859375\times10^{-2}$ & $7.109375\times10^{-3}$ \\
0.75 & $1.58671875\times10^{-2}$ & $1.877734375\times10^{-2}$ & $5.828125\times10^{-3}$ \\
0.8 & $1.1671875\times10^{-2}$ & $1.6671875\times10^{-2}$ & $4.578125\times10^{-3}$ \\
0.85 & $7.86328125\times10^{-3}$ & $1.43359375\times10^{-2}$ & $3.28125\times10^{-3}$ \\
0.9 & $4.6953125\times10^{-3}$ & $1.187109375\times10^{-2}$ & $1.9375\times10^{-3}$ \\
0.95 & $2.14453125\times10^{-3}$ & $8.4296875\times10^{-3}$ & $7.5\times10^{-4}$ \\
0.99 & $4.921875\times10^{-4}$ & $4.0546875\times10^{-3}$ & $3.73527547\times10^{-4}$ \\
0.995 & $2.8125\times10^{-4}$ & $2.9375\times10^{-3}$ & $2.18535188\times10^{-4}$ \\
0.999 & $8.59375\times10^{-5}$ & $1.44140625\times10^{-3}$ & $6.25\times10^{-5}$ \\
\bottomrule
\end{tabular}
\end{table}

\begin{table}[p]
\centering
\scriptsize
\caption{All Compositional Token Presence events and reference probabilities for GPT-2 Small at $L=10$. Conjunctive events require every listed token in any order; Ordered Sequence events require the listed order but allow intervening tokens. Leading spaces are part of each target token.}
\label{tab:all-compositional-events}
\begin{tabular}{@{}llr@{}}
\toprule
Event type & Target tokens & Ground truth $\mu$ \\
\midrule
Conjunctive (AND) & $\{$\texttt{\eventtokenfont ␣connect}, \texttt{\eventtokenfont ␣the}, \texttt{\eventtokenfont ␣was}$\}$ & $3.7222278697068845\times10^{-6}$ \\
Conjunctive (AND) & $\{$\texttt{\eventtokenfont ␣the}, \texttt{\eventtokenfont ␣was}, \texttt{\eventtokenfont ␣village}$\}$ & $8.061106367685\times10^{-5}$ \\
Conjunctive (AND) & $\{$\texttt{\eventtokenfont ␣a}, \texttt{\eventtokenfont ␣the}, \texttt{\eventtokenfont ␣king}$\}$ & $1.981384612154\times10^{-4}$ \\
Conjunctive (AND) & $\{$\texttt{\eventtokenfont ␣the}, \texttt{\eventtokenfont ␣was}, \texttt{\eventtokenfont ␣people}$\}$ & $5.053878165782\times10^{-4}$ \\
Conjunctive (AND) & $\{$\texttt{\eventtokenfont ␣the}, \texttt{\eventtokenfont ␣was}, \texttt{\eventtokenfont ␣and}$\}$ & $6.57273504138\times10^{-3}$ \\
Ordered Sequence (SEQ) & \texttt{\eventtokenfont ␣the} $\rightarrow$ \texttt{\eventtokenfont ␣was} $\rightarrow$ \texttt{\eventtokenfont ␣connect} & $1.246940501071\times10^{-6}$ \\
Ordered Sequence (SEQ) & \texttt{\eventtokenfont ␣the} $\rightarrow$ \texttt{\eventtokenfont ␣was} $\rightarrow$ \texttt{\eventtokenfont ␣village} & $5.749512319003\times10^{-6}$ \\
Ordered Sequence (SEQ) & \texttt{\eventtokenfont ␣a} $\rightarrow$ \texttt{\eventtokenfont ␣the} $\rightarrow$ \texttt{\eventtokenfont ␣king} & $3.01940971556\times10^{-5}$ \\
Ordered Sequence (SEQ) & \texttt{\eventtokenfont ␣the} $\rightarrow$ \texttt{\eventtokenfont ␣was} $\rightarrow$ \texttt{\eventtokenfont ␣people} & $1.033016124585\times10^{-4}$ \\
Ordered Sequence (SEQ) & \texttt{\eventtokenfont ␣the} $\rightarrow$ \texttt{\eventtokenfont ␣was} $\rightarrow$ \texttt{\eventtokenfont ␣and} & $2.589375125567\times10^{-3}$ \\
\bottomrule
\end{tabular}
\end{table}
\clearpage